\documentclass[10pt,twocolumn,letterpaper]{article}

\usepackage{LaTeX/iccv}
\usepackage{times}
\usepackage{epsfig}
\usepackage{graphicx}
\usepackage{amsmath}
\usepackage{amssymb}

\usepackage{multirow}
\usepackage{adjustbox}
\usepackage{booktabs}
\usepackage{color}
\usepackage{comment}
\usepackage{subcaption}
\newcommand{\Real}{\mathbb{R}}
\newcommand{\tr}{\mathbf{t}}   % translation vector
\newcommand{\loss}{{\cal L}}  % loss function 
\newcommand{\X}{\mathbf{X}}   % 3D points
\newcommand{\x}{\mathbf{x}}   % 2D points
\newcommand{\z}{\mathbf{z}}  % 
\newcommand{\vv}{\mathbf{v}} 
\newcommand{\bias}{\mathbf{b}}  % bias

\definecolor{mycolor2}{rgb}{0.6, 0.6, 1}

\definecolor{mycolor}{rgb}{1, 0.6, 0}

\newcommand{\norm}[1]{\left\lVert#1\right\rVert}

\usepackage[pagebackref=true,breaklinks=true,letterpaper=true,colorlinks,bookmarks=false]{hyperref}

\iccvfinalcopy % *** Uncomment this line for the final submission

\def\iccvPaperID{9310} % *** Enter the ICCV Paper ID here

\ificcvfinal\fi

\begin{document}

%%%%%%%%% TITLE
\title{Deep Permutation Equivariant Structure from Motion}

\author{Dror Moran$^1$* \hspace{1cm}  
Hodaya Koslowsky$^1$*  \hspace{1cm}
Yoni Kasten$^1$ \\       
Haggai Maron$^2$ \hspace{1cm}
Meirav Galun$^1$ \hspace{1cm}  Ronen Basri$^1$ \\ \\   $^1$Weizmann Institute of Science \hspace{2cm} $^2$NVIDIA Research 
}

%Weizmann Institute of Science  NVIDIA Research \\
% \and

\maketitle
\ificcvfinal\thispagestyle{empty}\fi

{\let\thefootnote\relax\footnotetext{*Equal contributors}}

%%%%%%%%% ABSTRACT
\begin{abstract}
Existing deep methods produce highly accurate 3D reconstructions in stereo and multiview stereo settings, i.e., when cameras are both internally and externally calibrated. Nevertheless, the challenge of simultaneous recovery of camera poses and 3D scene structure in multiview settings with deep networks is still outstanding. 
Inspired by projective factorization for Structure from Motion (SFM) and by deep matrix completion techniques, we propose a neural network architecture that, given a set of point tracks in multiple images of a static scene, recovers both the camera parameters and a (sparse) scene structure by minimizing an unsupervised reprojection loss. Our network architecture is designed to respect the structure of the problem: the sought output is equivariant to permutations of both cameras and scene points. 
Notably, our method does not require initialization of camera parameters or 3D point locations. We test our architecture in two setups: (1) single scene reconstruction and (2)  learning from multiple scenes. Our experiments, conducted on a variety of datasets in both internally calibrated and uncalibrated settings, indicate that our method accurately recovers pose and structure, on par with classical state of the art methods. Additionally, we show that a pre-trained network can be used to reconstruct novel scenes using inexpensive fine-tuning with no loss of accuracy.
\end{abstract}

%%%%%%%%% BODY TEXT
\section{Introduction}
\emph{Structure from motion} (SFM), i.e., the problem of camera pose and 3D structure recovery from images of a stationary scene, is a fundamental problem that was traditionally approached with tools from projective geometry and optimization \cite{HartleyZisserman,snavely2006photo,onur2017survey}. The rise of deep neural networks has led to a flux of new, network-based algorithms for pose estimation and 3D reconstruction. Owing to their ability to encode suitable priors and to their effective optimization with stochastic gradient descent, these algorithms were shown to achieve state of the art results in a number of tasks including binocular and multiview stereo, i.e., in reconstruction problems in which camera parameters are known~\cite{cheng2020stereo,Huang2018DeepMVS,Kar2017,Yao2018MVSNet}. Nevertheless, despite few attempts (see Section~\ref{sec:pervious_work}), the challenge of simultaneous recovery of camera poses and 3D scene structure in multiview settings with deep networks is still outstanding.

This paper introduces a deep neural network that addresses SFM in its classical setting. The goal of SFM is, given a collection of point tracks in multiple images of a static scene, to compute the parameters of the cameras and the 3D locations of the points. The quality of a reconstruction is typically evaluated by a reprojection error function, which measures how close the predicted locations of the projected points are to the image positions of the input tracks. Existing methods apply bundle adjustment (BA) either at the end or as part of the algorithm to minimize this reprojection loss \cite{snavely2006photo, schoenberger2016sfm, kasten2019algebraic}.

%The objective of this paper is twofold. Our first objective is to construct a neural network that, through the optimization of its parameters, can solve this classical SFM problem in the single scene scenario. \hm{Maybe: Indeed , we present the first deep architecture that succeeds in this task. As this is single scene optimization may not be scalable to larger scenes, we propose...}\rb{rephrased. let me know what you think} Additionally, since network parameter optimization typically requires many iterations, we seek a network architecture that can be trained effectively on a collection of scenes and then generalize to novel scenes to enable fast recovery of camera pose and 3D structure.

\begin{figure*}
    \centering
    \includegraphics[width=0.75\textwidth]{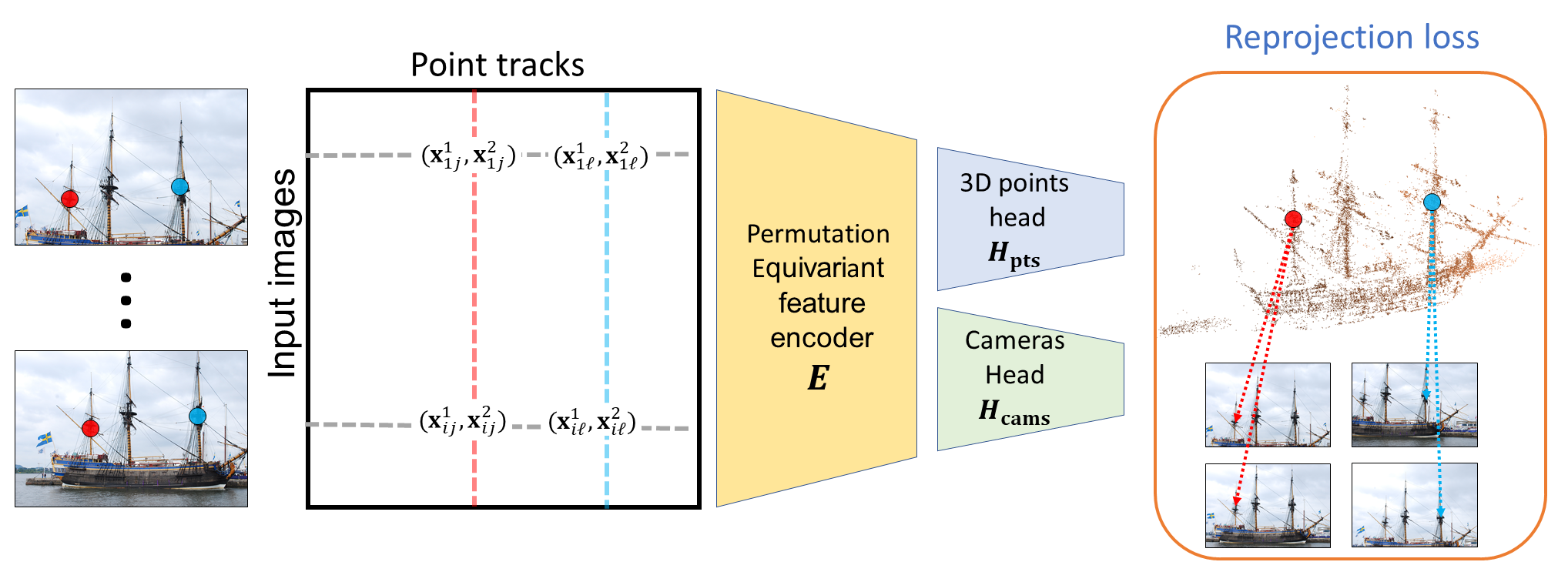}
    \caption{\small Overview of our method. The input is a sparse $m\times n \times 2$ tensor whose rows represent images (cameras) and columns represent point tracks.  Each filled entry in the tensor specifies the 2D coordinates of a 3D point in a specific image. This tensor is processed by a permutation equivariant feature encoder that respects the data symmetries: reoredeing of the rows and the columns of the input tensor. The model  outputs a set of points and a set of camera matrices using two dedicated sub-networks (heads). The model is optimized using an unsupervised reprojection loss, which minimizes the discrepancies in projecting the recovered 3D points relatively to the 2D positions of the input tracks.}
    \label{fig:architecture}
\end{figure*}

Our work is inspired by both projective factorization approaches to SFM (e.g., \cite{christy1996euclidean,lin2017factorization,magerand2017practical}) and by deep matrix completion networks for, e.g., collaborative filtering applications \cite{berg2017graph, Hartford2018}.  Projective factorization techniques seek to express a matrix of track positions as a function of two matrices, a matrix of camera pose parameters and a matrix of 3D point positions. Deep matrix completion methods opt to predict the missing values in an input matrix by learning a function of its unknown factors. Key to deep matrix completion architectures is their \emph{equivariant} architecture; i.e., reordering the rows or columns of the input matrix should yield the same predictions, reordered accordingly. Analogously to convolutional networks for images, such an equivariant architecture is efficient, requiring fewer parameters than standard MLP alternatives, and imposes an appropriate inductive bias that promotes training with fewer samples. 

Below we introduce an equivariant network architecture for SFM (see Figure~\ref{fig:architecture}) that enables simultaneous recovery of camera parameters and scene structure in both calibrated and uncalibrated settings. Specifically, given a tensor of track positions, possibly with many missing elements, we seek to express this tensor as a function of the unknown camera pose parameters and 3D point positions. We use an equivariant network architecture; changing the order of images (rows of the track tensor) yields the same reordering of the recovered cameras; changing the order of the point tracks (tensor columns) yields an equivalent reorder of the recovered 3D points. Our network can be applied in a single scene scenario, in which for a given scene the weights are optimized to directly minimize a reprojection loss. This minimization does not require initialization of either camera parameters or scene structure, yet it achieves accurate recovery of poses and scene structure, on par with state of the art methods. 

A significant contribution of this paper is in showing that this type of architecture also enables generalization to novel scenes. Specifically, we use unsupervised learning to train our network on multiple scenes and then apply the network at inference time to novel scenes. Our experiments indicate that a short refinement step and standard bundle adjustment enables accurate prediction of camera poses and 3D point positions for novel scenes in just a few minutes for scenes including hundreds of images and hundreds of thousands of track points. We finally compare our method, in addition to state of the art baselines, also to two novel network architectures: a network based on DeepSets \cite{zaheer2017deep} and a graph neural network. We discuss the advantages and weaknesses of these approaches.

We believe that our method is a step forward in using the deep learning machinery for solving large and challenging SFM problems. Specifically, our method can be leveraged in the future for constructing an end-to-end model that takes a set of images and outputs the camera positions and 3D points coordinates directly without any additional pre-processing and post-processing steps. Our code and data
are available at \url{https://github.com/drormoran/Equivariant-SFM}.

\section{Previous work}  \label{sec:pervious_work}

\paragraph{Classical SFM methods.} Structure from motion has a long history in computer vision~\cite{LonguetHiggins}. Advances in projective geometry and optimization~\cite{HartleyZisserman} yielded many effective algorithms. A popular approach involves a sequential processing of the input images~\cite{snavely2006photo,wu2013towards,schoenberger2016sfm,magerand2017practical}, so that at each step one image is added, the respective camera pose is recovered, and bundle adjustment is applied to refine the accumulated poses and 3D structure. More recently global techniques based on \emph{motion averaging} were proposed. Early methods first denoise the relative rotation between pairs of views, and subsequently denoise the relative translations and scales~\cite{martinec2007robust,arie2012global,chatterjee2013efficient,eriksson2018rotation,wilson2014robust,Ozyesil_2015_CVPR,jiang2013global, birdal2020synchronizing, Tron_2016_CVPR_Workshops}. Recent works apply averaging directly to the fundamental and essential matrices~\cite{kasten2019algebraic,kasten2019gpsfm,geifman2020averaging,sengupta2017}. Another approach uses \emph{factorization-based methods} \cite{christy1996euclidean,lin2017factorization,dai2013projective,kennedy2016online,magerand2017practical,oliensis2007iterative,sturm1996factorization,ueshiba1998factorization, ARRIGONI201895}. These methods factor a tensor of point tracks %\yk{\cite{lin2017factorization,dai2013projective}  handle missing data} 
into a product of unknown camera matrices, depth values, and 3D point locations. These methods typically yield very large optimization problems and are often approached by splitting the problem into smaller subproblems. Finally,  \emph{Initialization-free-bundle-adjustment methods} \cite{hong2016projective,hong2017revisiting,hong2018pose} use a variable projection method, designed to optimally eliminate subsets of the unknowns, to allow better convergence of bundle adjustment from arbitrary initializations.

\paragraph{Deep SFM methods.} The rise of deep neural networks has led to new, network-based algorithms for pose estimation and 3D reconstruction. Specifically, \cite{ummenhofer2017demon,vijayanarasimhan2017sfmnet,zhou2017ego} recover camera pose for two or three input images, and a dense depth map for one image. BA-net \cite{tang2018ba} and DeepSFM~\cite{Wei2019DeepSFM} are intended to recover pose in multiview settings, but due to memory limitations (the algorithms rely on fine sampling of the cost volume) their method can handle only a handful of views. The latter also requires initialization of the pose parameters.
IDR~\cite{yariv2020multiview} recovers pose and dense structure (using an implicit neural representation), but it requires near accurate initialization of camera pose as well as masked images, and hence it is unsuitable for general scenes. NeuRoRA \cite{purkait2019neurora} introduced a graph neural network architecture for rotation averaging from noisy and incomplete set of relative pairwise orientations. However, their method does not recover camera locations. Of some relevance also is Posenet~\cite{kendall2015posenet}, which is trained to estimate absolute camera pose for a particular (trained) scene. Finally, \cite{wang2021nerfmm} enables initialization-free camera and depth recovery for setups that involve cameras facing roughly the same direction.  In contrast to these, our method is applied to general scenes with hundreds of images and no initialization. However, like classical methods it only produces a sparse reconstruction.

\paragraph{Learning on sets and equivariance.} Using equivariance as a design principle is a popular approach for constructing efficient neural architectures, see, for example, \cite{cohen2016group,worrall2017harmonic,ravanbakhsh2017equivariance,kondor2018generalization,esteves2018learning,maron2018invariant}. Here, we focus on equivariant architectures for set-structured data. 

Learning on set-structured data, where each data point consists of several items and the learning task is invariant or equivariant to their order, is a prominent research direction in recent years. \cite{ravanbakhsh2016deep, zaheer2017deep, qi2017pointnet} pioneered this area by suggesting the first universal deep architectures for this setup. \cite{lee2019set} extended these works by incorporating attention mechanisms. More closely related to our work is \cite{Hartford2018}, which considered a setup that models interaction across several sets: the input is a matrix (more generally, a tensor) and the learning task is equivariant to permutations of both its rows and its columns (and other dimensions in the general case). The paper characterized the maximally expressive linear equivariant layers for this setup and used them for constructing equivariant deep models. These models were shown to perform well in matrix completion and recommender system applications. This learning setup was later generalized to sets of arbitrary symmetric elements \cite{maron2020learning} and to hierarchical structures \cite{wang2020equivariant}.

\section{Approach}
%\subsection{Motivation}
%\hm{discuss matrix factorization methods - accurate but heavy. we propose an approach that builds on the same data representation but is more scalable and enables learning. }

\subsection{Problem definition}
In structure from motion we assume a stationary scene is viewed by $m$ unknown camera matrices, $P_1,...,P_m$. Each camera is represented by a $3 \times 4$ matrix representing maps between projective spaces $P_i:\mathbb{P}^3 \longrightarrow \mathbb{P}^2$. In uncalibrated settings these are general matrices, defined up to scale, while in calibrated settings these represent camera positions  and orientations in the following format $P_i=[R_i|\tr_i]$ with $R_i \in SO(3)$ representing camera orientation and $\tr_i \in \Real^3$ so that camera position given by $-R_i^T\tr_i$. Let $\X$ denote a 3D scene point represented in homogeneous coordinates as $\X=(X^1,X^2,X^3,1)^T \in \mathbb{P}^3$. Its projection onto the $i$'th image is given by $\x = (x^1,x^2,1)^T \propto P\X \in \mathbb{P}^2$, where the symbol '$\propto$' denotes equality up to scale.

As with common SFM algorithms, given images of a static scene, we address the SFM problem after feature points are detected and matched, and after outlier matches are removed by robust recovery of essential or fundamental matrices between pairs of images. The input to our algorithm includes a set of tracks $T_1,...,T_n$, where each track is a set of 2D point positions, $T_j=\{\x_{ij}\}_{i \in C_j}$, and $C_j \subseteq [m]$ denotes images in which  $\X_j$ is detected. The collection of tracks is arranged in a (typically sparse) \emph{measurement tensor} $M$ of size $m \times n \times 2$ such that $(M_{ij1},M_{ij2},1)^T=\x_{ij}$, or these entries are empty if $\X_j$ is not detected in $I_i$. Given $M$, we aim to recover the camera matrices  ${\cal P} = \{ P_1,...,P_m\}$ and the 3D positions ${\cal X} =  \{ \X_1,\X_2,...\X_n \}$ of the respective point tracks.

\subsection{Model}
Our main design motivation is building a model that respects the symmetries of the task above. We will now describe these symmetries, define a suitable model and discuss its advantages.

\begin{figure}
    \centering
    \includegraphics[width=0.3\textwidth]{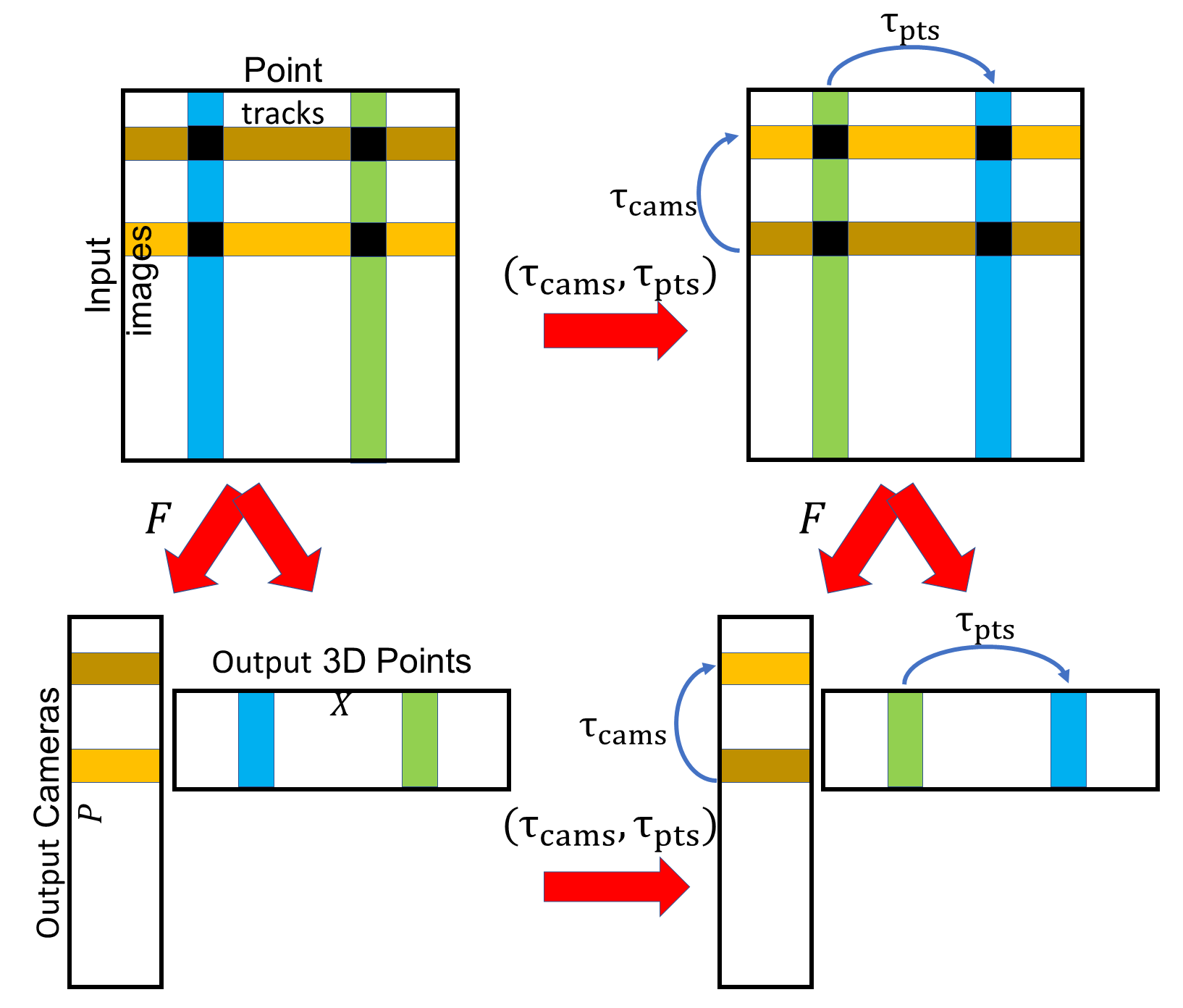}
    \caption{\small Predicting a set of camera positions and 3D points from an input measurement tensor $M$ is equivariant to reordering of the points and the cameras, represented by the pair of permutations $(\tau_{\mathrm{cams}} ,\tau_{\mathrm{pts}})$. This is illustrated in the commutative diagram above.}
    \label{fig:equi}
\end{figure}

\paragraph{Task symmetries.} Our network takes as input a sparse $m \times n \times 2$ tensor $M$ and outputs two matrices: $P\in \Real^{m \times 12}$ that represents the set of cameras and $X\in \Real^{n \times 3}$ that represents a set of points. Importantly, our model should respect the order of the cameras ${\cal P}$ and the scene points ${\cal X}$ in the measurements tensor $M$. More formally, let $S_d$ denote the group of permutations on $d$ elements and let $\tau_{\mathrm{cams}} \in S_m$ and $\tau_{\mathrm{pts}} \in S_n$.  Intuitively, we can think of these permutations as a specific ordering of the cameras and the 3D points. $S_n$ and $S_m$ act on our measurement tensor $M$ by permuting its rows and columns (Figure~\ref{fig:equi}): $((\tau_{\mathrm{cams}},\tau_\mathrm{pts})\cdot M)_{ij}=M_{\tau_{\mathrm{cams}}^{-1}(i),\tau_{\mathrm{pts}}^{-1}(j)}$. As the order of cameras and points is arbitrary, it is natural to expect that the outputs will be reordered according to any given input order. This argument suggests that our network $F$ should follow the following transformation rule for all $\tau_{\mathrm{cams}}\in S_m, \tau_{\mathrm{pts}} \in S_n$ and $M\in \Real ^{m \times n \times 2}$:
\begin{equation*}
    F((\tau_{\mathrm{cams}},\tau_{\mathrm{pts}})\cdot M) = (\tau_{\mathrm{cams}},\tau_{\mathrm{pts}})\cdot F(M).
\end{equation*}
Put differently, the network should be equivariant to the action of the direct product of the two groups, namely to $G=S_m \times S_n$. The action of $G$ on the output space should be understood here as $(\tau_{\mathrm{cams}},\tau_{\mathrm{pts}})\cdot P = \tau_{\mathrm{cams}} \cdot P$ and $(\tau_{\mathrm{cams}},\tau_{\mathrm{pts}}) \cdot X = \tau_{\mathrm{pts}} \cdot X$, i.e., the permutations act only on a single dimension of each output matrix. Figure~\ref{fig:equi} illustrates this type of equivariance. We note that equivariance to direct products of permutation groups was first studied in \cite{Hartford2018}.

\paragraph{Equivariant Layers.} Linear equivariant maps serve as the main building blocks for equivariant networks. As shown in \cite{Hartford2018}, assuming a single input and output channel, the space of linear maps that are equivariant to $G$ is 4-dimensional, and it is spanned by the identity, the row sums, the column sums, and the matrix sum. More generally, with multiple input and output channels, any $G$-equivariant affine map $L$ from a feature space with $d$ channels  $\Real^{m \times n \times d}$ to a feature space with $d'$ channels $\Real^{m \times n \times d'}$ can be represented in the following way:
\begin{equation} \label{eq:equilayer}
\begin{split}
 L(\tilde M)_{ij}= & W_1\tilde M_{ij} + W_2\sum_{k=1}^m \tilde M_{kj}+ \\
            & W_3\sum_{l=1}^n \tilde M_{il} + W_4\sum_{k=1}^m\sum_{l=1}^n \tilde M_{kl} + \bias,
\end{split}
\end{equation}
where for the input tensor $\tilde M_{ij}$ is a vector in $\Real^d$, $W_i\in \Real^{d'\times d}$ for $i=1,...,4$, and $\bias\in\Real^{d'}$ are learnable parameters. Equivalently, $L$ can be represented as an $mnd' \times mnd$ matrix with a parameter sharing scheme governed by \eqref{eq:equilayer} (see \cite{ravanbakhsh2017equivariance,maron2018invariant} for the connection between equivariance and parameter sharing) and possibly an additional bias term. Here we replace sums by averages over the non-empty entries of $M$ to maintain indifference to missing data.

\paragraph{Network architecture.} Our network is composed of three main parts (see Figure \ref{fig:architecture}): a shared feature encoder $E$ and two sub-networks (heads), $H_{\mathrm{cams}}$ and $H_{\mathrm{pts}}$, that produce the recovered cameras and 3D points from the features encoded by $E$. The shared feature encoder takes the input tensor $M\in \Real^{m \times n \times 2}$ and outputs a latent representation $\Real^{m \times n \times d}$ while maintaining the sparsity pattern of the input.  As usually done when constructing equivariant networks \cite{zaheer2017deep,Hartford2018}, the feature encoder is comprised of a composition of several linear equivariant layers interleaved with pointwise nonlinearities such as the ReLU function:
\begin{equation}
    E=L_k\circ \sigma \circ \cdots \circ \sigma \circ L_1,
\end{equation}
where $L_i,~ i=1...k$, are $G$-equivariant layers as in \eqref{eq:equilayer} and $\sigma$ is the pointwise nonlinearity.
Importantly, the pointwise nonlinearities maintain equivariance to the permutation action, implying that the complete composition is equivariant. 

The first head $H_{\mathrm{cams}}$ takes as input a pooled average of the latent representation, $(1/|\bar C_i|)\sum_{j \in \bar C_i} E(M)_{i,j}\in \Real^{m\times d}$, where $\bar C_i$ denotes the set of track points detected in image $i$, and outputs a matrix representation of the camera parameters. In the uncalibrated setting this results in a matrix $P\in \Real^{m \times 12}$, and each row is used to populate the corresponding $3 \times 4$ camera matrix $P_i$. In the calibrated setting this head produces an $m \times 7$ matrix, and we use the first four components in each row to construct a quaternion vector that represents the $i$'th camera orientation and the last three components to construct the camera translation vector. The second head $H_{\mathrm{pts}}$ takes as input $(1/|C_j|)\sum_{i \in C_j} E(M)_{i,j} \in \Real^{n\times d}$, where $C_j$ denotes the track points corresponding to $X_j$, and outputs a matrix representation of the 3D points $X\in \Real^{n \times 3}$.
%In practice, to maintain indifference to missing data, the pooled versions of the latent representations are replaced by averages over the non-missing entries of $M$.
As the pooled versions of the latent representations used by the sub-networks result in single set structures, both heads are implemented as standard equivariant set networks (element-wise fully connected layers) \cite{qi2017pointnet}. 

\paragraph{Discussion.} There are several important advantages to using our equivariant architecture compared to standard fully connected architectures. Above all, our architecture encodes the structure of the task into the model, thus providing a strong inductive bias. Additional benefits include (1) considerable reduction in the number of parameters: for example, the linear part of the layer described in \eqref{eq:equilayer} has $4dd'$ free parameters instead of $n^2m^2dd'$ free parameters in a suitable fully connected architecture; (2) improved efficiency due to small matrix multiplications as seen in Equation \eqref{eq:equilayer}; (3) better generalization: training on any input tensor exposes the network to all its different realizations by different point and camera orders; and (4) Our architecture has the crucial ability to handle variable sized inputs and varying patterns of missing entries in the input tensor. Finally, while equivariant/invariant models might suffer from loss of expressive power \cite{xu2018powerful,maron2019universality}, it was proven that the network architecture we use enjoys a universal approximation property (for full tensors under mild assumptions on the input domain) \cite{maron2020learning}, i.e., with sufficient layers and channels and with a proper set of weights it is capable of approximating any continuous $G$-equivariant function.

\subsection{Loss function}  \label{sec:loss}

Our network minimizes a loss function made of two terms, a reprojection loss and a hinge loss. The reprojection loss, similar to bundle adjustment, minimizes the discrepancies in projecting the recovered 3D points $\X_k$ compared to their detected locations in the images (see Figure \ref{fig:architecture}). Additionally, we use a hinge loss to discourage the recovered depth values from vanishing or becoming negative. 

Let $\x_{ij}$ be a point from track $T_j$ detected in image $I_i$. Let the unknown $P_i$ denote the respective camera, and let $\X_j$ denote the unknown 3D position of the points in $T_j$. To properly measure projective depth in the uncalibrated case (see discussion in \cite{HartleyZisserman}, Chapter 21 Cheirality) we normalize each camera matrix $P_i$ so that its left $3 \times 3$ block has positive determinant and unit norm third row. (These conditions are fulfilled by construction in the calibrated case.)

We next let
\begin{equation*}  %\label{eq:reproj}
r_{ij} =
\norm{\left(\x_{ij}^1-\frac{P_i^1 \X_j}{P_i^3 \X_j}, \x_{ij}^2-\frac{P_i^2 \X_j}{P_i^3 \X_j} \right)},
\end{equation*}
where $P_i^k$  denotes the $k$'th row of the camera matrix $P_i$. Thus, $r_{ij}$ measures the reprojection error of $\X_j$ in image $I_i$. 
Next, we define $h_{ij} = \max(0,h-P_i^3 \X_j)$,
%\begin{equation*}
%h_{ij} = \max(0,h-P_i^3 \X_j),
%\end{equation*}
where $h>0$ is a small constant (in our setting, $h=0.0001$). $h_{ij}$ therefore is the hinge loss for the depth value of $\X_j$ in image $I_i$. 

We combine $r_{ij}$ and $h_{ij}$ as follows,
\begin{eqnarray*}
s_{ij} =
    \begin{cases}
    r_{ij},~~ & P_i^3 \X_j \geq h\\
    h_{ij},~~ & P_i^3 \X_j < h,
    \end{cases}
\end{eqnarray*}
and finally define our loss function as
\begin{equation} \label{eq:loss}
  \loss(\{P_i\},\{\X_j\}) =\frac{1}{p} \sum_{i=1}^m \sum_{j=1}^n \xi_{ij}  s_{ij},
\end{equation}
where $m$ denotes the number of cameras, $n$ the number of point tracks, $p=\sum_{j=1}^n|T_j|$ the number of measured projections, and $\xi_{ij} \in \{0,1\}$ is indicating whether point $\X_j$ is detected in image $I_i$. For each detected point $\x_{ij}$, therefore, the loss measures either reprojection error, if its recovered depth value exceeds the threshold $h$, or it measures the hinge loss, if the recovered depth is either negative or near zero. Note that for robustness we use in the reprojection loss the $\ell_2$ measure, as opposed to the standard MSE.
We note that our loss function is not continuous at points in which a depth value is $h$. It is however differentiable almost everywhere with respect to the network weights.

\subsection{Optimization and learning strategies}

We apply our model in two setups. In a \emph{single scene setup} given point tracks from a single scene we initialize the network with random weights and use back propagation to minimize the loss function \eqref{eq:loss}. At the end we use triangulation to estimate the locations of the 3D points and apply bundle adjustment.

In the \emph{learning scenario} we train the model with tracks from multiple scenes. At train time we minimize our loss function \eqref{eq:loss} while alternating between the input scenes. We evaluate success on a validation set and use early stopping to keep the model that reaches the best reprojection error on this set. To test the model on novel scenes, given a measurement tensor we either apply the network to the tensor to get an initial reconstruction, or, we fine tune the model by running back propagation for a predetermined number of epochs. We complete these with bundle adjustment. Inference with the network is extremely fast, taking roughly hundredths of a second on a DGX machine, therefore the total test time is dominated by the time of the optional fine tuning and the bundle adjustment.

\section{Experimental setup} \label{sec:experiments}

We tested our model on a variety of datasets in both single scene and learning scenarios and in both calibrated and uncalibrated settings.

\subsection{Datasets and baselines}

\subsubsection{Uncalibrated setting} 

For this setting we used 39 scans from Olsson's \cite{olsson2011stable} and the VGG \cite{vgg_dataset} datasets. The number of images in these scans vary from 10 to 400 and number of points in the track tensors from 300 to 150K. We compare our method to the following baselines.

\noindent \textbf{VARPRO} \cite{hong2016projective}.
An initialization-free optimization strategy based on variable projection (VarPro) applied to projective bundle adjustment.

\noindent \textbf{GPSFM} \cite{kasten2019gpsfm}. A global algorithm based on averaging of fundamental matrices. The algorithm minimizes an algebraic loss, i.e., it seeks the nearest collection of fundamental matrices that can be realized with projective cameras.

\noindent \textbf{PPSFM} \cite{magerand2017practical}. A sequential method based on incrementally optimizing for projective structure and cameras while incorporating constraints on the sought projective depths.

\subsubsection{Calibrated setting} 

For this setting we used 36 scans from Olsson's dataset \cite{olsson2011stable}. The dataset includes ``ground truth'' camera parameters, including intrinsic and extrinsic (locations and orientations) parameters, which in fact are reconstructed with Olsson's SFM method. The number of cameras ranges from 12 to 419 and the number of points in the track tensor from 319 to 156k.

\noindent\textbf{GESFM} \cite{kasten2019algebraic}. Analogously to GPSFM, this is a global algorithm based on averaging essential matrices. The algorithm minimizes an algebraic loss, i.e., it seeks the nearest collection of essential matrices that can be realized with Euclidean cameras. 

\noindent  \textbf{Linear} \cite{jiang2013global}.
This algorithm enforces the consistency of camera triplets while minimizing an approximate geometric error. (For this algorithm we were unable to find one set of threshold parameters that works on all the datasets, and so for each scan we report the best run.)  

\noindent \textbf{COLMAP} \cite{schoenberger2016sfm,schonberger2016pixelwise,Colmap_implementation} A state-of-the-art sequential method.

\subsection{Implementation details}

\noindent
\textbf {Framework.} Our method was run on NVIDIA Quadro RTX 8000/ RTX 6000/ DGX V100 GPUs. We used PyTorch \cite{paszke2019pytorch} as the deep learning framework, and ADAM optimizer with normalized gradients. For experiments involving graphs (Section \ref{sec:alternative}) we use PyTorch Geometric \cite{fey2019fast}.

\noindent
\textbf{Training.}
In both calibrated and uncalibrated settings we randomly divided the datasets to three parts: 10 scenes for test, 3 for validation, and the rest for training. During training we alternated between the different training scenes, where in each epoch we trained on a random subset of 10-20 images in a scene. We used validation for early stopping. Validation and test were applied to the complete scenes.

\noindent
\textbf{Optimization.}
Since dividing by $P^3_i \X_j$ can result in exploding gradients, during back-propagation we normalized the gradient of $P_i \X_j$ at each step.

\noindent\textbf{Architecture details.} 
The input to our method includes normalized point tracks $\x_{ij}$. In the calibrated setting we normalized the points using the known intrinsic parameters. In the unclibrated setting we used Hartley normalization \cite{hartley1997defense}. For efficiency we designed our encoder to work with sparse matrices. The shared features encoder $E$ has 3 layers each with respectively 256 or 512 feature channels in the single scene and training setups and ReLU activation. The camera head $H_{\mathrm{cams}}$ and 3D point head $H_{\mathrm{pts}}$ each have 2 layers with 256 or 512 channels for optimization and training, respectively. After each layer in $E$ we normalize its features by subtracting their mean.

\noindent\textbf{Hyper-parameter search.} 
We tested our model with different hyper-parameters including: (1) learning rates in $\{1e-2, 1e-3, 1e-4\}$, (2) network width for the encoder $E$ and heads in $\{128, 256, 512\}$, (3) number of layers in these networks $\{2, 3\}$, (4) depth threshold $h \in \{1e-2, 1e-3, 1e-4\}$, (5) std normalization for the layer output in \eqref{eq:equilayer}.

\noindent\textbf{Bundle adjustment}
For bundle adjustment we used the Ceres BA implementation \cite{ceres-solver} with the Huber loss (with parameter 0.1) for robustness and limited the number of iterations to 100.

\subsection{Evaluation}
In the uncalibrated setting we measure accuracy with the average reprojection error, measured in pixels as follows 
\begin{equation*}
    \frac{1}{p} \sum_{i=1}^m \sum_{j=1}^n \xi_{ij} \norm{\left(\x_{ij}^1-\frac{P_i^1 \X_j}{P_i^3 \X_j}, \x_{ij}^2-\frac{P_i^2 \X_j}{P_i^3 \X_j} \right)}.
\end{equation*}
For notation see Sec.~\ref{sec:loss}.
%While our network outputs the positions $\X_j$, we found that re-triangulation of the points using the predicted cameras yields more stable estimates. We therefore measure the reprojection error for our method before BA using these triangulated points.
In the calibrated case we further evaluate our predictions of the external camera parameters. Specifically, we compare our camera orientation predictions with ground truth ones by measuring their angular difference in degrees, as well as the difference between our predicted and ground truth camera locations in meters. For fair comparison, both our method and all the baseline methods were run with the same set of point tracks. For all methods we apply a final post-processing step of bundle adjustment. Below we show results both before and after BA (results before BA for the baseline methods are provided in the supplementary material).

\section{Results}

\begin{table}[tb]
    \hspace{-8pt}
    \setlength\tabcolsep{2pt} % default value: 6pt
    \tiny
    \centering
    \begin{tabular}{c}
        \begin{adjustbox}{max width=\textwidth}
        \aboverulesep=0ex
        \belowrulesep=0ex
        \renewcommand{\arraystretch}{1}
        \begin{tabular}[t]{|l|r|r|rrrrr|}
            \hline
            \multirow{3}{3em}{Scan}& \multirow{3}{3em}{\#Images} & \multirow{3}{3em}{\#Points} & \multicolumn{5}{c|}{\textbf{Reprojection error (pixels)}}\\
            & & & \multicolumn{1}{c}{\textbf{Ours}} & \multirow{2}{2.5em}{\textbf{Ours}} & \multirow{2}{*}{GPSFM} & \multirow{2}{3em}{PPSFM} & \multirow{2}{*}{VarPro} \\
            & & & \multicolumn{1}{c}{\textbf{No BA}} & & & & \\
			\hline
Alcatraz Courtyard & $ 133 $ & $ 23674 $ & $ 1.55 $ & $\mathbf{0.52}$ & $\mathbf{0.52}$ & $ 0.57 $ & $\mathbf{0.52}$\\
Alcatraz Water Tower & $ 172 $ & $ 14828 $ & $ 2.18 $ & $\mathbf{0.47}$ & $ 0.63 $ & $ 0.59 $ & $\mathbf{0.47}$\\
Alcatraz West Side Gardens & $ 419 $ & $ 65072 $ & $ 9.54 $ & $\mathbf{0.76}$ & $ 326.99 $ & $ 1.77 $ & -\\
Basilica Di San Petronio & $ 334 $ & $ 46035 $ & $ 7.9 $ & $ 0.96 $ & $ 60.69 $ & $\mathbf{0.63}$ & -\\
Buddah Statue & $ 322 $ & $ 156356 $ & $ 18.88 $ & $ 2.93 $ & $ 96.96 $ & $\mathbf{0.41}$ & -\\
Buddah Tooth Relic Temple Singapore & $ 162 $ & $ 27920 $ & $ 4.59 $ & $\mathbf{0.6}$ & $ 0.62 $ & $ 0.71 $ & $\mathbf{0.6}$\\
Corridor & $ 11 $ & $ 737 $ & $ 0.3 $ & $\mathbf{0.26}$ & $\mathbf{0.26}$ & $ 0.27 $ & $\mathbf{0.26}$\\
Ecole Superior De Guerre & $ 35 $ & $ 13477 $ & $ 0.75 $ & $\mathbf{0.26}$ & $\mathbf{0.26}$ & $ 0.28 $ & $\mathbf{0.26}$\\
Dinosaur 319 & $ 36 $ & $ 319 $ & $ 2.35 $ & $ 1.53 $ & $\mathbf{0.43}$ & $ 0.47 $ & $\mathbf{0.43}$\\
Dinosaur 4983 & $ 36 $ & $ 4983 $ & $ 1.96 $ & $ 0.57 $ & $\mathbf{0.42}$ & $ 0.47 $ & $\mathbf{0.42}$\\
Doge Palace Venice & $ 241 $ & $ 67107 $ & $ 3.6 $ & $\mathbf{0.6}$ & $ 3.52 $ & $ 0.67 $ & -\\
Eglise du dome & $ 85 $ & $ 84792 $ & $ 1.1 $ & $\mathbf{0.24}$ & $\mathbf{0.24}$ & $ 0.25 $ & -\\
Drinking Fountain Somewhere in Zurich & $ 14 $ & $ 5302 $ & $ 0.33 $ & $\mathbf{0.28}$ & $\mathbf{0.28}$ & $ 0.31 $ & $\mathbf{0.28}$\\
East Indiaman Goteborg & $ 179 $ & $ 25655 $ & $ 3.31 $ & $ 0.99 $ & $ 5.11 $ & $\mathbf{0.67}$ & -\\
Folke Filbyter & $ 40 $ & $ 21150 $ & $ 8.87 $ & $ 8.58 $ & $ 0.82 $ & $\mathbf{0.33}$ & $ 277.89 $\\
Golden Statue Somewhere In Hong Kong & $ 18 $ & $ 39989 $ & $ 0.35 $ & $\mathbf{0.22}$ & $\mathbf{0.22}$ & $ 0.24 $ & $\mathbf{0.22}$\\
Gustav Vasa & $ 18 $ & $ 4249 $ & $ 0.23 $ & $\mathbf{0.16}$ & $\mathbf{0.16}$ & $ 0.17 $ & $\mathbf{0.16}$\\
GustavIIAdolf & $ 57 $ & $ 5813 $ & $ 14.77 $ & $ 5.83 $ & $\mathbf{0.23}$ & $ 0.24 $ & $\mathbf{0.23}$\\
Model House & $ 10 $ & $ 672 $ & $ 0.37 $ & $\mathbf{0.34}$ & $ 1.12 $ & $ 0.4 $ & $\mathbf{0.34}$\\
Jonas Ahlstromer & $ 40 $ & $ 2021 $ & $ 14.38 $ & $ 4.72 $ & $\mathbf{0.18}$ & $ 0.2 $ & $\mathbf{0.18}$\\
Lund University Sphinx & $ 70 $ & $ 32668 $ & $ 3.64 $ & $\mathbf{0.34}$ & $ 0.45 $ & $ 0.37 $ & $\mathbf{0.34}$\\
Nijo Castle Gate & $ 19 $ & $ 7348 $ & $ 0.71 $ & $\mathbf{0.39}$ & $\mathbf{0.39}$ & $ 0.43 $ & $\mathbf{0.39}$\\
Pantheon Paris & $ 179 $ & $ 29383 $ & $ 1.75 $ & $\mathbf{0.49}$ & $ 2.85 $ & $ 0.62 $ & -\\
Park Gate Clermont Ferrand & $ 34 $ & $ 9099 $ & $ 0.61 $ & $\mathbf{0.31}$ & $ 0.32 $ & $ 0.49 $ & $\mathbf{0.31}$\\
Plaza De Armas Santiago & $ 240 $ & $ 26969 $ & $ 5.1 $ & $\mathbf{0.64}$ & $ 3.14 $ & $ 0.71 $ & -\\
Porta San Donato Bologna & $ 141 $ & $ 25490 $ & $ 1.58 $ & $\mathbf{0.4}$ & $ 0.61 $ & $ 3.75 $ & $\mathbf{0.4}$\\
The Pumpkin & $ 195 $ & $ 69335 $ & $ 14.45 $ & $\mathbf{0.38}$ & $\mathbf{0.38}$ & $ 0.42 $ & -\\
Skansen Kronan Gothenburg & $ 131 $ & $ 28371 $ & $ 1.19 $ & $\mathbf{0.41}$ & $ 0.44 $ & $ 0.44 $ & -\\
Skansen Lejonet Gothenburg & $ 368 $ & $ 74423 $ & $ 10.82 $ & $ 2.05 $ & $ 7.48 $ & $\mathbf{1.28}$ & -\\
Smolny Cathedral St Petersburg & $ 131 $ & $ 51115 $ & $ 1.66 $ & $\mathbf{0.46}$ & $\mathbf{0.46}$ & $ 0.5 $ & -\\
Some Cathedral In Barcelona & $ 177 $ & $ 30367 $ & $ 3.67 $ & $\mathbf{0.51}$ & $\mathbf{0.51}$ & $ 0.54 $ & -\\
Sri Mariamman Singapore & $ 222 $ & $ 56220 $ & $ 7.06 $ & $\mathbf{0.61}$ & $ 0.78 $ & $ 0.85 $ & -\\
Sri Thendayuthapani Singapore & $ 98 $ & $ 88849 $ & $ 2.12 $ & $\mathbf{0.31}$ & $ 0.56 $ & $ 0.33 $ & -\\
Sri Veeramakaliamman Singapore & $ 157 $ & $ 130013 $ & $ 6.47 $ & $\mathbf{0.52}$ & $ 1.78 $ & $ 0.66 $ & -\\
Thian Hook Keng Temple Singapore & $ 138 $ & $ 34288 $ & $ 7.59 $ & $\mathbf{0.54}$ & $ 0.55 $ & $ 0.66 $ & $\mathbf{0.54}$\\
King's College U. of Toronto & $ 77 $ & $ 7087 $ & $ 2.27 $ & $ 0.78 $ & $ 2.35 $ & $ 0.26 $ & $\mathbf{0.24}$\\
Tsar Nikolai I & $ 98 $ & $ 37857 $ & $ 6.04 $ & $ 2.43 $ & $ 0.33 $ & $ 0.31 $ & $\mathbf{0.29}$\\
Urban II & $ 96 $ & $ 22284 $ & $ 16.91 $ & $ 6.84 $ & $\mathbf{0.27}$ & $ 0.31 $ & $ 3.61 $\\
			\bottomrule
	\end{tabular} 
	\end{adjustbox}
	\vspace{3pt}      
    \end{tabular}
    \caption{\small Single scene results with our method against baselines in the uncalibrated setup. The table shows average reprojection error before and after BA. (\textit{Smaller is better.}) In a number of experiments VarPro exceeded either memory or runtime limitations. These experiments are marked by the missing entries.}
    \label{tab:Projective_Results}
\end{table}
\begin{table*}[tb]
    \hspace{-8pt}
    \setlength\tabcolsep{1pt} % default value: 6pt
    \tiny
    \centering
    \begin{tabular}{c}
        \begin{adjustbox}{max width=\textwidth}
        \aboverulesep=0ex
        \belowrulesep=0ex
        \renewcommand{\arraystretch}{0.9}
        \begin{tabular}[t]{|l|r|r||rrrrr|rrrrr|rrrrr|}
            \hline
            \multirow{3}{3em}{Scan}& \multirow{3}{3em}{\#Images} & \multirow{3}{3em}{\#Points} & \multicolumn{5}{c|}{$\tr_{\text{error}}$} & \multicolumn{5}{c|}{$R_{\text{error}}$} & \multicolumn{5}{c|}{Reprojection Error} \\
            & & & \multicolumn{1}{c}{\textbf{Ours}} & \multirow{2}{3em}{\textbf{Ours}} & \multirow{2}{3.5em}{GESFM} & \multirow{2}{3em}{Linear} & \multirow{2}{3.5em}{Colmap} &\multicolumn{1}{c}{\textbf{Ours}} & \multirow{2}{3em}{\textbf{Ours}} & \multirow{2}{3.5em}{GESFM} & \multirow{2}{3em}{Linear} & \multirow{2}{3.5em}{Colmap} & \multicolumn{1}{c}{\textbf{Ours}} & \multirow{2}{2.5em}{\textbf{Ours}} & \multirow{2}{3.5em}{GESFM} & \multirow{2}{3em}{Linear} & \multirow{2}{*}{Colmap}  \\
            & & & \multicolumn{1}{c}{\textbf{No BA}} & & & & & \multicolumn{1}{c}{\textbf{No BA}} & & & & & \multicolumn{1}{c}{\textbf{No BA}} & & & & \\
			\hline
Alcatraz Courtyard & $ 133 $ & $ 23674 $ & $ 0.16 $ & $ 0.015 $ & $ 0.259 $ & $\mathbf{0.014}$ & $\mathbf{0.014}$ & $0.619 $ & $ 0.049 $ & $ 0.533 $ & $\mathbf{0.042}$ & $ 0.043 $ & $ 1.64 $ & $ \mathbf{0.81}$ & $ 4.67 $ & $ 1.27 $ & $\mathbf{0.81}$\\
Alcatraz Water Tower & $ 172 $ & $ 14828 $ & $ 0.518$ & $ 0.116 $ & $ 9.147 $ & $ 1.643 $ & $\mathbf{0.115}$ & $0.933 $ & $ 0.23 $ & $ 9.997 $ & $ 1.525 $ & $\mathbf{0.228}$ & $ 2.13 $ & $\mathbf{0.55}$ & $ 25.93 $ & $ 73.72 $ & $\mathbf{0.55}$\\
Buddah Tooth Relic Temple Singapore & $ 162 $ & $ 27920 $ & $0.233 $ & $\mathbf{0.014}$ & $ 1.429 $ & $ 0.125 $ & $ 0.015 $ & $1.03 $ & $\mathbf{0.081}$ & $ 4.709 $ & $ 0.551 $ & $ 0.083 $ & $ 2.06 $ & $\mathbf{0.85}$ & $ 13.22 $ & $ 2.66 $ & $\mathbf{0.85}$\\
Doge Palace Venice & $ 241 $ & $ 67107 $ & $0.342 $ & $ 0.029 $ & $ 1.608 $ & - & $\mathbf{0.012}$ & $1.163 $ & $ 0.211 $ & $ 5.317 $ & - & $\mathbf{0.031}$ & $ 3.62 $ & $ 1.0 $ & $ 22.32 $ & - & $\mathbf{0.98}$\\
Door Lund & $ 12 $ & $ 17650 $ & $ 0.006$ & $\mathbf{0.001}$ & $ (0.973) $ & $\mathbf{0.001}$ & $\mathbf{0.001}$ & $ 0.024$ & $ 0.006 $ & $ (7.552) $ & $\mathbf{0.005}$ & $\mathbf{0.005}$ & $0.32 $ & $\mathbf{0.3}$ & $ (9.21) $ & $\mathbf{0.3}$ & $\mathbf{0.3}$\\
Drinking Fountain Somewhere In Zurich & $ 14 $ & $ 5302 $ & $0.004 $ & $\mathbf{0.002}$ & $ (0.002) $ & $\mathbf{0.002}$ & $\mathbf{0.002}$ & $0.031 $ & $\mathbf{0.007}$ & $ (0.01) $ & $\mathbf{0.007}$ & $\mathbf{0.007}$ & $0.33 $ & $ 0.31 $ & $ (0.27) $ & $ 0.31 $ & $ 0.31 $\\
East Indiaman Goteborg & $ 179 $ & $ 25655 $ & $0.621 $ & $ 0.509 $ & $ 3.099 $ & $ (2.235) $ & $\mathbf{0.065}$ & $3.814 $ & $ 3.117 $ & $ 12.396 $ & $ (3.284) $ & $\mathbf{0.251}$ & $4.13 $ & $ 1.85 $ & $ 32.37 $ & $ (312.9) $ & $\mathbf{0.89}$\\
Ecole Superior De Guerre & $ 35 $ & $ 13477 $ & $0.081 $ & $ 0.005 $ & $ (0.002) $ & $ 0.005 $ & $ 0.005 $ & $0.318 $ & $\mathbf{0.024}$ & $ (0.035) $ & $\mathbf{0.024}$ & $\mathbf{0.024}$ & $0.72 $ & $ 0.34 $ & $ (0.14) $ & $ 0.34 $ & $ 0.34 $\\
Eglise du dome & $ 85 $ & $ 84792 $ & $0.205 $ & $\mathbf{0.01}$ & $ (1.425) $ & $ 0.046 $ & $\mathbf{0.01}$ & $0.808 $ & $ 0.037 $ & $ (3.631) $ & $ 0.162 $ & $\mathbf{0.036}$ & $0.91 $ & $\mathbf{0.27}$ & $ (6.21) $ & $ 0.76 $ & $\mathbf{0.27}$\\
Folke Filbyter & $ 40 $ & $ 21150 $ & $0.125 $ & $ 0.118 $ & $ (0.0) $ & $ 0.123 $ & $\mathbf{0.0}$ & $74.596 $ & $ 70.157 $ & $ (0.148) $ & $ 4.484 $ & $\mathbf{0.036}$ & $10.37 $ & $ 4.29 $ & $ (0.41) $ & $ 6.06 $ & $\mathbf{0.29}$\\
Fort Channing Gate Singapore & $ 27 $ & $ 23627 $ & $0.093 $ & $\mathbf{0.008}$ & $\mathbf{0.008}$ & $ 0.013 $ & $\mathbf{0.008}$ & $0.207 $ & $\mathbf{0.02}$ & $\mathbf{0.02}$ & $ 0.029 $ & $\mathbf{0.02}$ & $0.52 $ & $\mathbf{0.25}$ & $\mathbf{0.25}$ & $ 0.45 $ & $\mathbf{0.25}$\\
Golden Statue Somewhere In Hong Kong & $ 18 $ & $ 39989 $ & $0.073 $ & $\mathbf{0.004}$ & $\mathbf{0.004}$ & $\mathbf{0.004}$ & $\mathbf{0.004}$ & $0.292 $ & $ 0.031 $ & $ 0.03 $ & $\mathbf{0.022}$ & $ 0.031 $ & $0.4 $ & $\mathbf{0.27}$ & $\mathbf{0.27}$ & $ 0.3 $ & $\mathbf{0.27}$\\
Gustav Vasa & $ 18 $ & $ 4249 $ & $1.085 $ & $ 1.145 $ & $ (0.101) $ & $\mathbf{0.099}$ & $ 0.1 $ & $34.181 $ & $ 32.266 $ & $ (0.751) $ & $ 0.839 $ & $ 0.841 $ & $3.52 $ & $ 3.15 $ & $ (0.31) $ & $ 0.48 $ & $ 0.48 $\\
GustavIIAdolf & $ 57 $ & $ 5813 $ & $9.714 $ & $ 8.524 $ & $\mathbf{0.004}$ & $\mathbf{0.004}$ & $\mathbf{0.004}$ & $67.784 $ & $ 58.458 $ & $\mathbf{0.021}$ & $\mathbf{0.021}$ & $\mathbf{0.021}$ & $ 13.91$ & $ 11.49 $ & $\mathbf{0.26}$ & $\mathbf{0.26}$ & $\mathbf{0.26}$\\
Jonas Ahlstromer & $ 40 $ & $ 2021 $ & $10.888 $ & $ 10.451 $ & $ (0.01) $ & $ 1.259 $ & $ 0.011 $ & $50.19 $ & $ 47.117 $ & $ (0.082) $ & $ 5.391 $ & $\mathbf{0.036}$ & $10.82 $ & $ 8.41 $ & $ (0.69) $ & $ 4.69 $ & $\mathbf{0.22}$\\
King's College University Of Toronto & $ 77 $ & $ 7087 $ & $0.235 $ & $ 0.017 $ & $ (0.005) $ & $ (1.877) $ & $ 0.017 $ & $0.989 $ & $ 0.085 $ & $ (0.059) $ & $ (4.624) $ & $ 0.084 $ & $0.9 $ & $\mathbf{0.34}$ & $ (0.35) $ & $ (7.12) $ & $\mathbf{0.34}$\\
Lund University Sphinx & $ 70 $ & $ 32668 $ & $4.585 $ & $ 2.191 $ & $ 0.016 $ & $ 1.512 $ & $\mathbf{0.009}$ & $19.522 $ & $ 8.752 $ & $ 0.058 $ & $ 5.452 $ & $\mathbf{0.033}$ & $4.78 $ & $ 1.36 $ & $ 0.4 $ & $ 4.58 $ & $\mathbf{0.39}$\\
Nijo Castle Gate & $ 19 $ & $ 7348 $ & $0.286 $ & $ 0.012 $ & $\mathbf{0.011}$ & $ 0.19 $ & $\mathbf{0.011}$ &  $1.495 $ & $ 0.069 $ & $\mathbf{0.064}$ & $ 0.744 $ & $\mathbf{0.064}$ & $1.7 $ & $\mathbf{0.73}$ & $\mathbf{0.73}$ & $ 4.84 $ & $\mathbf{0.73}$\\
Pantheon Paris & $ 179 $ & $ 29383 $ & $0.05 $ & $\mathbf{0.005}$ & $ 0.595 $ & $ 0.011 $ & - & $0.192 $ & $\mathbf{0.04}$ & $ 3.208 $ & $ 0.072 $ & - & $ 1.47 $ & $\mathbf{0.49}$ & $ 9.71 $ & $ 0.82 $ & -\\
Park Gate Clermont Ferrand & $ 34 $ & $ 9099 $ & $0.125 $ & $\mathbf{0.022}$ & $\mathbf{0.022}$ & $\mathbf{0.022}$ & $\mathbf{0.022}$ & $0.391 $ & $\mathbf{0.049}$ & $\mathbf{0.049}$ & $\mathbf{0.049}$ & $\mathbf{0.049}$ & $ 0.57$ & $\mathbf{0.35}$ & $\mathbf{0.35}$ & $\mathbf{0.35}$ & $\mathbf{0.35}$\\
Plaza De Armas Santiago & $ 240 $ & $ 26969 $ & $2.944 $ & $ 1.383 $ & $ 2.244 $ & - & $\mathbf{0.048}$ & $6.782 $ & $ 2.556 $ & $ 6.344 $ & - & $\mathbf{0.122}$ & $7.4 $ & $ 4.9 $ & $ 15.61 $ & - & $\mathbf{1.13}$\\
Porta San Donato Bologna & $ 141 $ & $ 25490 $ & $0.388 $ & $\mathbf{0.046}$ & $ 0.169 $ & $ 0.067 $ & $ 0.047 $ & $ 2.153$ & $\mathbf{0.095}$ & $ 0.513 $ & $ 0.149 $ & $ 0.099 $ & $2.28 $ & $\mathbf{0.75}$ & $ 3.23 $ & $ 1.16 $ & $\mathbf{0.75}$\\
Round Church Cambridge & $ 92 $ & $ 84643 $ & $1.003 $ & $ 0.582 $ & $ 0.493 $ & $\mathbf{0.012}$ & $\mathbf{0.012}$ & $2.451 $ & $ 1.107 $ & $ 1.851 $ & $\mathbf{0.033}$ & $ 0.035 $ & $2.66 $ & $ 1.54 $ & $ 2.03 $ & $ 0.41 $ & $\mathbf{0.39}$\\
Skansen Kronan Gothenburg & $ 131 $ & $ 28371 $ & $0.226 $ & $ 0.008 $ & $ 0.008 $ & $ (0.007) $ & $ 0.008 $ & $0.736 $ & $ 0.026 $ & $ 0.025 $ & $ (0.02) $ & $ 0.025 $ & $1.24 $ & $\mathbf{0.67}$ & $\mathbf{0.67}$ & $ (0.69) $ & $\mathbf{0.67}$\\
Smolny Cathedral St Petersburg & $ 131 $ & $ 51115 $ & $0.051 $ & $\mathbf{0.006}$ & $ 0.007 $ & - & $\mathbf{0.006}$ & $0.554 $ & $ 0.033 $ & $\mathbf{0.028}$ & - & $ 0.029 $ & $1.66 $ & $\mathbf{0.81}$ & $ 1.0 $ & - & $\mathbf{0.81}$\\
Some Cathedral In Barcelona & $ 177 $ & $ 30367 $ & $0.315 $ & $ 0.011 $ & $ 0.013 $ & $ 0.024 $ & $\mathbf{0.01}$ & $0.88 $ & $ 0.026 $ & $ 0.031 $ & $ 0.057 $ & $\mathbf{0.025}$ & $2.87 $ & $\mathbf{0.89}$ & $ 1.09 $ & $ 2.09 $ & $\mathbf{0.89}$\\
Sri Mariamman Singapore & $ 222 $ & $ 56220 $ & $0.683 $ & $\mathbf{0.023}$ & $ 0.614 $ & $ 0.025 $ & $\mathbf{0.023}$ & $2.302 $ & $\mathbf{0.077}$ & $ 2.158 $ & $ 0.083 $ & $ 0.078 $ & $4.13 $ & $ 0.91 $ & $ 7.4 $ & $ 1.17 $ & $\mathbf{0.89}$\\
Sri Thendayuthapani Singapore & $ 98 $ & $ 88849 $ & $3.812 $ & $ 2.87 $ & $ (0.053) $ & $\mathbf{0.034}$ & $\mathbf{0.034}$ & $ 46.269$ & $ 44.17 $ & $ (0.329) $ & $\mathbf{0.138}$ & $\mathbf{0.138}$ & $23.37 $ & $ 8.44 $ & $ (0.56) $ & $ 0.72 $ & $ 0.67 $\\
Sri Veeramakaliamman Singapore & $ 157 $ & $ 130013 $ & $0.597 $ & $ 0.04 $ & $ (1.388) $ & $ 0.095 $ & $\mathbf{0.038}$ & $2.559 $ & $ 0.175 $ & $ (3.41) $ & $ 0.288 $ & $\mathbf{0.169}$ & $3.47 $ & $ 0.73 $ & $ (34.72) $ & $ 2.2 $ & $\mathbf{0.71}$\\
Statue Of Liberty & $ 134 $ & $ 49250 $ & $20.012 $ & $ 4.122 $ & $ (4.782) $ & $ 28.049 $ & $\mathbf{0.099}$ & $46.887 $ & $ 9.091 $ & $ (8.281) $ & $ 2.945 $ & $\mathbf{0.213}$ & $26.16 $ & $ 6.97 $ & $ (52.05) $ & $ 5.08 $ & $\mathbf{1.25}$\\
The Pumpkin & $ 196 $ & $ 69341 $ & $14.89 $ & $ 14.952 $ & $\mathbf{0.022}$ & $ (14.862) $ & $\mathbf{0.022}$ & $94.672 $ & $ 98.862 $ & $ 0.092 $ & $ (3.123) $ & $\mathbf{0.091}$ & $33.41 $ & $ 24.85 $ & $\mathbf{0.57}$ & $ (24.19) $ & $\mathbf{0.57}$\\
Thian Hook Keng Temple Singapore & $ 138 $ & $ 34288 $ & $0.082 $ & $\mathbf{0.008}$ & $ 0.024 $ & $ 0.043 $ & $\mathbf{0.008}$ & $0.832 $ & $\mathbf{0.081}$ & $ 0.245 $ & $ 0.424 $ & $ 0.084 $ & $2.75 $ & $ 1.13 $ & $ 3.32 $ & $ 4.92 $ & $\mathbf{1.12}$\\
Tsar Nikolai I & $ 98 $ & $ 37857 $ & $9.467 $ & $ 7.836 $ & $\mathbf{0.005}$ & $\mathbf{0.005}$ & $\mathbf{0.005}$ & $48.499 $ & $ 36.28 $ & $\mathbf{0.018}$ & $\mathbf{0.018}$ & $\mathbf{0.018}$ & $9.79 $ & $ 6.53 $ & $\mathbf{0.33}$ & $\mathbf{0.33}$ & $\mathbf{0.33}$\\
Urban II & $ 96 $ & $ 22284 $ & $9.467 $ & $ 9.586 $ & $ 0.036 $ & $ 3.038 $ & $\mathbf{0.021}$ & $47.49 $ & $ 48.214 $ & $ 0.175 $ & $ 16.348 $ & $\mathbf{0.107}$ & $9.38 $ & $ 6.92 $ & $ 0.72 $ & $ 17.61 $ & $\mathbf{0.38}$\\
Vercingetorix & $ 69 $ & $ 10754 $ & $8.788 $ & $ 3.104 $ & $ 0.3 $ & $ 1.564 $ & $\mathbf{0.011}$ & $69.328 $ & $ 17.706 $ & $ 1.431 $ & $ 7.138 $ & $\mathbf{0.048}$ & $5.08 $ & $ 1.5 $ & $ 0.54 $ & $ 2.93 $ & $\mathbf{0.23}$\\
Yueh Hai Ching Temple Singapore & $ 43 $ & $ 13774 $ & $0.098 $ & $\mathbf{0.014}$ & $ (0.023) $ & $ 0.059 $ & $\mathbf{0.014}$ & $0.72 $ & $\mathbf{0.043}$ & $ (0.075) $ & $ 0.26 $ & $\mathbf{0.043}$ & $ 0.94$ & $\mathbf{0.65}$ & $ (1.64) $ & $ 2.06 $ & $\mathbf{0.65}$\\
    \bottomrule
	\end{tabular} 
	\end{adjustbox}
	\vspace{3pt}      
    \end{tabular}
    \caption{\small Single scene results with our method before and after bundle adjustment against baselines in the calibrated setup. The table shows mean camera location error (denoted $\tr_{\mathrm{error}}$) in meters, mean orientation error (denoted $R_{\mathrm{error}}$) in degrees, and mean reprojection error in pixels. (\textit{Smaller is better.}) In parenthesis experiments in which at least 10\% of the cameras are removed.}
    \label{tab:Euclidean_Results}
\end{table*}

We next present the results of our experiments. We show results in the single scene recovery and learning from multiple scenes setups. We then present an ablation study and a comparison to alternative novel neural architectures.

\subsection{Single scene recovery}

Tables~\ref{tab:Projective_Results} and~\ref{tab:Euclidean_Results} respectively show results in the uncalibrated and calibrated settings. In the majority of the cases our method achieves sub-pixel accuracies, on par with classical state of the art methods such as VarPro and Colmap. Notably, as we show in the supplementary material, already before BA our method often achieves sub-pixel accuracies, significantly surpassing GPSFM. We also note that unlike the baseline methods, which often remove a subset of the cameras (we marked experiments in which at least 10\% of the cameras are removed with parenthesis), our results are evaluated on all cameras. However, in a few cases our method did not converge to a favorable solution. We observe in these cases that the network reached a minimum in which a subset of the cameras are close to their ground truth positions, while the others appear to be displaced by a different global transformation. This could be resolved using sequential optimization, see supplementary material for more details and results.
We note that our comparison to VarPro is partial, since in a number of experiments it exceeded either memory or runtime limitations.

\subsection{Learning from multiple scenes}

\begin{table}[t]
    \hspace{-8pt}
    \setlength\tabcolsep{2pt} % default value: 6pt
    \tiny
    \centering
    \begin{tabular}{c}
        \begin{adjustbox}{max width=\textwidth}
        \aboverulesep=0ex
        \belowrulesep=0ex
        \renewcommand{\arraystretch}{1}
        \begin{tabular}[t]{|l|r|r|rrrr|}
            \hline
            \multirow{2}{3em}{Scan}& \multirow{2}{3em}{\#Images} & \multirow{2}{3em}{\#Points} & \multicolumn{4}{c|}{\textbf{Reprojection error (pixels)}} \\
            \cline{4-7}
            & & & Inference & Fine tuning & Opt. (short) & VarPro\\
            \hline
Alcatraz Water Tower & 172 & 14828 & $ 7.37 $ & $\mathbf{0.47}$ & $ 6.46 $ & $\mathbf{0.47}$\\
Dinosaur 319 & 36 & 319 & $ 1.58 $ & $ 1.30 $ & $ 1.71 $ & $\mathbf{0.43}$\\
Dinosaur 4983 & 36 & 4983 & $ 3.99 $ & $ 1.14 $ & $ 4.26 $ & $\mathbf{0.42}$\\
Eglise du dome & 85 & 84792 & $ 2.1 $ & $\mathbf{1.27}$ & $ 3.77 $ & -\\
Drinking Fountain Somewhere in Zurich & 14 & 5302 & $ 14.39 $ & $\mathbf{0.28}$ & $ 29.9 $ & $\mathbf{0.28}$\\
Gustav Vasa & 18 & 4249 & $ 6.3 $ & $\mathbf{0.16}$ & $\mathbf{0.16}$ & $\mathbf{0.16}$\\
Nijo Castle Gate & 19 & 7348 & $ 3.27 $ & $\mathbf{0.39}$ & $ 28.68 $ & $\mathbf{0.39}$\\
Skansen Kronan Gothenburg & 131 & 28371 & $ 1.64 $ & $\mathbf{0.41}$ & $ 0.52 $ & -\\
Some Cathedral in Barcelona & 177 & 30367 & $ 14.87 $ & $\mathbf{0.51}$ & $ 19.79 $ & -\\
Sri Veeramakaliamman Singapore & 157 & 130013 & $ 18.25 $ & $\mathbf{5.45}$ & $ 6.95 $ & -\\
			\bottomrule
	\end{tabular} 
	\end{adjustbox}
	\vspace{3pt}      
    \end{tabular}
    \caption{\small Reprojection errors obtained by applying our trained model to test data in the uncalibrated setup. Accuracies are shown after inference and fine tuning, compared to optimization with the same number of epochs starting with random initialization as well as results with VarPro. (\textit{Smaller is better.}) }
    \label{tab:Generalizing_proj_Results}
%\end{table}
%\begin{table}[t]
    \hspace{-8pt}
    \setlength\tabcolsep{2pt} % default value: 6pt
    \tiny
    \centering
    \begin{tabular}{c}
        \begin{adjustbox}{max width=\textwidth}
        \aboverulesep=0ex
        \belowrulesep=0ex
        \renewcommand{\arraystretch}{1}
        \begin{tabular}[t]{|l|r|r|rrrr|}
            \hline
            \multirow{2}{3em}{Scan}& \multirow{2}{3em}{\#Images} & \multirow{2}{3em}{\#Points} & \multicolumn{4}{c|}{\textbf{Reprojection error (pixels)}} \\
            \cline{4-7}
            & & & Inference & Fine tuning & Opt. (short) & Colmap\\
            \hline
Alcatraz Courtyard & 133 & 23674 & $ 0.82 $ & $\mathbf{0.81}$ & $ 1.61 $ & $\mathbf{0.81}$\\
Alcatraz Water Tower & 172 & 14828 & $\mathbf{0.55}$ & $\mathbf{0.55}$ & $ 4.6 $ & $\mathbf{0.55}$\\
Drinking Fountain Somewhere In Zurich & 14 & 5302 & $ 7.21 $ & $\mathbf{0.31}$ & $ 110.85 $ & $\mathbf{0.31}$\\
Nijo Castle Gate & 19 & 7348 & $ 5.81 $ & $\mathbf{0.73}$ & $ 19.0 $ & $\mathbf{0.73}$\\
Porta San Donato Bologna & 141 & 25490 & $ 1.1 $ & $ 0.79 $ & $ 83.02 $ & $\mathbf{0.75}$\\
Round Church Cambridge & 92 & 84643 & $ 0.5 $ & $ 1.51 $ & $ 28.17 $ & $\mathbf{0.39}$\\
Smolny Cathedral St Petersburg & 131 & 51115 & $ 15.15 $ & $\mathbf{0.81}$ & $ 2.86 $ & $\mathbf{0.81}$\\
Some Cathedral In Barcelona & 177 & 30367 & $ 21.46 $ & $\mathbf{0.89}$ & $ 1.06 $ & $\mathbf{0.89}$\\
Sri Veeramakaliamman Singapore & 157 & 130013 & $ 16.92 $ & $ 17.26 $ & $ 90.41 $ & $\mathbf{0.71}$\\
Yueh Hai Ching Temple Singapore & 43 & 13774 & $ 1.16 $ & $\mathbf{0.65}$ & $ 37.31 $ & $\mathbf{0.65}$\\
			\bottomrule
	\end{tabular} 
	\end{adjustbox}
	\vspace{3pt}      
    \end{tabular}
    \caption{\small Reprojection errors obtained by applying our trained model to test data in the calibrated setup. Accuracies are shown after inference and fine tuning, compared to optimization with the same number of epochs starting with random initialization as well as results with Colmap. (\textit{Smaller is better.})}
    \label{tab:Generalizing_euc_Results}
\end{table}

\begin{figure}[tb]
    \centering
    \includegraphics[width=0.45\linewidth]{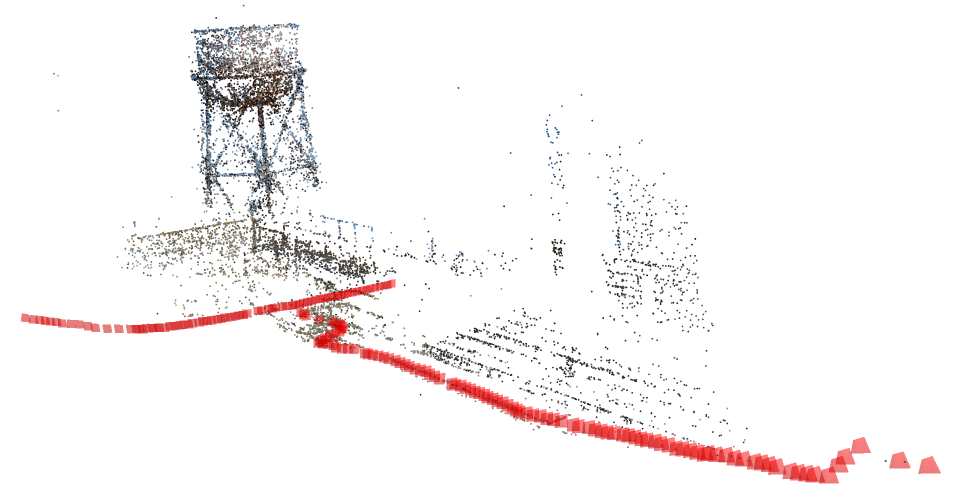}
    \includegraphics[width=0.45\linewidth]{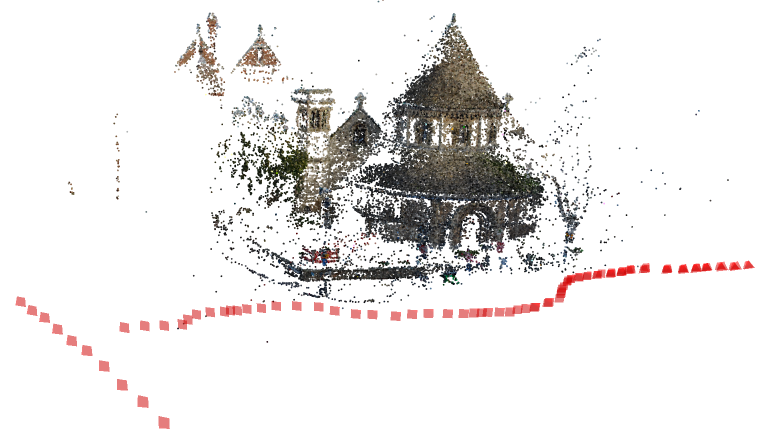}
    \caption{\small Recovery of 3D point clouds and camera poses in the calibrated setup obtained with our learning scheme, including inference, fine tuning and BA. 
    Left: Alcatraz Water Tower. Right: Round Church Cambridge. 
    %From left to right: Alcatraz Water Tower, Round Church Cambridge, and Some Cathedral In Barcelona. 
    Points are colored according to image intensity (darkened for better visibility) and cameras are colored in red. 
    }
    \label{fig:results}
\end{figure}

Tables~\ref{tab:Generalizing_proj_Results} and \ref{tab:Generalizing_euc_Results} respectively show results of testing with our trained model in both calibrated and uncalibrated settings. In each case we report reprojection error in pixels.  For our method we show results both after inference with our model and after additional 500 epochs fine tuning (+ BA in both cases). These are compared to 500 epochs single scene optimization (starting with random initialization) and to state of the art VarPro and Colmap. It can be seen that with the additional fine tuning in the majority of cases our network reached sub-pixel accuracy, often on par with state of the art methods. Figure~\ref{fig:results} shows several reconstruction results for test data.

We note that inference with our model typically takes only small fractions of a second and so the total runtime is dominated by the fine tuning and BA. For example, for the Some Cathedral In Barcelona dataset (with 177 cameras and $\sim$30K points) inference consumes only 0.007 seconds. 500 epoch fine tuning + BA take 263 seconds on a DGX machine, somewhat faster than Colmap (451s on an 8 thread, 2.2GHz CPUs), and significantly faster than our full, single scene optimization, which for this dataset with 70K epochs consumed nearly 3 hours. Run times for all tested datasets are reported in the supplementary material.

\subsection{Ablation study}

We have further conducted an ablation study. We tested our single scene method (a) as a simple elementwise feature encoder, i.e., with $W_2, W_3$ and $W_4$ removed from \eqref{eq:equilayer}, 
%(b) with no hinge loss, i.e., with all points penalized by the reprojection loss \eqref{eq:reproj}, (c)
(b) directly optimizing \eqref{eq:loss} with the Adam optimizer with cameras and 3D points treated as free variables (i.e., no network), 
%(d) directly optimizing \eqref{eq:reproj} with the Adam optimizer (i.e., no network and no hinge loss), and (e)
and (c) standard bundle adjustment with random initialization.
Table~\ref{tab:Projective_Ablation} shows results in the uncalibrated setup. The results emphasize the importance of the global features, produced by $W_2, W_3$ and $W_4$. Also, perhaps surprisingly, the Adam optimizer alone with our loss \eqref{eq:loss} achieves a significant improvement over standard bundle adjustment.

\begin{table}[t]
\centering
    \hspace{-8pt}
    \setlength\tabcolsep{2pt} % default value: 6pt
    \tiny
    \centering
    \begin{tabular}{c}
        \begin{adjustbox}{max width=\textwidth}
        \aboverulesep=0ex
        \belowrulesep=0ex
        \renewcommand{\arraystretch}{1}
        \begin{tabular}[t]{|l|rrrr|rr|}
            \hline
            \multirow{3}{3em}{Scan}& \multicolumn{6}{c|}{\textbf{Error (pixels)}}\\
            \cline{2-7}
            & \multirow{2}{3em}{\textbf{Ours}} & \multicolumn{1}{c}{\textbf{Elementwise}} & \multicolumn{1}{c}{\textbf{No}} & \multicolumn{1}{c|}{\textbf{BA}} & \multicolumn{1}{c}{\textbf{Set}} & \multicolumn{1}{c|}{\textbf{Graph}} \\
			&  & \multicolumn{1}{c}{\textbf{network}} & \multicolumn{1}{c}{\textbf{network}} & \multicolumn{1}{c|}{\textbf{only}} & \multicolumn{1}{c}{\textbf{network}} & \multicolumn{1}{c|}{\textbf{network}} \\
			\hline
			Alcatraz Courtyard & $\mathbf{0.52}$ & $\mathbf{0.52}$ & $ 1.31 $ & $ 191.42 $ & $\mathbf{0.52}$ & $\mathbf{0.52}$\\
The Pumpkin & $\mathbf{0.38}$ & $ 6.82 $ & $ 18.67 $ & $ 324.2 $ & $\mathbf{0.38}$ & $\mathbf{0.38}$\\
Alcatraz Water Tower & $\mathbf{0.47}$ & $\mathbf{0.47}$ & $ 1.73 $ & $ 54.03 $ & $ 1.56 $ & $ 0.68 $\\
Folke Filbyter & $\mathbf{8.58}$ & $ 22.8 $ & $ 35.41 $ & $ 325.62 $ & $ 30.19 $ & $ 92.24 $\\
Gustav Vasa & $\mathbf{0.16}$ & $\mathbf{0.16}$ & $ 9.06 $ & $ 34.47 $ & $ 69.11 $ & $\mathbf{0.16}$\\
Dinosaur 319 & $ 1.53 $ & $ 1.46 $ & $ 1.29 $ & $ 13.43 $ & $\mathbf{0.59}$ & -\\
Park Gate Clermont Ferrand & $\mathbf{0.31}$ & $\mathbf{0.31}$ & $\mathbf{0.31}$ & $ 60.13 $ & $\mathbf{0.31}$ & $\mathbf{0.31}$\\
Skansen Kronan Gothenburg & $\mathbf{0.41}$ & $ 1.04 $ & $\mathbf{0.41}$ & $ 236.75 $ & $\mathbf{0.41}$ & $ 2.08 $\\
Smolny Cathedral St Petersburg & $\mathbf{0.46}$ & $\mathbf{0.46}$ & $ 3.9 $ & $ 179.58 $ & $\mathbf{0.46}$ & $\mathbf{0.46}$\\
Sri Thendayuthapani Singapore & $\mathbf{0.31}$ & $ 8.85 $ & $ 0.42 $ & $ 222.95 $ & $\mathbf{0.31}$ & $\mathbf{0.31}$\\
King's College University Of Toronto & $ 0.78 $ & $ 0.86 $ & $ 17.02 $ & $ 36.8 $ & $ 1.07 $ & $\mathbf{0.63}$\\
Model House & $\mathbf{0.34}$ & $\mathbf{0.34}$ & $ 0.92 $ & $ 26.23 $ & $ 0.52 $ & $\mathbf{0.34}$\\
Ecole Superior De Guerre & $\mathbf{0.26}$ & $ 1.83 $ & $\mathbf{0.26}$ & $ 134.41 $ & $\mathbf{0.26}$ & $\mathbf{0.26}$\\
Golden Statue Somewhere In Hong Kong & $\mathbf{0.22}$ & $ 0.86 $ & $ 0.89 $ & $ 24.81 $ & $\mathbf{0.22}$ & $\mathbf{0.22}$\\
Dinosaur 4983 & $\mathbf{0.57}$ & $ 0.92 $ & $ 1.08 $ & $ 16.86 $ & $ 0.87 $ & $ 0.84 $\\
Tsar Nikolai I & $\mathbf{2.43}$ & $ 5.88 $ & $ 22.09 $ & $ 99.73 $ & $ 7.4 $ & $ 2.47 $\\
Buddah Tooth Relic Temple Singapore & $\mathbf{0.6}$ & $ 3.69 $ & $ 4.68 $ & $ 256.71 $ & $\mathbf{0.6}$ & $\mathbf{0.6}$\\
Drinking Fountain Somewhere In Zurich & $\mathbf{0.28}$ & $\mathbf{0.28}$ & $ 1.41 $ & $ 35.01 $ & $\mathbf{0.28}$ & $\mathbf{0.28}$\\
Jonas Ahlstromer & $ 4.72 $ & $\mathbf{3.8}$ & $ 19.8 $ & $ 72.63 $ & $ 4.31 $ & $ 5.16 $\\
Porta San Donato Bologna & $\mathbf{0.4}$ & $ 1.92 $ & $ 19.22 $ & $ 135.2 $ & $\mathbf{0.4}$ & $\mathbf{0.4}$\\
Nijo Castle Gate & $\mathbf{0.39}$ & $\mathbf{0.39}$ & $ 0.57 $ & $ 114.39 $ & $\mathbf{0.39}$ & $\mathbf{0.39}$\\
Corridor & $\mathbf{0.26}$ & $\mathbf{0.26}$ & $ 0.39 $ & $ 7.99 $ & $\mathbf{0.26}$ & $\mathbf{0.26}$\\
Eglise du dome & $\mathbf{0.24}$ & $ 1.2 $ & $ 0.9 $ & $ 61.92 $ & $\mathbf{0.24}$ & $\mathbf{0.24}$\\
Doge Palace Venice & $\mathbf{0.6}$ & $ 5.36 $ & $ 20.33 $ & $ 378.06 $ & $\mathbf{0.6}$ & $ 0.73 $\\
East Indiaman Goteborg & $\mathbf{0.99}$ & $ 2.33 $ & $ 21.79 $ & $ 67.96 $ & $ 2.6 $ & $ 1.13 $\\
GustavIIAdolf & $ 5.83 $ & $ 5.51 $ & $\mathbf{0.23}$ & $ 74.63 $ & $ 5.32 $ & $ 5.28 $\\
Lund University Sphinx & $\mathbf{0.34}$ & $ 1.74 $ & $ 12.24 $ & $ 77.69 $ & $ 0.77 $ & $ 4.62 $\\
Pantheon Paris & $\mathbf{0.49}$ & $ 0.53 $ & $ 31.08 $ & $ 152.99 $ & $\mathbf{0.49}$ & $\mathbf{0.49}$\\
Plaza De Armas Santiago & $\mathbf{0.64}$ & $ 2.36 $ & $ 2.41 $ & $ 212.0 $ & $\mathbf{0.64}$ & $ 3.55 $\\
Some Cathedral In Barcelona & $\mathbf{0.51}$ & $ 0.54 $ & $ 8.13 $ & $ 265.96 $ & $ 2.27 $ & $\mathbf{0.51}$\\
Sri Mariamman Singapore & $\mathbf{0.61}$ & $ 6.83 $ & $ 35.09 $ & $ 274.74 $ & $\mathbf{0.61}$ & $ 0.62 $\\
Sri Veeramakaliamman Singapore & $\mathbf{0.52}$ & $ 2.04 $ & $ 17.39 $ & $ 445.98 $ & $ 3.06 $ & $ 2.66 $\\
Thian Hook Keng Temple Singapore & $\mathbf{0.54}$ & $ 0.83 $ & $ 7.45 $ & $ 405.92 $ & $\mathbf{0.54}$ & $\mathbf{0.54}$\\
Urban II & $ 6.84 $ & $\mathbf{6.39}$ & $ 22.95 $ & $ 117.47 $ & $ 31.64 $ & $ 141.92 $\\
Alcatraz West Side Gardens & $\mathbf{0.76}$ & $ 7.54 $ & $ 8.05 $ & $ 303.1 $ & $ 0.86 $ & $ 5.81 $\\
Skansen Lejonet Gothenburg & $\mathbf{2.05}$ & $ 14.5 $ & $ 18.54 $ & $ 268.78 $ & $ 2.32 $ & $ 3.33 $\\
Basilica Di San Petronio & $ 0.96 $ & $ 2.99 $ & $ 27.87 $ & $ 241.3 $ & $ 1.26 $ & $\mathbf{0.85}$\\
Buddah Statue & $\mathbf{2.93}$ & $ 7.85 $ & $ 11.18 $ & $ 186.99 $ & $ 5.15 $ & $ 6.31 $\\
%Golden Statue somewhere in Hong Kong -> Golden Statue in Hong Kong
%Buddah Tooth Relic Temple Singapore - > Buddah Tooth Relic Temple Sing.
%Drinking Fountain somewhere in Zurich -> Drinking Fountain in Zurich
%Thian Hook Keng Temple Singapore -> Thian Hook Keng Temple Sing. 
            \hline
            \textbf{Geo. Mean Ratio} & $ 1.0 $ & $ 2.35 $ & $ 5.95 $ & $ 150.8 $ & $ 1.36 $ & $ 1.52 $\\

			\bottomrule
	\end{tabular} 
	\end{adjustbox}
	\vspace{3pt}      
    \end{tabular}
    \caption{\small Ablation Study, uncalibrated setting: reprojection errors with different components removed (see text) and with alternative network models. The last row shows for each condition the geometric mean of the ratios of reprojection errors and the corresponding errors of the full method. ({\textit{Smaller is better.}})}
    \label{tab:Projective_Ablation}
    %\vspace*{-0.5cm}
\end{table}

\subsection{Comparison to alternative deep architectures} \label{sec:alternative}

We have further compared our method to two novel deep architectures for SFM. The first architecture uses a set neural network \cite{zaheer2017deep} akin to our camera head $H_\mathrm{cams}$ to which we add a global feature. This network optimizes the same loss~\eqref{eq:loss} with the 3D point locations treated as free variables. This network can be seen as a simpler version of our method. A second method further adds pairwise relations between the cameras, inducing a \emph{viewing graph}. The input to this network further includes fundamental matrices, which are computed from the point tracks. We use a standard graph neural network \cite{gilmer2017neural} to optimize the loss~\eqref{eq:loss}. More details about these algorithms are provided in the supplementary material.

Results with these models are shown in the right two columns of Table \ref{tab:Projective_Ablation}. Both methods were inferior to our proposed approach, with the set network performing somewhat better than the graph network. Interestingly, simplifying our method by treating the 3D points as free variable hurts performance, and injecting the fundamental matrices to the network does not improve performance. We finally note in addition that both of these methods do not support learning.

\section{Conclusion}

We have introduced a novel deep-based method for multiview SFM, in both the calibrated and unclibrated settings.  Starting with a (sparse) measurement tensor of point tracks, our method minimizes the reprojection loss to yield a simultaneous recovery of both camera parameters and a sparse 3D reconstruction. Importantly, we use an equivariant network architecture that respects the symmetries of the task, i.e., equivariant to permutation of either the rows (cameras) or the columns (3D points) of the measurement tensor.  We have tested our method in two setups, single-scene optimization and inference with a trained model. Our experiments indicate that our method can achieve accurate pose and structure recovery, on par with classical, state-of-the-art techniques. In future work, we plan to extend this work to allow direct, end-to-end recovery from raw images and to produce dense 3D reconstruction. 

\section*{Acknowledgement}
This research was partially supported by the Israeli Council for Higher Education (CHE) via the Weizmann Data Science Research Center, by the Dan and Betty Kahn Michigan-Israel Partnership, and by the U.S.-Israel Binational Science Foundation, grant number 2018680.

{\small
\bibliographystyle{LaTeX/ieee_fullname}
\bibliography{arxiv.bib}
}

\onecolumn
\appendix
\section*{Appendix}

Below we provide implementation details for the baseline methods and the alternative deep architectures tested in our paper. We further include additional results, including examples of reconstructions with our method.

\section{Baselines}
\paragraph{Colmap baseline.}
In the calibrated experiments (Table 2 in the paper), for fair comparison, we applied Colmap \cite{schoenberger2016sfm} directly to the points tracks provided by Olsson’s dataset \cite{olsson2011stable} and  fixed the intrinsic camera parameters to those provided as ground truth.

\paragraph{Linear baseline.}
We tested Jiang et al.'s method \cite{jiang2013global} while ignoring viewing graph edges for which the number of matching points was lower than a certain threshold. We used thresholds of 30, 200, 500 matching points and report those results for which the lowest reprojection error, before bundle adjustment, was obtained.

\section{Alternative deep architectures}

\noindent
The two right most columns in Table 5 in the paper show results of two novel deep architectures which were developed for comparison to our deep network architecture. The details are given below. 

\paragraph{Set neural network.}
For a scene with $m$ cameras, the input to this network is a set of $m$ random feature vectors of size 12 that provide unique ids to each camera. Inspired by \cite{qi2017pointnet,zaheer2017deep}, our set network is composed of three sub-networks where each sub-network is an equivariant set network. The first sub-network is applied to each feature vector and calculates a local feature for each camera. The second sub-network is applied to each such local feature. The outputs for all cameras are then averaged, producing a global scene feature vector. Finally, the camera parameters are predicted by applying the third sub-network to both the local and global feature. In summary, the set network prediction for camera $i$ is defined as follows
\begin{eqnarray*}
    \z_i &=& S_1(\vv_i) \\
    \z_g &=& \frac{1}{m} \sum_{i=1}^{m}{S_2(\z_i)} \\
    P_i &=& S_3(\z_i, \z_g).
\end{eqnarray*}
Each $S_k$ is a fully connected network and $v_i$ is the initial random vector of camera $i$.

\paragraph{Graph neural network.}
Here, the cameras are represented by the nodes of a graph, called the \emph{viewing graph}. An edge connects a pair of nodes if the respective images share at least 30 tracks, in which case a fundamental matrix is computed. The fundamental matrices are used as edge input features, while as with the set network model, random vectors form the node input features. We use a message-passing scheme \cite{gilmer2017neural} and global feature as described for the set network model. Each message-passing layer is of the following form 
\begin{equation*}
    \z_i^{l} = \frac{1}{|N_i|}\sum_{j\in N_i}{\mathrm {mlp }}_l(\z_i^{l-1},\z_j^{l-1},F_{ij})
\end{equation*}

where $\z_i^l$ is the local feature of node $i$ in layer $l$, $N_i$ are the neighbors of node $i$ and $F_{ij}$ is the fundamental matrix measured between cameras $i$ and $j$.

Both the set and the graph models predict camera parameters, while the 3D points are treated as free variables. In both cases we minimize the reprojection loss defined in equation (3) in the paper.

\section{Results}

\begin{table*}[t]
    \hspace{-8pt}
    \setlength\tabcolsep{2pt} % default value: 6pt
    %\small
    \centering
    \scriptsize
    \begin{tabular}{c}
        \begin{adjustbox}{max width=\textwidth}
        \aboverulesep=0ex
        \belowrulesep=0ex
        \renewcommand{\arraystretch}{1}
        \begin{tabular}[t]{|l|r|r|rr|rrrr|}
            \hline
            \multirow{3}{3em}{Scan}& \multirow{3}{3em}{\#Images} & \multirow{3}{3em}{\#Points} & \multicolumn{6}{c|}{\textbf{Error (pixels)}}\\ \cline{4-9}
            & & & \multicolumn{2}{c|}{Before BA} & \multicolumn{4}{c|}{After BA}\\
            \cline{4-9}
            & & & \textbf{Ours} & GPSFM & \textbf{Ours} & GPSFM & PPSFM & VarPro \\
			\hline
Alcatraz Courtyard & $ 133 $ & $ 23674 $ & $\mathbf{1.55}$ & $ 20.34 $ & $\mathbf{0.52}$ & $\mathbf{0.52}$ & $ 0.57 $ & $\mathbf{0.52}$\\
Alcatraz Water Tower & $ 172 $ & $ 14828 $ & $\mathbf{2.18}$ & $ 16.5 $ & $\mathbf{0.47}$ & $ 0.63 $ & $ 0.59 $ & $\mathbf{0.47}$\\
Alcatraz West Side Gardens & $ 419 $ & $ 65072 $ & $\mathbf{9.54}$ & $ 1007.5 $ & $\mathbf{0.76}$ & $ 326.99 $ & $ 1.77 $ & -\\
Basilica Di San Petronio & $ 334 $ & $ 46035 $ & $\mathbf{7.9}$ & $ 1871.41 $ & $ 0.96 $ & $ 60.69 $ & $\mathbf{0.63}$ & -\\
Buddah Statue & $ 322 $ & $ 156356 $ & $\mathbf{18.88}$ & $ 919.26 $ & $ 2.93 $ & $ 96.96 $ & $\mathbf{0.41}$ & -\\
Buddah Tooth Relic Temple Singapore & $ 162 $ & $ 27920 $ & $\mathbf{4.59}$ & $ 18.53 $ & $\mathbf{0.6}$ & $ 0.62 $ & $ 0.71 $ & $\mathbf{0.6}$\\
Corridor & $ 11 $ & $ 737 $ & $\mathbf{0.3}$ & $ 0.64 $ & $\mathbf{0.26}$ & $\mathbf{0.26}$ & $ 0.27 $ & $\mathbf{0.26}$\\
Ecole Superior De Guerre & $ 35 $ & $ 13477 $ & $\mathbf{0.75}$ & $ 1.88 $ & $\mathbf{0.26}$ & $\mathbf{0.26}$ & $ 0.28 $ & $\mathbf{0.26}$\\
Dinosaur 319 & $ 36 $ & $ 319 $ & $\mathbf{2.35}$ & $ 4.66 $ & $ 1.53 $ & $\mathbf{0.43}$ & $ 0.47 $ & $\mathbf{0.43}$\\
Dinosaur 4983 & $ 36 $ & $ 4983 $ & $ 1.96 $ & $\mathbf{1.54}$ & $ 0.57 $ & $\mathbf{0.42}$ & $ 0.47 $ & $\mathbf{0.42}$\\
Doge Palace Venice & $ 241 $ & $ 67107 $ & $\mathbf{3.6}$ & $ 170.93 $ & $\mathbf{0.6}$ & $ 3.52 $ & $ 0.67 $ & -\\
Eglise du dome & $ 85 $ & $ 84792 $ & $\mathbf{1.1}$ & $ 8.41 $ & $\mathbf{0.24}$ & $\mathbf{0.24}$ & $ 0.25 $ & -\\
Drinking Fountain Somewhere In Zurich & $ 14 $ & $ 5302 $ & $\mathbf{0.33}$ & $ 1.29 $ & $\mathbf{0.28}$ & $\mathbf{0.28}$ & $ 0.31 $ & $\mathbf{0.28}$\\
East Indiaman Goteborg & $ 179 $ & $ 25655 $ & $\mathbf{3.31}$ & $ 99.38 $ & $ 0.99 $ & $ 5.11 $ & $\mathbf{0.67}$ & -\\
Folke Filbyter & $ 40 $ & $ 21150 $ & $ 8.87 $ & $\mathbf{1.78}$ & $ 8.58 $ & $ 0.82 $ & $\mathbf{0.33}$ & $ 277.89 $\\
Golden Statue Somewhere In Hong Kong & $ 18 $ & $ 39989 $ & $\mathbf{0.35}$ & $ 0.81 $ & $\mathbf{0.22}$ & $\mathbf{0.22}$ & $ 0.24 $ & $\mathbf{0.22}$\\
Gustav Vasa & $ 18 $ & $ 4249 $ & $\mathbf{0.23}$ & $ 1.82 $ & $\mathbf{0.16}$ & $\mathbf{0.16}$ & $ 0.17 $ & $\mathbf{0.16}$\\
GustavIIAdolf & $ 57 $ & $ 5813 $ & $ 14.77 $ & $\mathbf{5.91}$ & $ 5.83 $ & $\mathbf{0.23}$ & $ 0.24 $ & $\mathbf{0.23}$\\
Model House & $ 10 $ & $ 672 $ & $\mathbf{0.37}$ & $ 3.66 $ & $\mathbf{0.34}$ & $ 1.12 $ & $ 0.4 $ & $\mathbf{0.34}$\\
Jonas Ahlstromer & $ 40 $ & $ 2021 $ & $\mathbf{14.38}$ & $ 28.83 $ & $ 4.72 $ & $\mathbf{0.18}$ & $ 0.2 $ & $\mathbf{0.18}$\\
Lund University Sphinx & $ 70 $ & $ 32668 $ & $\mathbf{3.64}$ & $ 10.0 $ & $\mathbf{0.34}$ & $ 0.45 $ & $ 0.37 $ & $\mathbf{0.34}$\\
Nijo Castle Gate & $ 19 $ & $ 7348 $ & $\mathbf{0.71}$ & $ 20.08 $ & $\mathbf{0.39}$ & $\mathbf{0.39}$ & $ 0.43 $ & $\mathbf{0.39}$\\
Pantheon Paris & $ 179 $ & $ 29383 $ & $\mathbf{1.75}$ & $ 44.85 $ & $\mathbf{0.49}$ & $ 2.85 $ & $ 0.62 $ & -\\
Park Gate Clermont Ferrand & $ 34 $ & $ 9099 $ & $\mathbf{0.61}$ & $ 13.82 $ & $\mathbf{0.31}$ & $ 0.32 $ & $ 0.49 $ & $\mathbf{0.31}$\\
Plaza De Armas Santiago & $ 240 $ & $ 26969 $ & $\mathbf{5.1}$ & $ 81.01 $ & $\mathbf{0.64}$ & $ 3.14 $ & $ 0.71 $ & -\\
Porta San Donato Bologna & $ 141 $ & $ 25490 $ & $\mathbf{1.58}$ & $ 33.36 $ & $\mathbf{0.4}$ & $ 0.61 $ & $ 3.75 $ & $\mathbf{0.4}$\\
The Pumpkin & $ 195 $ & $ 69335 $ & $ 14.45 $ & $\mathbf{8.97}$ & $\mathbf{0.38}$ & $\mathbf{0.38}$ & $ 0.42 $ & -\\
Skansen Kronan Gothenburg & $ 131 $ & $ 28371 $ & $\mathbf{1.19}$ & $ 8.9 $ & $\mathbf{0.41}$ & $ 0.44 $ & $ 0.44 $ & -\\
Skansen Lejonet Gothenburg & $ 368 $ & $ 74423 $ & $\mathbf{10.82}$ & $ 69.81 $ & $ 2.05 $ & $ 7.48 $ & $\mathbf{1.28}$ & -\\
Smolny Cathedral St Petersburg & $ 131 $ & $ 51115 $ & $\mathbf{1.66}$ & $ 83.78 $ & $\mathbf{0.46}$ & $\mathbf{0.46}$ & $ 0.5 $ & -\\
Some Cathedral In Barcelona & $ 177 $ & $ 30367 $ & $\mathbf{3.67}$ & $ 14.77 $ & $\mathbf{0.51}$ & $\mathbf{0.51}$ & $ 0.54 $ & -\\
Sri Mariamman Singapore & $ 222 $ & $ 56220 $ & $\mathbf{7.06}$ & $ 39.89 $ & $\mathbf{0.61}$ & $ 0.78 $ & $ 0.85 $ & -\\
Sri Thendayuthapani Singapore & $ 98 $ & $ 88849 $ & $\mathbf{2.12}$ & $ 13.25 $ & $\mathbf{0.31}$ & $ 0.56 $ & $ 0.33 $ & -\\
Sri Veeramakaliamman Singapore & $ 157 $ & $ 130013 $ & $\mathbf{6.47}$ & $ 99.99 $ & $\mathbf{0.52}$ & $ 1.78 $ & $ 0.66 $ & -\\
Thian Hook Keng Temple Singapore & $ 138 $ & $ 34288 $ & $\mathbf{7.59}$ & $ 26.78 $ & $\mathbf{0.54}$ & $ 0.55 $ & $ 0.66 $ & $\mathbf{0.54}$\\
King's College University Of Toronto & $ 77 $ & $ 7087 $ & $\mathbf{2.27}$ & $ 22.89 $ & $ 0.78 $ & $ 2.35 $ & $ 0.26 $ & $\mathbf{0.24}$\\
Tsar Nikolai I & $ 98 $ & $ 37857 $ & $\mathbf{6.04}$ & $ 13.21 $ & $ 2.43 $ & $ 0.33 $ & $ 0.31 $ & $\mathbf{0.29}$\\
Urban II & $ 96 $ & $ 22284 $ & $\mathbf{16.91}$ & $ 87.25 $ & $ 6.84 $ & $\mathbf{0.27}$ & $ 0.31 $ & $ 3.61 $\\
			\bottomrule
	\end{tabular} 
	\end{adjustbox}
	\vspace{3pt}      
    \end{tabular}
    \caption{\small Single scene experiments in the uncalibrated setup. The table shows mean reprojection errors obtained with our method before and after BA, compared to GPSFM \cite{kasten2019gpsfm}, PPSFM \cite{magerand2017practical} and VarPro \cite{hong2016projective}. ({\textit{Smaller is better.}}) Our comparison to VarPro is partial, since in a number of experiments it exceeded either memory or runtime limitations.}
    \label{tab:Projective_Results_No_BA}
\end{table*}
\begin{table*}[tb]
    \hspace{-8pt}
    \setlength\tabcolsep{1pt} % default value: 6pt
    \tiny
    \begin{tabular}{c}
        \begin{adjustbox}{max width=\textwidth}
        \aboverulesep=0ex
        \belowrulesep=0ex
        \renewcommand{\arraystretch}{0.9}
        \begin{tabular}[t]{|l|r|r||rrr|rrr|rrr||rrrr|rrrr|rrrr|}
            \hline
            \multirow{3}{3em}{Scan}& \multirow{3}{3em}{\#Images} & \multirow{3}{3em}{\#Points} & \multicolumn{9}{c||}{\textbf{Before BA}} & \multicolumn{12}{c|}{\textbf{After BA}}\\
            \cline{4-24}
            & & & \multicolumn{3}{c|}{$\tr_{\text{error}}$} & \multicolumn{3}{|c|}{$R_{\text{error}}$} & \multicolumn{3}{|c||}{Reprojection Err.} & \multicolumn{4}{c|}{$\tr_{\text{error}}$} & \multicolumn{4}{c|}{$R_{\text{error}}$} & \multicolumn{4}{c|}{Reprojection Err.} \\
            & & & \textbf{Ours} & GESFM & Linear & \textbf{Ours} & GESFM & Linear & \textbf{Ours} & GESFM & Linear & \textbf{Ours} & GESFM & Linear & Colmap & \textbf{Ours} & GESFM & Linear & Colmap & \textbf{Ours} & GESFM & Linear & Colmap  \\
			\hline
Alcatraz Courtyard & $ 133 $ & $ 23674 $ & $\mathbf{0.16}$ & $ 0.767 $ & $ 0.378 $ & $\mathbf{0.619}$ & $ 1.851 $ & $ 0.729 $ & $\mathbf{1.64}$ & $ 66.5 $ & $ 16.58 $ & $ 0.015 $ & $ 0.259 $ & $\mathbf{0.014}$ & $\mathbf{0.014}$ & $ 0.049 $ & $ 0.533 $ & $\mathbf{0.042}$ & $ 0.043 $ & $\mathbf{0.81}$ & $ 4.67 $ & $ 1.27 $ & $\mathbf{0.81}$\\
Alcatraz Water Tower & $ 172 $ & $ 14828 $ & $\mathbf{0.518}$ & $ 8.332 $ & $ 1.643 $ & $\mathbf{0.933}$ & $ 1.136 $ & $ 1.525 $ & $\mathbf{2.13}$ & $ 131.81 $ & $ 56.26 $ & $ 0.116 $ & $ 9.147 $ & $ 1.643 $ & $\mathbf{0.115}$ & $ 0.23 $ & $ 9.997 $ & $ 1.525 $ & $\mathbf{0.228}$ & $\mathbf{0.55}$ & $ 25.93 $ & $ 73.72 $ & $\mathbf{0.55}$\\
Buddah Tooth Relic Temple Singapore & $ 162 $ & $ 27920 $ & $\mathbf{0.233}$ & $ 2.124 $ & $ 1.325 $ & $\mathbf{1.03}$ & $ 2.95 $ & $ 2.058 $ & $\mathbf{2.06}$ & $ 89.94 $ & $ 47.5 $ & $\mathbf{0.014}$ & $ 1.429 $ & $ 0.125 $ & $ 0.015 $ & $\mathbf{0.081}$ & $ 4.709 $ & $ 0.551 $ & $ 0.083 $ & $\mathbf{0.85}$ & $ 13.22 $ & $ 2.66 $ & $\mathbf{0.85}$\\
Doge Palace Venice & $ 241 $ & $ 67107 $ & $\mathbf{0.342}$ & $ 1.688 $ & - & $\mathbf{1.163}$ & $ 2.75 $ & - & $\mathbf{3.62}$ & $ 123.53 $ & - & $ 0.029 $ & $ 1.608 $ & - & $\mathbf{0.012}$ & $ 0.211 $ & $ 5.317 $ & - & $\mathbf{0.031}$ & $ 1.0 $ & $ 22.32 $ & - & $\mathbf{0.98}$\\
Door Lund & $ 12 $ & $ 17650 $ & $\mathbf{0.006}$ & $ (1.603) $ & $ 0.226 $ & $\mathbf{0.024}$ & $ (2.041) $ & $ 1.148 $ & $\mathbf{0.32}$ & $ (227.0) $ & $ 20.89 $ & $\mathbf{0.001}$ & $ (0.973) $ & $\mathbf{0.001}$ & $\mathbf{0.001}$ & $ 0.006 $ & $ (7.552) $ & $\mathbf{0.005}$ & $\mathbf{0.005}$ & $\mathbf{0.3}$ & $ (9.21) $ & $\mathbf{0.3}$ & $\mathbf{0.3}$\\
Drinking Fountain Somewhere In Zurich & $ 14 $ & $ 5302 $ & $\mathbf{0.004}$ & $ (0.016) $ & $ 0.024 $ & $\mathbf{0.031}$ & $ (0.054) $ & $ 0.077 $ & $\mathbf{0.33}$ & $ (0.94) $ & $ 0.58 $ & $\mathbf{0.002}$ & $ (0.002) $ & $\mathbf{0.002}$ & $\mathbf{0.002}$ & $\mathbf{0.007}$ & $ (0.01) $ & $\mathbf{0.007}$ & $\mathbf{0.007}$ & $ 0.31 $ & $ (0.27) $ & $ 0.31 $ & $ 0.31 $\\
East Indiaman Goteborg & $ 179 $ & $ 25655 $ & $\mathbf{0.621}$ & $ 2.783 $ & $ (2.235) $ & $ 3.814 $ & $ 11.129 $ & $ (3.284) $ & $\mathbf{4.13}$ & $ 170.63 $ & $ (94.46) $ & $ 0.509 $ & $ 3.099 $ & $ (2.235) $ & $\mathbf{0.065}$ & $ 3.117 $ & $ 12.396 $ & $ (3.284) $ & $\mathbf{0.251}$ & $ 1.85 $ & $ 32.37 $ & $ (312.9) $ & $\mathbf{0.89}$\\
Ecole Superior De Guerre & $ 35 $ & $ 13477 $ & $ 0.081 $ & $ (0.006) $ & $ 0.048 $ & $ 0.318 $ & $ (0.057) $ & $ 0.182 $ & $ 0.72 $ & $ (0.35) $ & $ 1.48 $ & $ 0.005 $ & $ (0.002) $ & $ 0.005 $ & $ 0.005 $ & $\mathbf{0.024}$ & $ (0.035) $ & $\mathbf{0.024}$ & $\mathbf{0.024}$ & $ 0.34 $ & $ (0.14) $ & $ 0.34 $ & $ 0.34 $\\
Eglise du dome & $ 85 $ & $ 84792 $ & $ 0.205 $ & $ (1.958) $ & $\mathbf{0.128}$ & $\mathbf{0.808}$ & $ (2.851) $ & $ 0.903 $ & $\mathbf{0.91}$ & $ (90.83) $ & $ 26.4 $ & $\mathbf{0.01}$ & $ (1.425) $ & $ 0.046 $ & $\mathbf{0.01}$ & $ 0.037 $ & $ (3.631) $ & $ 0.162 $ & $\mathbf{0.036}$ & $\mathbf{0.27}$ & $ (6.21) $ & $ 0.76 $ & $\mathbf{0.27}$\\
Folke Filbyter & $ 40 $ & $ 21150 $ & $ 0.125 $ & $ (0.003) $ & $ 0.021 $ & $ 74.596 $ & $ (0.332) $ & $ 1.94 $ & $ 10.37 $ & $ (5.74) $ & $ 72.06 $ & $ 0.118 $ & $ (0.0) $ & $ 0.123 $ & $\mathbf{0.0}$ & $ 70.157 $ & $ (0.148) $ & $ 4.484 $ & $\mathbf{0.036}$ & $ 4.29 $ & $ (0.41) $ & $ 6.06 $ & $\mathbf{0.29}$\\
Fort Channing Gate Singapore & $ 27 $ & $ 23627 $ & $ 0.093 $ & $\mathbf{0.092}$ & $ 0.139 $ & $\mathbf{0.207}$ & $ 0.295 $ & $ 0.659 $ & $\mathbf{0.52}$ & $ 2.57 $ & $ 22.69 $ & $\mathbf{0.008}$ & $\mathbf{0.008}$ & $ 0.013 $ & $\mathbf{0.008}$ & $\mathbf{0.02}$ & $\mathbf{0.02}$ & $ 0.029 $ & $\mathbf{0.02}$ & $\mathbf{0.25}$ & $\mathbf{0.25}$ & $ 0.45 $ & $\mathbf{0.25}$\\
Golden Statue Somewhere In Hong Kong & $ 18 $ & $ 39989 $ & $\mathbf{0.073}$ & $ 0.118 $ & $ 1.153 $ & $\mathbf{0.292}$ & $ 0.669 $ & $ 8.264 $ & $\mathbf{0.4}$ & $ 4.98 $ & $ 73.7 $ & $\mathbf{0.004}$ & $\mathbf{0.004}$ & $\mathbf{0.004}$ & $\mathbf{0.004}$ & $ 0.031 $ & $ 0.03 $ & $\mathbf{0.022}$ & $ 0.031 $ & $\mathbf{0.27}$ & $\mathbf{0.27}$ & $ 0.3 $ & $\mathbf{0.27}$\\
Gustav Vasa & $ 18 $ & $ 4249 $ & $ 1.085 $ & $ (0.079) $ & $ 0.266 $ & $ 34.181 $ & $ (0.841) $ & $ 1.658 $ & $\mathbf{3.52}$ & $ (5.21) $ & $ 11.99 $ & $ 1.145 $ & $ (0.101) $ & $\mathbf{0.099}$ & $ 0.1 $ & $ 32.266 $ & $ (0.751) $ & $ 0.839 $ & $ 0.841 $ & $ 3.15 $ & $ (0.31) $ & $ 0.48 $ & $ 0.48 $\\
GustavIIAdolf & $ 57 $ & $ 5813 $ & $ 9.714 $ & $\mathbf{0.134}$ & $ 0.333 $ & $ 67.784 $ & $\mathbf{0.435}$ & $ 1.398 $ & $ 13.91 $ & $\mathbf{6.49}$ & $ 31.08 $ & $ 8.524 $ & $\mathbf{0.004}$ & $\mathbf{0.004}$ & $\mathbf{0.004}$ & $ 58.458 $ & $\mathbf{0.021}$ & $\mathbf{0.021}$ & $\mathbf{0.021}$ & $ 11.49 $ & $\mathbf{0.26}$ & $\mathbf{0.26}$ & $\mathbf{0.26}$\\
Jonas Ahlstromer & $ 40 $ & $ 2021 $ & $ 10.888 $ & $ (0.35) $ & $ 0.895 $ & $ 50.19 $ & $ (1.994) $ & $ 10.154 $ & $\mathbf{10.82}$ & $ (36.48) $ & $ 236.41 $ & $ 10.451 $ & $ (0.01) $ & $ 1.259 $ & $ 0.011 $ & $ 47.117 $ & $ (0.082) $ & $ 5.391 $ & $\mathbf{0.036}$ & $ 8.41 $ & $ (0.69) $ & $ 4.69 $ & $\mathbf{0.22}$\\
King's College University Of Toronto & $ 77 $ & $ 7087 $ & $ 0.235 $ & $ (0.152) $ & $ (1.781) $ & $ 0.989 $ & $ (0.645) $ & $ (1.07) $ & $\mathbf{0.9}$ & $ (11.87) $ & $ (27.29) $ & $ 0.017 $ & $ (0.005) $ & $ (1.877) $ & $ 0.017 $ & $ 0.085 $ & $ (0.059) $ & $ (4.624) $ & $ 0.084 $ & $\mathbf{0.34}$ & $ (0.35) $ & $ (7.12) $ & $\mathbf{0.34}$\\
Lund University Sphinx & $ 70 $ & $ 32668 $ & $ 4.585 $ & $\mathbf{0.228}$ & $ 1.199 $ & $ 19.522 $ & $\mathbf{0.738}$ & $ 3.476 $ & $\mathbf{4.78}$ & $ 7.19 $ & $ 60.64 $ & $ 2.191 $ & $ 0.016 $ & $ 1.512 $ & $\mathbf{0.009}$ & $ 8.752 $ & $ 0.058 $ & $ 5.452 $ & $\mathbf{0.033}$ & $ 1.36 $ & $ 0.4 $ & $ 4.58 $ & $\mathbf{0.39}$\\
Nijo Castle Gate & $ 19 $ & $ 7348 $ & $ 0.286 $ & $\mathbf{0.141}$ & $ 0.348 $ & $ 1.495 $ & $\mathbf{0.399}$ & $ 2.097 $ & $\mathbf{1.7}$ & $ 11.18 $ & $ 154.96 $ & $ 0.012 $ & $\mathbf{0.011}$ & $ 0.19 $ & $\mathbf{0.011}$ & $ 0.069 $ & $\mathbf{0.064}$ & $ 0.744 $ & $\mathbf{0.064}$ & $\mathbf{0.73}$ & $\mathbf{0.73}$ & $ 4.84 $ & $\mathbf{0.73}$\\
Pantheon Paris & $ 179 $ & $ 29383 $ & $\mathbf{0.05}$ & $ 0.867 $ & $ 1.275 $ & $\mathbf{0.192}$ & $ 3.766 $ & $ 2.655 $ & $\mathbf{1.47}$ & $ 79.24 $ & $ 39.69 $ & $\mathbf{0.005}$ & $ 0.595 $ & $ 0.011 $ & - & $\mathbf{0.04}$ & $ 3.208 $ & $ 0.072 $ & - & $\mathbf{0.49}$ & $ 9.71 $ & $ 0.82 $ & -\\
Park Gate Clermont Ferrand & $ 34 $ & $ 9099 $ & $ 0.125 $ & $\mathbf{0.083}$ & $ 0.1 $ & $ 0.391 $ & $\mathbf{0.203}$ & $ 0.296 $ & $\mathbf{0.57}$ & $ 1.71 $ & $ 10.5 $ & $\mathbf{0.022}$ & $\mathbf{0.022}$ & $\mathbf{0.022}$ & $\mathbf{0.022}$ & $\mathbf{0.049}$ & $\mathbf{0.049}$ & $\mathbf{0.049}$ & $\mathbf{0.049}$ & $\mathbf{0.35}$ & $\mathbf{0.35}$ & $\mathbf{0.35}$ & $\mathbf{0.35}$\\
Plaza De Armas Santiago & $ 240 $ & $ 26969 $ & $ 2.944 $ & $\mathbf{2.45}$ & - & $ 6.782 $ & $\mathbf{6.291}$ & - & $\mathbf{7.4}$ & $ 146.56 $ & - & $ 1.383 $ & $ 2.244 $ & - & $\mathbf{0.048}$ & $ 2.556 $ & $ 6.344 $ & - & $\mathbf{0.122}$ & $ 4.9 $ & $ 15.61 $ & - & $\mathbf{1.13}$\\
Porta San Donato Bologna & $ 141 $ & $ 25490 $ & $\mathbf{0.388}$ & $ 0.949 $ & $ 1.588 $ & $ 2.153 $ & $\mathbf{1.013}$ & $ 1.381 $ & $\mathbf{2.28}$ & $ 29.5 $ & $ 46.12 $ & $\mathbf{0.046}$ & $ 0.169 $ & $ 0.067 $ & $ 0.047 $ & $\mathbf{0.095}$ & $ 0.513 $ & $ 0.149 $ & $ 0.099 $ & $\mathbf{0.75}$ & $ 3.23 $ & $ 1.16 $ & $\mathbf{0.75}$\\
Round Church Cambridge & $ 92 $ & $ 84643 $ & $ 1.003 $ & $ 0.486 $ & $\mathbf{0.217}$ & $ 2.451 $ & $ 1.021 $ & $\mathbf{0.634}$ & $\mathbf{2.66}$ & $ 19.04 $ & $ 9.6 $ & $ 0.582 $ & $ 0.493 $ & $\mathbf{0.012}$ & $\mathbf{0.012}$ & $ 1.107 $ & $ 1.851 $ & $\mathbf{0.033}$ & $ 0.035 $ & $ 1.54 $ & $ 2.03 $ & $ 0.41 $ & $\mathbf{0.39}$\\
Skansen Kronan Gothenburg & $ 131 $ & $ 28371 $ & $ 0.226 $ & $\mathbf{0.223}$ & $ (0.234) $ & $ 0.736 $ & $\mathbf{0.549}$ & $ (0.679) $ & $\mathbf{1.24}$ & $ 8.82 $ & $ (18.49) $ & $ 0.008 $ & $ 0.008 $ & $ (0.007) $ & $ 0.008 $ & $ 0.026 $ & $ 0.025 $ & $ (0.02) $ & $ 0.025 $ & $\mathbf{0.67}$ & $\mathbf{0.67}$ & $ (0.69) $ & $\mathbf{0.67}$\\
Smolny Cathedral St Petersburg & $ 131 $ & $ 51115 $ & $\mathbf{0.051}$ & $ 0.209 $ & - & $ 0.554 $ & $\mathbf{0.493}$ & - & $\mathbf{1.66}$ & $ 19.01 $ & - & $\mathbf{0.006}$ & $ 0.007 $ & - & $\mathbf{0.006}$ & $ 0.033 $ & $\mathbf{0.028}$ & - & $ 0.029 $ & $\mathbf{0.81}$ & $ 1.0 $ & - & $\mathbf{0.81}$\\
Some Cathedral In Barcelona & $ 177 $ & $ 30367 $ & $\mathbf{0.315}$ & $ 1.776 $ & $ 1.261 $ & $\mathbf{0.88}$ & $ 1.519 $ & $ 3.126 $ & $\mathbf{2.87}$ & $ 47.12 $ & $ 66.97 $ & $ 0.011 $ & $ 0.013 $ & $ 0.024 $ & $\mathbf{0.01}$ & $ 0.026 $ & $ 0.031 $ & $ 0.057 $ & $\mathbf{0.025}$ & $\mathbf{0.89}$ & $ 1.09 $ & $ 2.09 $ & $\mathbf{0.89}$\\
Sri Mariamman Singapore & $ 222 $ & $ 56220 $ & $\mathbf{0.683}$ & $ 1.758 $ & $ 0.721 $ & $ 2.302 $ & $\mathbf{1.433}$ & $ 1.615 $ & $\mathbf{4.13}$ & $ 52.13 $ & $ 37.16 $ & $\mathbf{0.023}$ & $ 0.614 $ & $ 0.025 $ & $\mathbf{0.023}$ & $\mathbf{0.077}$ & $ 2.158 $ & $ 0.083 $ & $ 0.078 $ & $ 0.91 $ & $ 7.4 $ & $ 1.17 $ & $\mathbf{0.89}$\\
Sri Thendayuthapani Singapore & $ 98 $ & $ 88849 $ & $ 3.812 $ & $ (0.285) $ & $ 0.375 $ & $ 46.269 $ & $ (1.561) $ & $ 1.581 $ & $ 23.37 $ & $ (15.93) $ & $ 19.57 $ & $ 2.87 $ & $ (0.053) $ & $\mathbf{0.034}$ & $\mathbf{0.034}$ & $ 44.17 $ & $ (0.329) $ & $\mathbf{0.138}$ & $\mathbf{0.138}$ & $ 8.44 $ & $ (0.56) $ & $ 0.72 $ & $ 0.67 $\\
Sri Veeramakaliamman Singapore & $ 157 $ & $ 130013 $ & $ 0.597 $ & $ (1.966) $ & $\mathbf{0.273}$ & $ 2.559 $ & $ (1.807) $ & $\mathbf{0.519}$ & $\mathbf{3.47}$ & $ (205.96) $ & $ 18.08 $ & $ 0.04 $ & $ (1.388) $ & $ 0.095 $ & $\mathbf{0.038}$ & $ 0.175 $ & $ (3.41) $ & $ 0.288 $ & $\mathbf{0.169}$ & $ 0.73 $ & $ (34.72) $ & $ 2.2 $ & $\mathbf{0.71}$\\
Statue Of Liberty & $ 134 $ & $ 49250 $ & $ 20.012 $ & $ (4.55) $ & $\mathbf{3.031}$ & $ 46.887 $ & $ (3.449) $ & $\mathbf{3.357}$ & $\mathbf{26.16}$ & $ (1031.8) $ & $ 133.81 $ & $ 4.122 $ & $ (4.782) $ & $ 28.049 $ & $\mathbf{0.099}$ & $ 9.091 $ & $ (8.281) $ & $ 2.945 $ & $\mathbf{0.213}$ & $ 6.97 $ & $ (52.05) $ & $ 5.08 $ & $\mathbf{1.25}$\\
The Pumpkin & $ 196 $ & $ 69341 $ & $ 14.89 $ & $\mathbf{0.513}$ & $ (1.656) $ & $ 94.672 $ & $\mathbf{2.036}$ & $ (4.215) $ & $ 33.41 $ & $\mathbf{9.71}$ & $ (122.54) $ & $ 14.952 $ & $\mathbf{0.022}$ & $ (14.862) $ & $\mathbf{0.022}$ & $ 98.862 $ & $ 0.092 $ & $ (3.123) $ & $\mathbf{0.091}$ & $ 24.85 $ & $\mathbf{0.57}$ & $ (24.19) $ & $\mathbf{0.57}$\\
Thian Hook Keng Temple Singapore & $ 138 $ & $ 34288 $ & $\mathbf{0.082}$ & $ 0.519 $ & $ 0.404 $ & $\mathbf{0.832}$ & $ 2.751 $ & $ 3.047 $ & $\mathbf{2.75}$ & $ 53.79 $ & $ 62.7 $ & $\mathbf{0.008}$ & $ 0.024 $ & $ 0.043 $ & $\mathbf{0.008}$ & $\mathbf{0.081}$ & $ 0.245 $ & $ 0.424 $ & $ 0.084 $ & $ 1.13 $ & $ 3.32 $ & $ 4.92 $ & $\mathbf{1.12}$\\
Tsar Nikolai I & $ 98 $ & $ 37857 $ & $ 9.467 $ & $\mathbf{0.219}$ & $ 0.261 $ & $ 48.499 $ & $\mathbf{0.475}$ & $ 1.437 $ & $ 9.79 $ & $\mathbf{5.19}$ & $ 32.86 $ & $ 7.836 $ & $\mathbf{0.005}$ & $\mathbf{0.005}$ & $\mathbf{0.005}$ & $ 36.28 $ & $\mathbf{0.018}$ & $\mathbf{0.018}$ & $\mathbf{0.018}$ & $ 6.53 $ & $\mathbf{0.33}$ & $\mathbf{0.33}$ & $\mathbf{0.33}$\\
Urban II & $ 96 $ & $ 22284 $ & $ 9.467 $ & $\mathbf{0.774}$ & $ 2.044 $ & $ 47.49 $ & $\mathbf{2.077}$ & $ 8.951 $ & $\mathbf{9.38}$ & $ 31.71 $ & $ 176.19 $ & $ 9.586 $ & $ 0.036 $ & $ 3.038 $ & $\mathbf{0.021}$ & $ 48.214 $ & $ 0.175 $ & $ 16.348 $ & $\mathbf{0.107}$ & $ 6.92 $ & $ 0.72 $ & $ 17.61 $ & $\mathbf{0.38}$\\
Vercingetorix & $ 69 $ & $ 10754 $ & $ 8.788 $ & $\mathbf{1.158}$ & $ 2.786 $ & $ 69.328 $ & $\mathbf{2.203}$ & $ 2.365 $ & $\mathbf{5.08}$ & $ 15.87 $ & $ 65.57 $ & $ 3.104 $ & $ 0.3 $ & $ 1.564 $ & $\mathbf{0.011}$ & $ 17.706 $ & $ 1.431 $ & $ 7.138 $ & $\mathbf{0.048}$ & $ 1.5 $ & $ 0.54 $ & $ 2.93 $ & $\mathbf{0.23}$\\
Yueh Hai Ching Temple Singapore & $ 43 $ & $ 13774 $ & $\mathbf{0.098}$ & $ (0.642) $ & $ 0.303 $ & $\mathbf{0.72}$ & $ (1.813) $ & $ 1.92 $ & $\mathbf{0.94}$ & $ (27.32) $ & $ 45.19 $ & $\mathbf{0.014}$ & $ (0.023) $ & $ 0.059 $ & $\mathbf{0.014}$ & $\mathbf{0.043}$ & $ (0.075) $ & $ 0.26 $ & $\mathbf{0.043}$ & $\mathbf{0.65}$ & $ (1.64) $ & $ 2.06 $ & $\mathbf{0.65}$\\
	        \bottomrule
	\end{tabular} 
	\end{adjustbox}
	\vspace{3pt}      
    \end{tabular}
    \caption{\small Single scene experiments in the calibrated setup. The table shows mean camera location error (denoted $\tr_{\mathrm{error}}$) in meters, mean orientation error (denoted $R_{\mathrm{error}}$) in degrees, and mean reprojection error in pixels obtained with our method before and after BA, compared to GESFM \cite{kasten2019algebraic}, Linear \cite{jiang2013global}, and Colmap \cite{Colmap_implementation}. ({\textit{ Smaller is better.}}) In parenthesis experiments in which at least 10\% of the cameras are removed.}
    \label{tab:Euclidean_Results_No_BA}
\end{table*}

\begin{figure}[h]

\begin{subfigure}{1\textwidth}
    \begin{subfigure}{0.5\textwidth}
    \centering
    \includegraphics[width=0.75\textwidth]{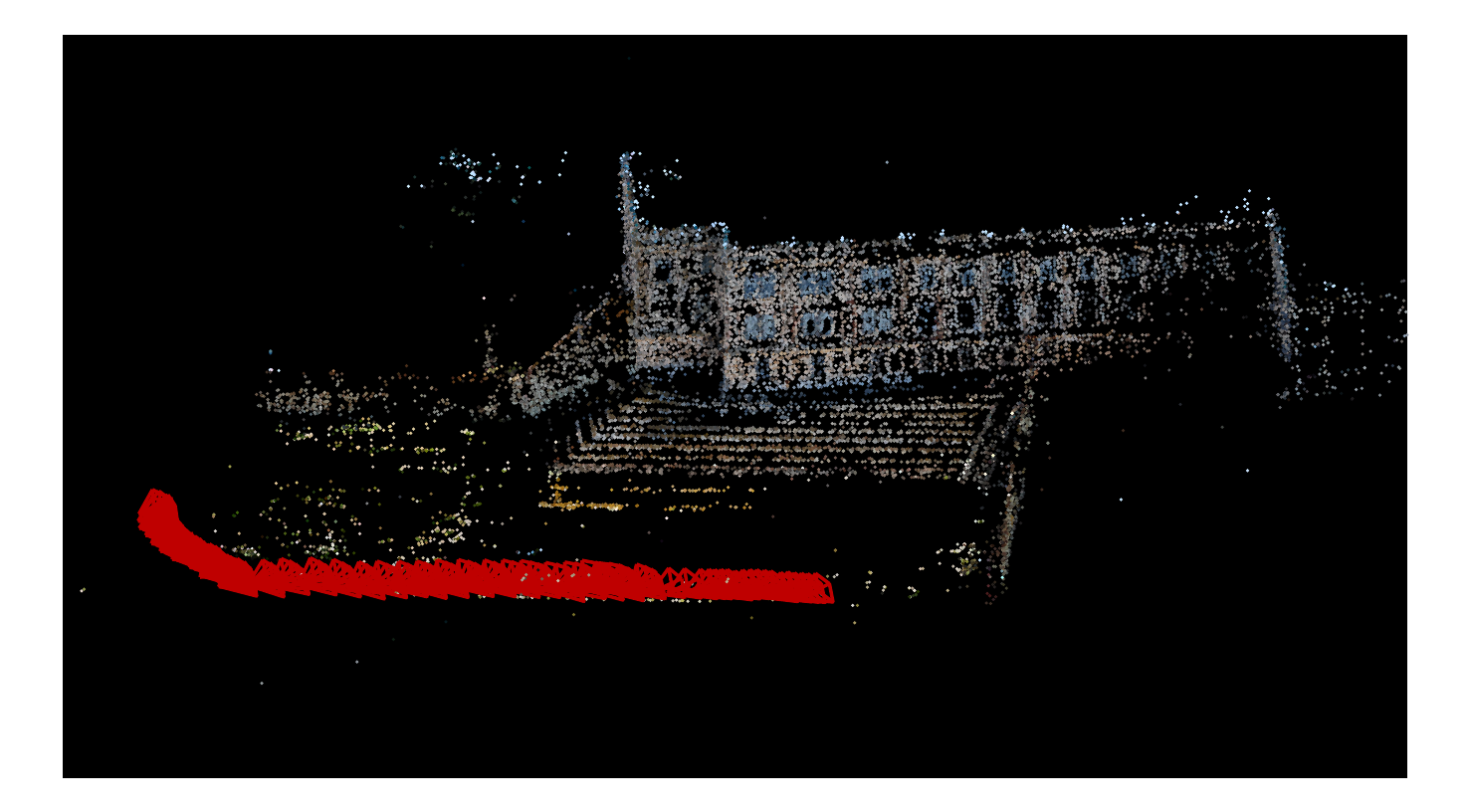} 
    \end{subfigure}
    \begin{subfigure}{0.5\textwidth}
    \centering
    \includegraphics[width=0.6\textwidth]{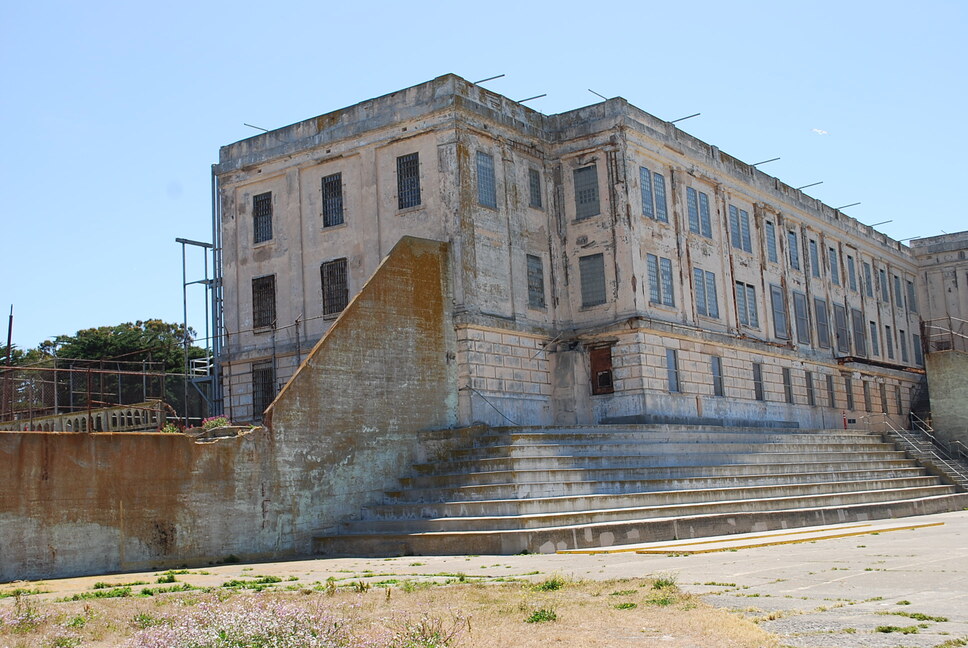} 
    \end{subfigure}
    \caption{Alcatraz Courtyard}
\end{subfigure}

\begin{subfigure}{1\textwidth}
    \begin{subfigure}{0.5\textwidth}
    \centering
    \includegraphics[width=0.75\textwidth]{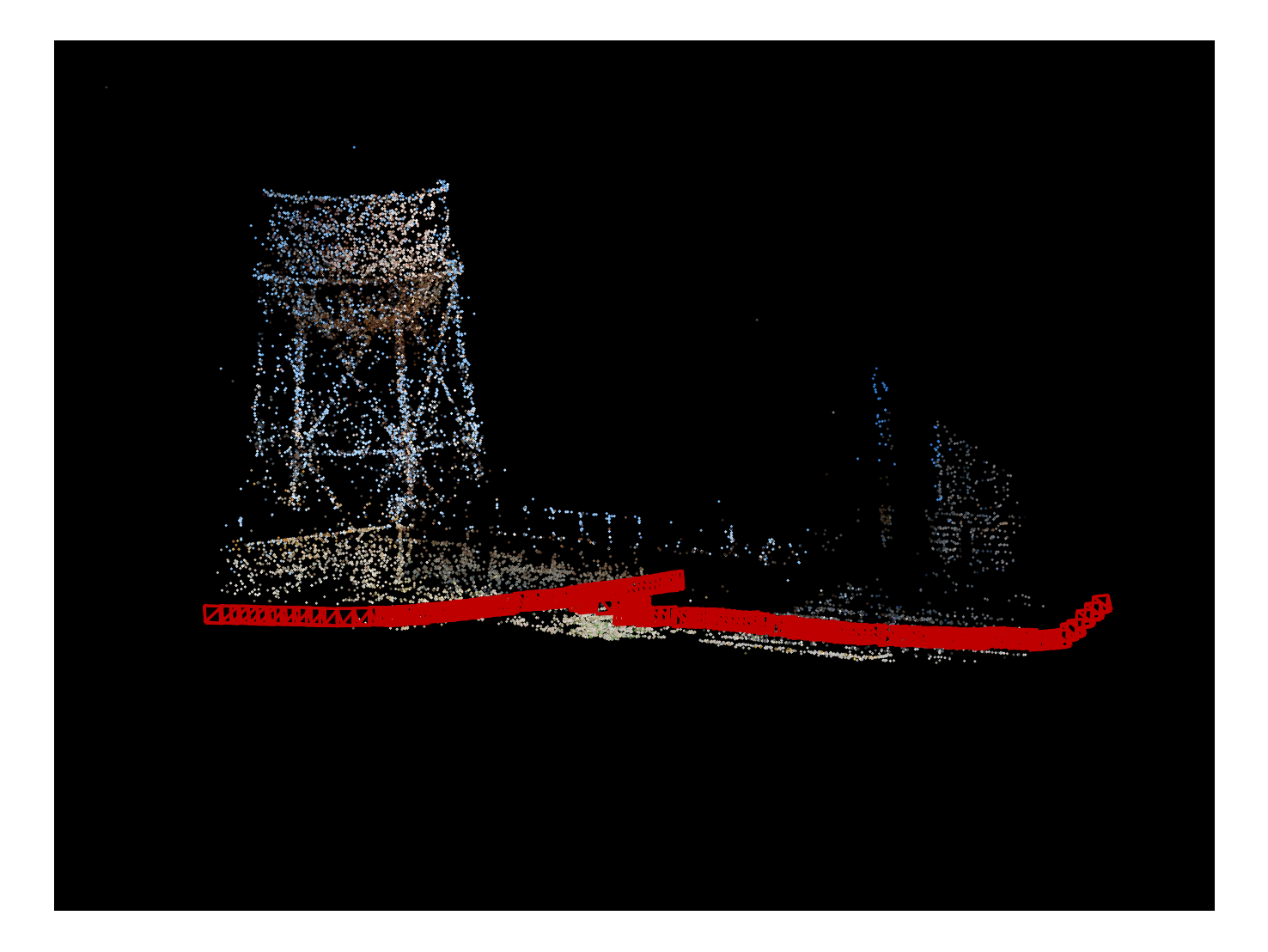}
    \end{subfigure}
    \begin{subfigure}{0.5\textwidth}
    \centering
    \includegraphics[width=0.4\textwidth]{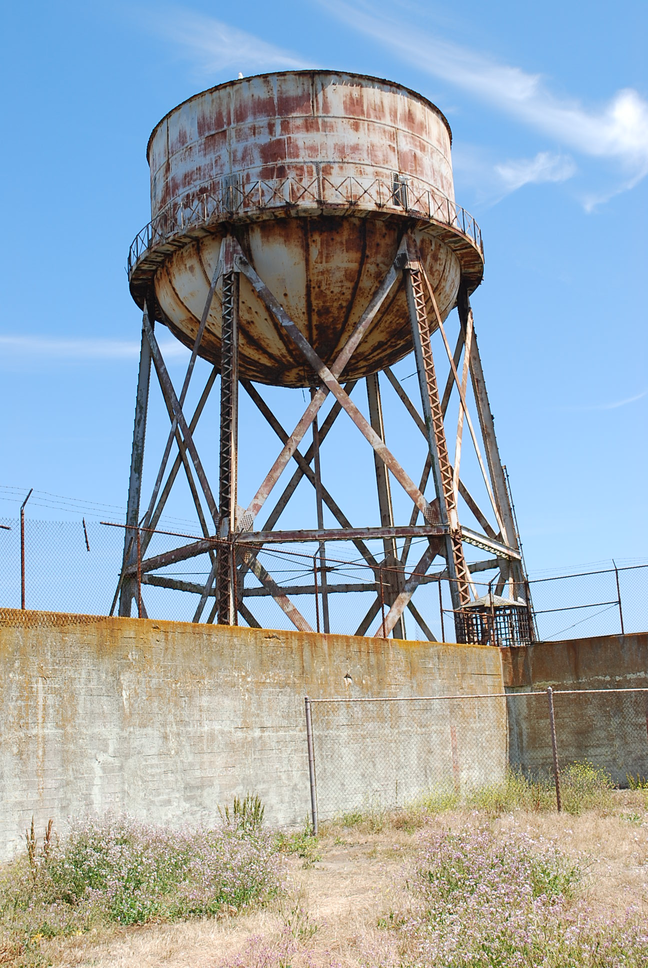}
    \end{subfigure}
    \caption{Alcatraz Water Tower}
\end{subfigure}

\begin{subfigure}{1\textwidth}
    \begin{subfigure}{0.5\textwidth}
    \centering
    \includegraphics[width=0.75\textwidth]{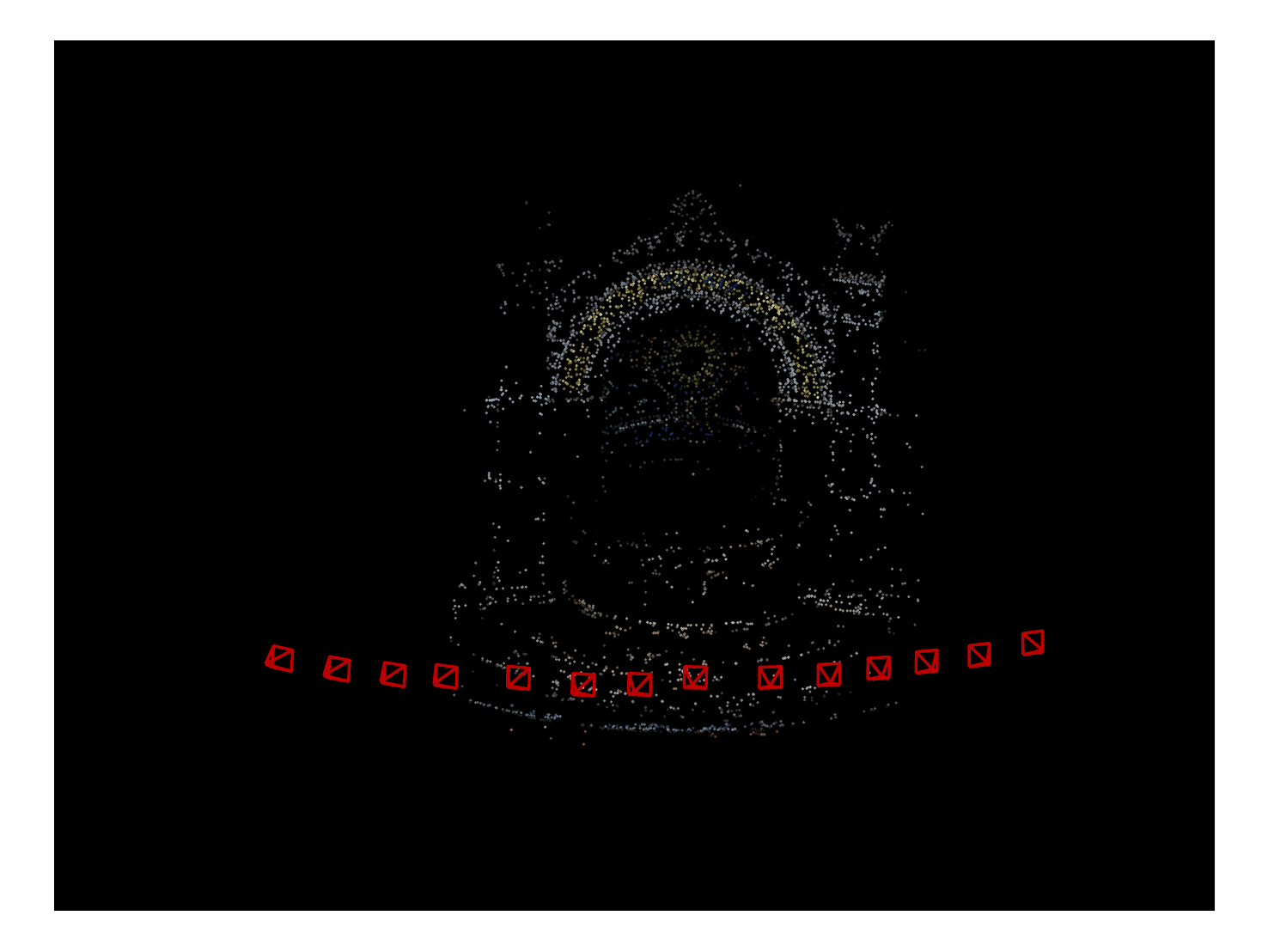}
    \end{subfigure}
    \begin{subfigure}{0.5\textwidth}
    \centering
    \includegraphics[width=0.4\textwidth]{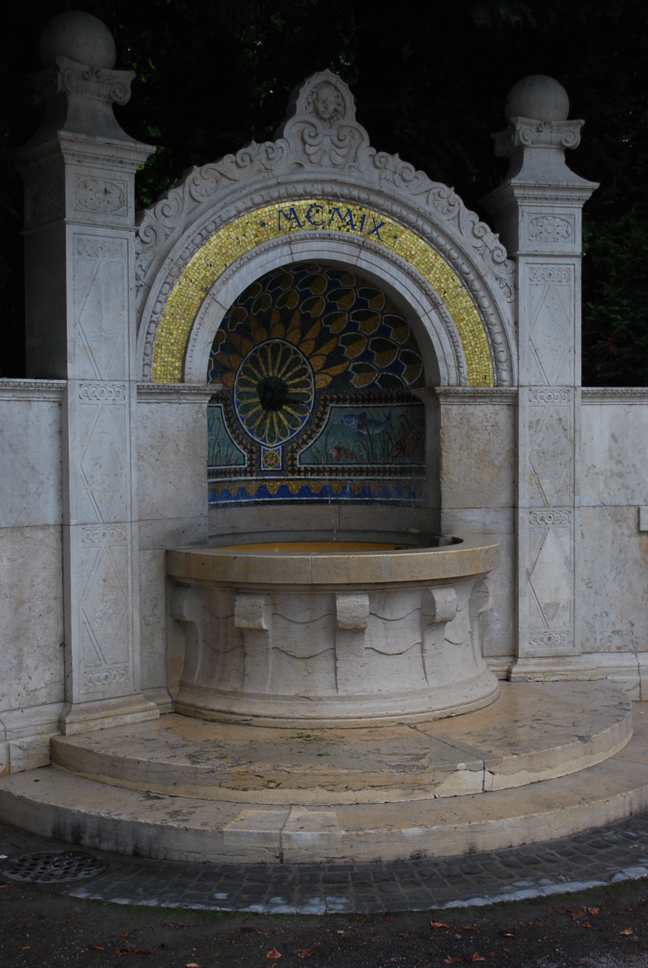}
    \end{subfigure}
    \caption{Drinking Fountain Somewhere In Zurich}
\end{subfigure}

\begin{subfigure}{1\textwidth}
    \begin{subfigure}{0.5\textwidth}
    \centering
    \includegraphics[width=0.75\textwidth]{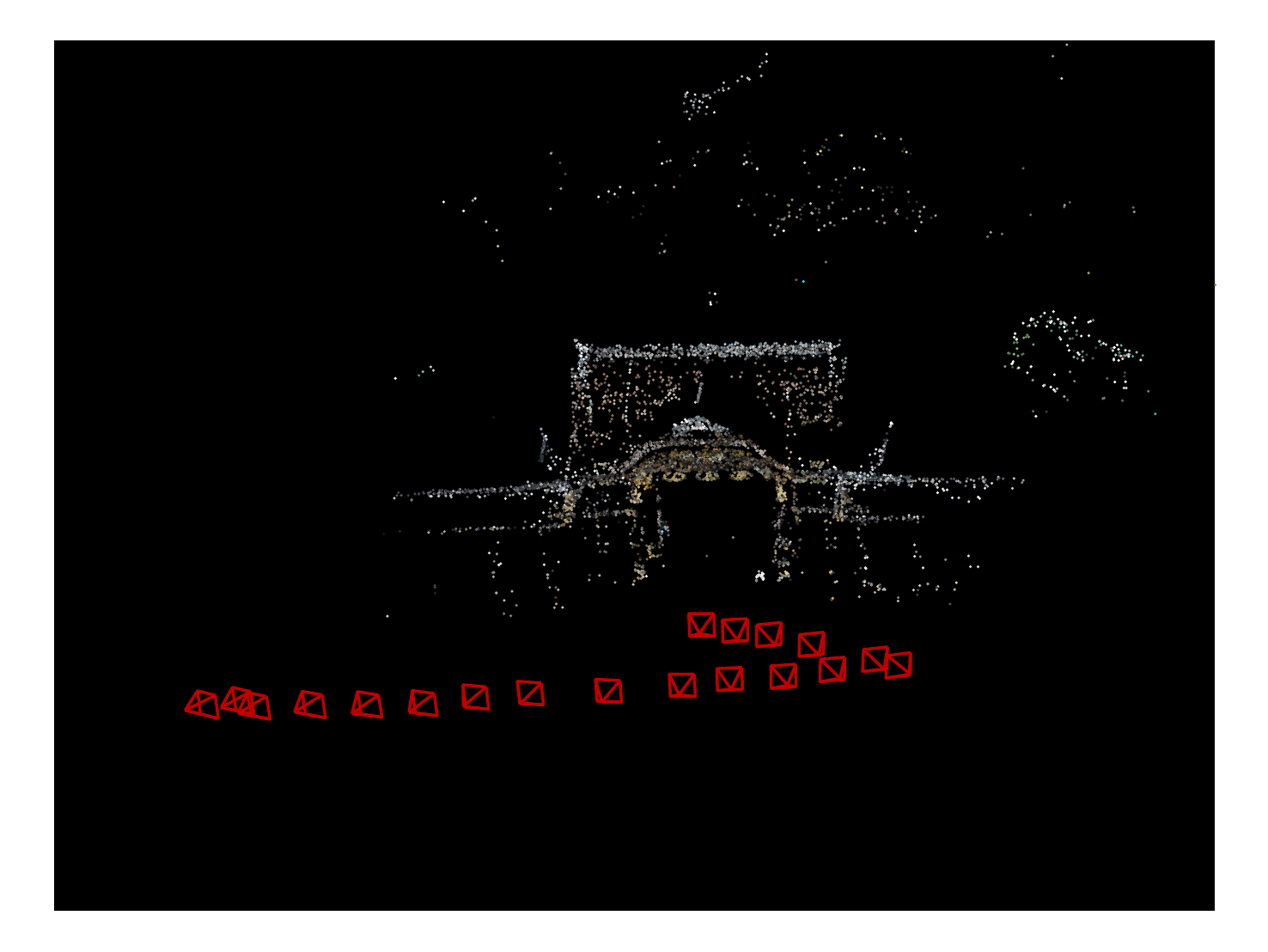} 
    \end{subfigure}
    \begin{subfigure}{0.5\textwidth}
    \centering
    \includegraphics[width=0.6\textwidth]{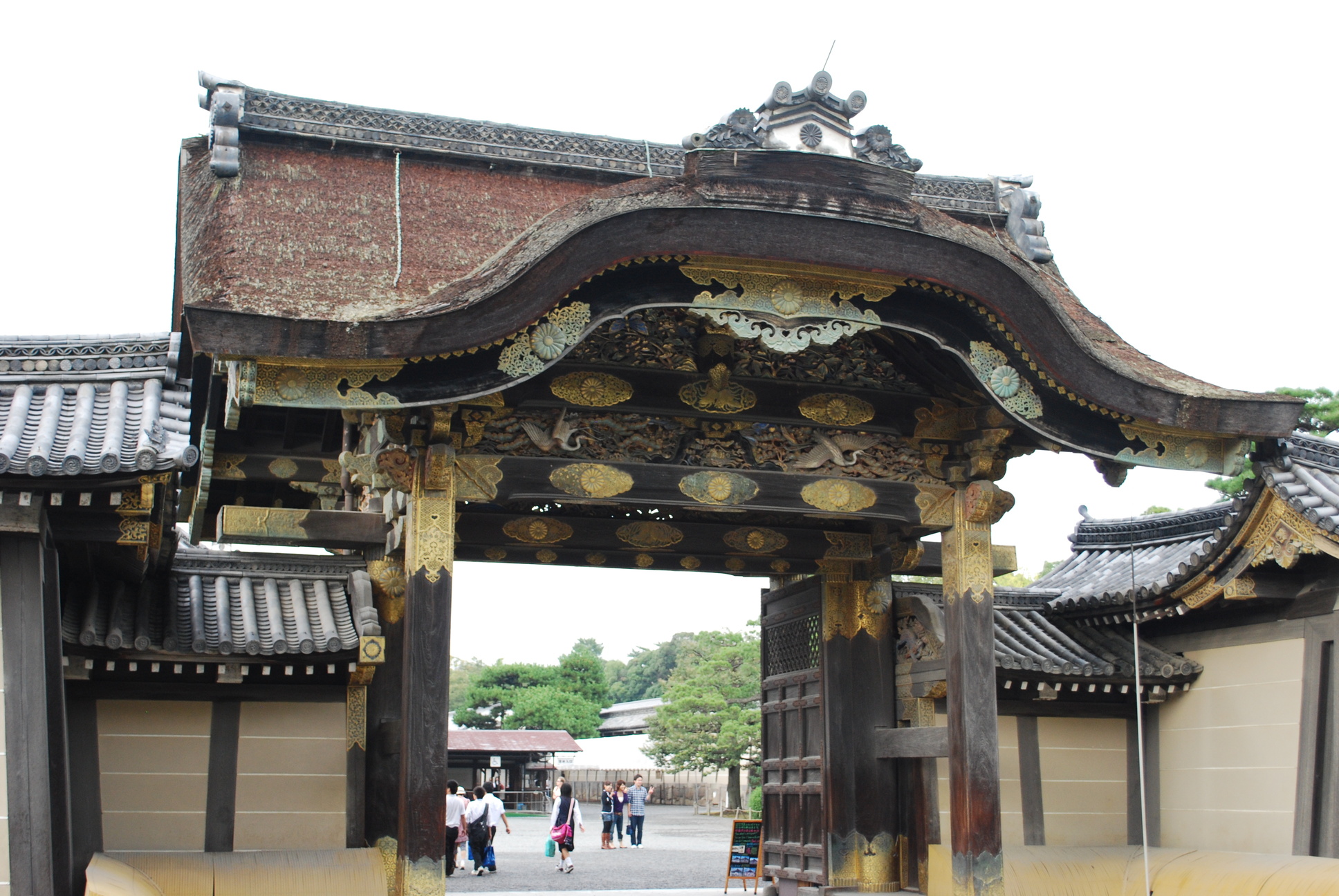}
    \end{subfigure}
    \caption{Nijo Castle Gate}
\end{subfigure}

\caption{\small Single scene 3D reconstructions and recovery of camera parameters  with our method. Each pair shows on the left the triangulated point cloud and the recovered camera locations and orientations (in red) and on the right one of the input images.}
\label{fig:optimization_reconstruction}
\end{figure}

\begin{figure}[h]
\begin{subfigure}{1\textwidth}
    \begin{subfigure}{0.5\textwidth}
    \centering
    \includegraphics[width=0.75\textwidth]{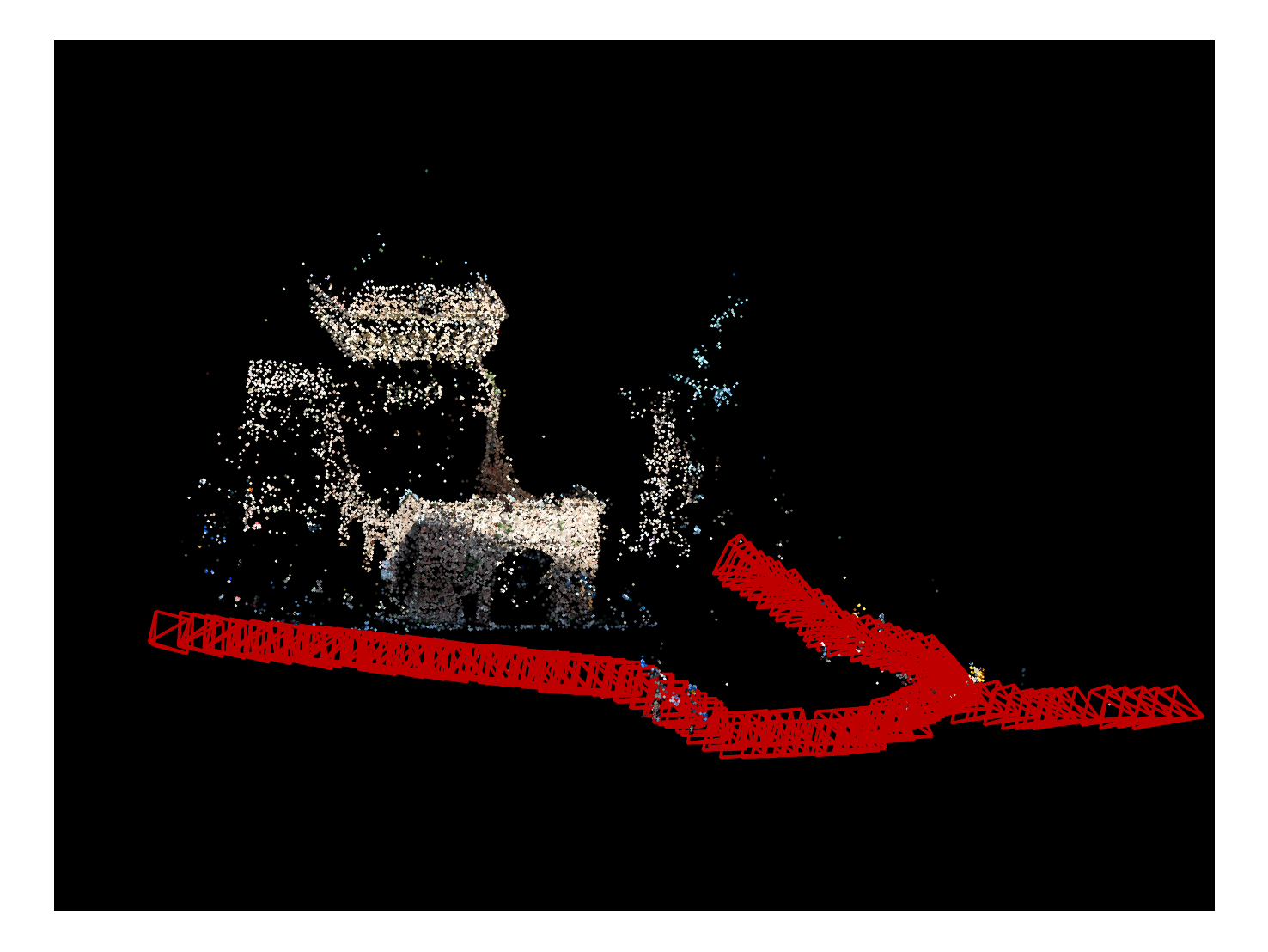} 
    \end{subfigure}
    \begin{subfigure}{0.5\textwidth}
    \centering
    \includegraphics[width=0.6\textwidth]{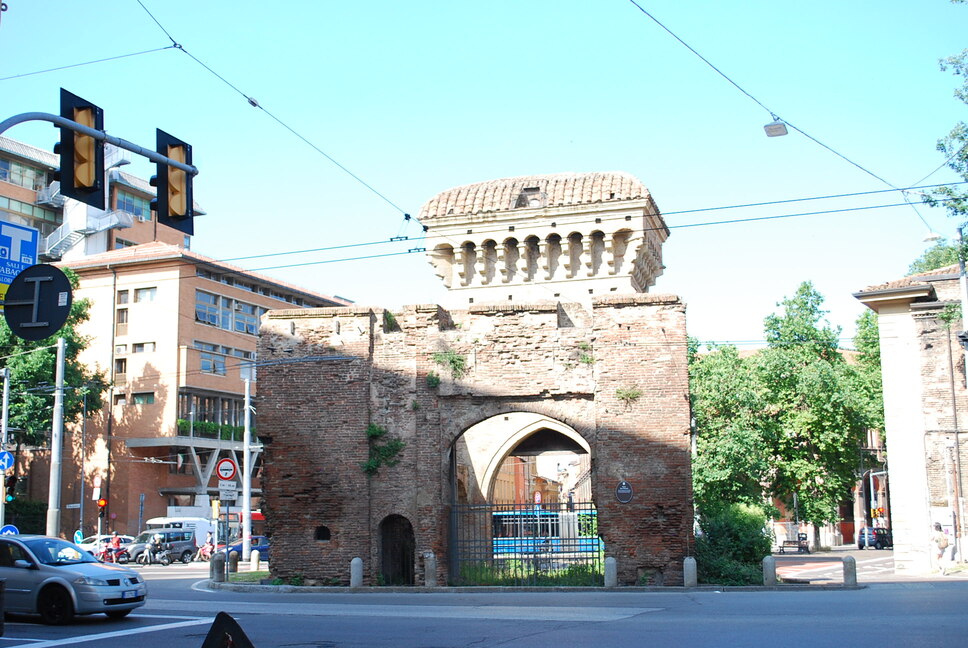}
    \end{subfigure}
    \caption{Porta San Donato Bologna}
\end{subfigure}

\begin{subfigure}{1\textwidth}
    \begin{subfigure}{0.5\textwidth}
    \centering
    \includegraphics[width=0.75\textwidth]{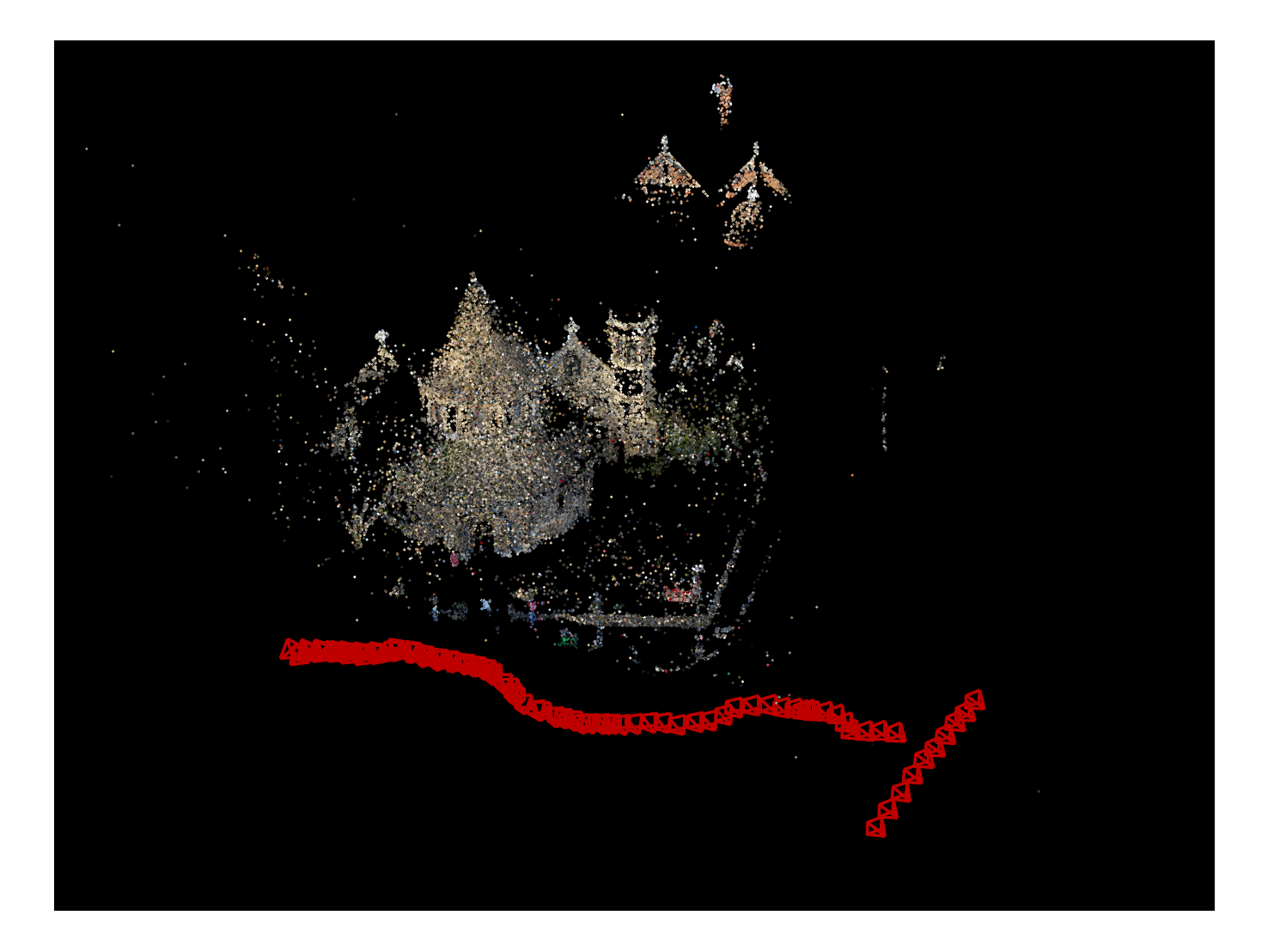} 
    \end{subfigure}
    \begin{subfigure}{0.5\textwidth}
    \centering
    \includegraphics[width=0.6\textwidth]{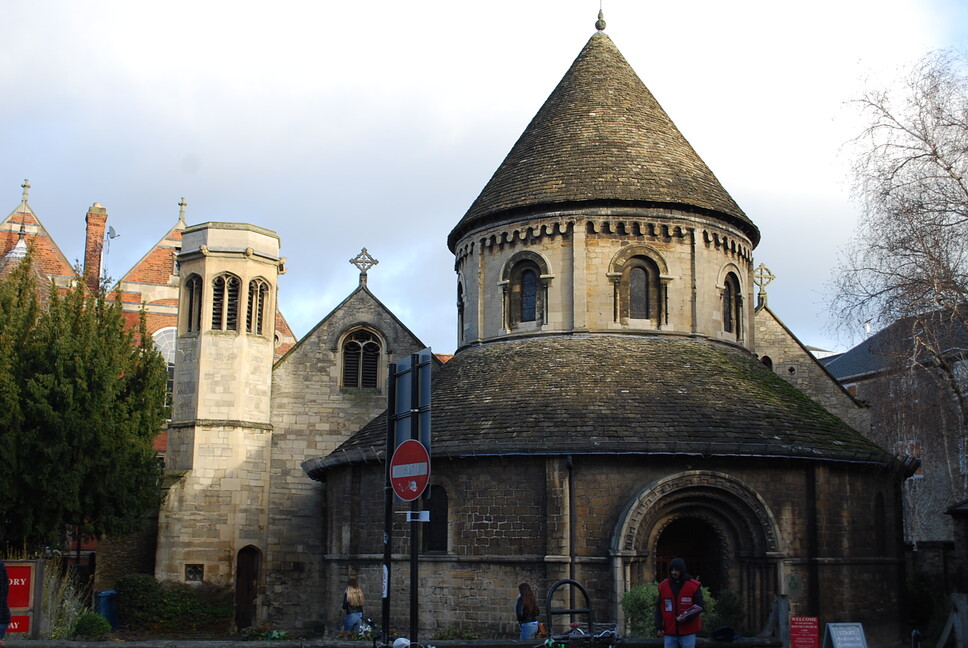}
    \end{subfigure}
    \caption{Round Church Cambridge}
\end{subfigure}

\begin{subfigure}{1\textwidth}
    \begin{subfigure}{0.5\textwidth}
    \centering
    \includegraphics[width=0.75\textwidth]{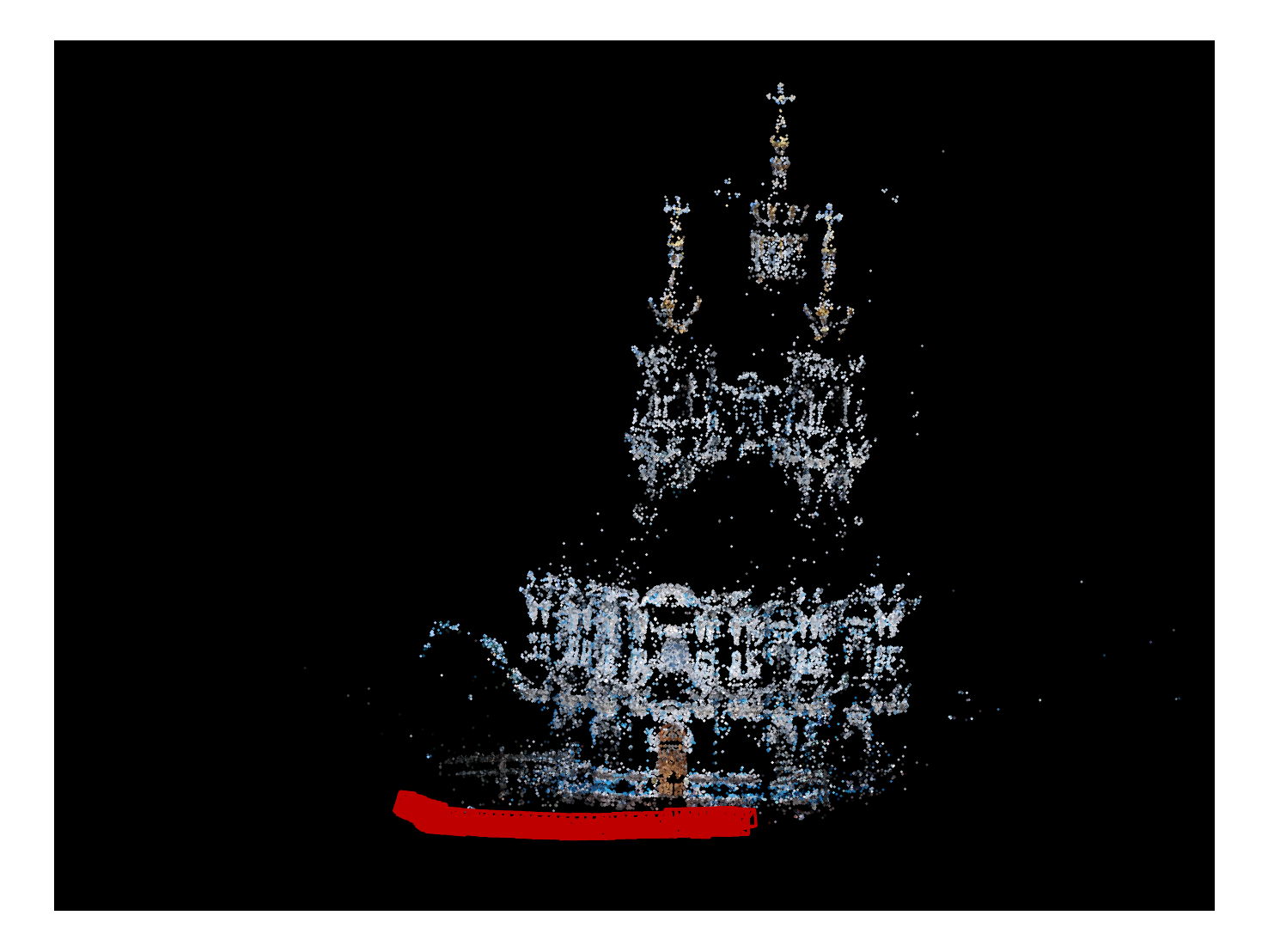} 
    \end{subfigure}
    \begin{subfigure}{0.5\textwidth}
    \centering
    \includegraphics[width=0.6\textwidth]{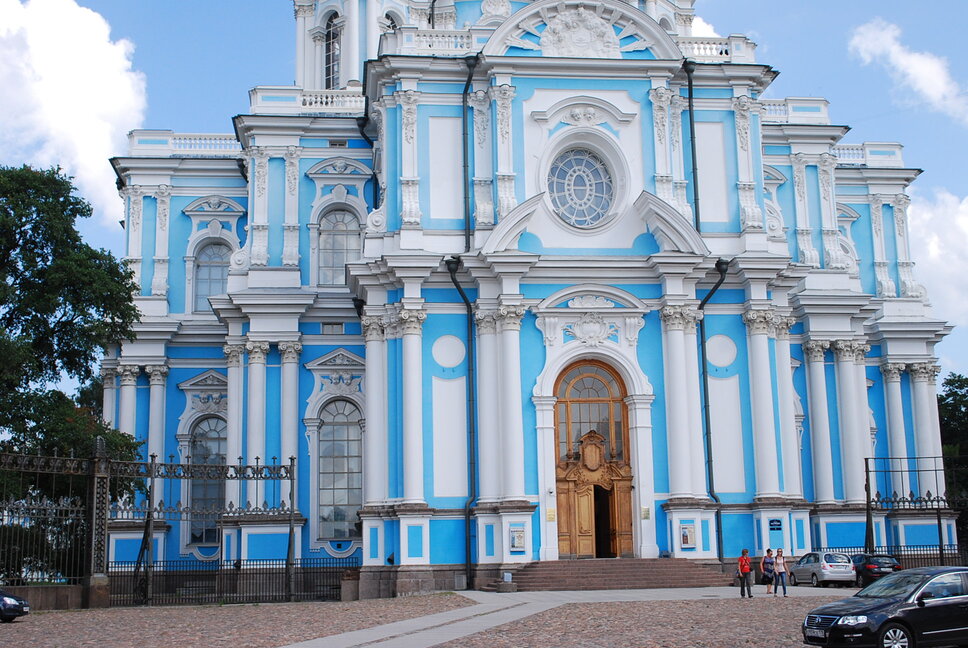}
    \end{subfigure}
    \caption{Smolny Cathedral St Petersburg}
\end{subfigure}

\begin{subfigure}{1\textwidth}
    \begin{subfigure}{0.5\textwidth}
    \centering
    \includegraphics[width=0.75\textwidth]{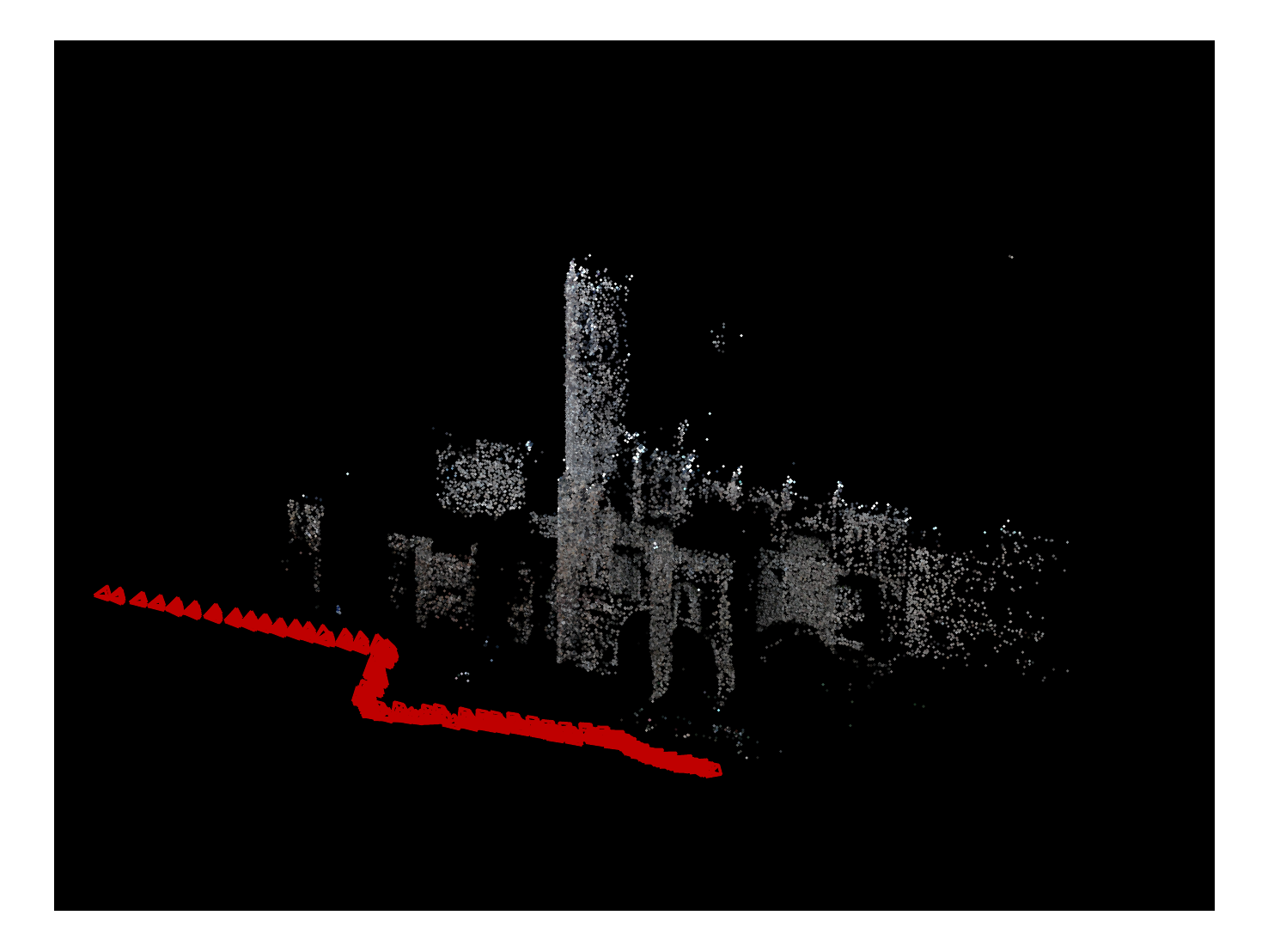} 
    \end{subfigure}
    \begin{subfigure}{0.5\textwidth}
    \centering
    \includegraphics[width=0.4\textwidth]{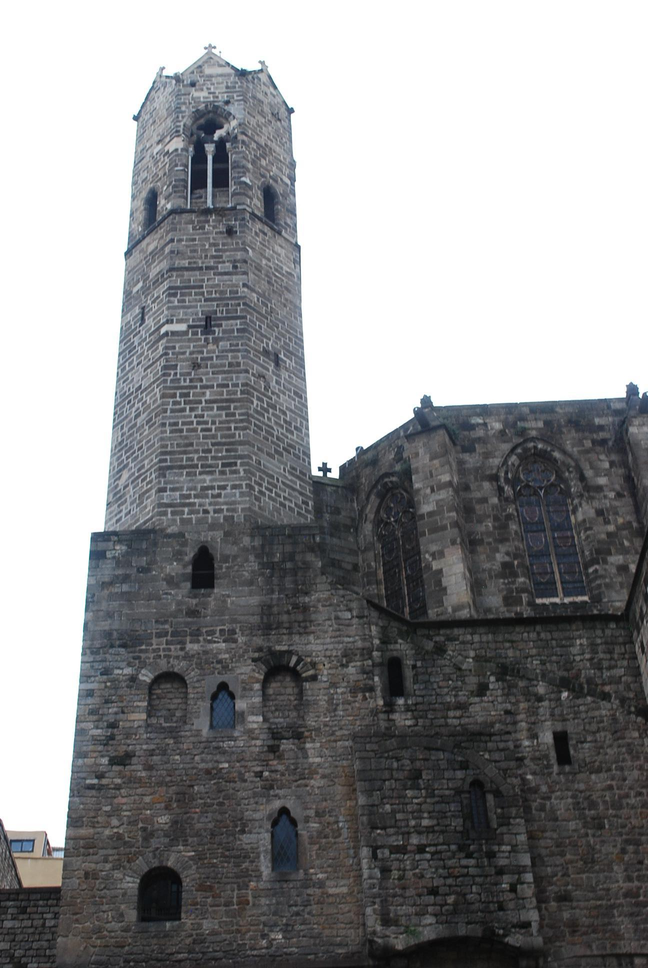}
    \end{subfigure}
    \caption{Some Cathedral In Barcelona}
\end{subfigure}

\caption{\small Single scene 3D reconstructions and recovery of camera parameters  with our method. Each pair shows on the left the triangulated point cloud and the recovered camera locations and orientations (in red) and on the right one of the input images.}
\label{fig:optimization_reconstruction2}
\end{figure}

\begin{figure}[h]

\begin{subfigure}{1\textwidth}
    \begin{subfigure}{0.5\textwidth}
    \centering
    \includegraphics[width=0.75\textwidth]{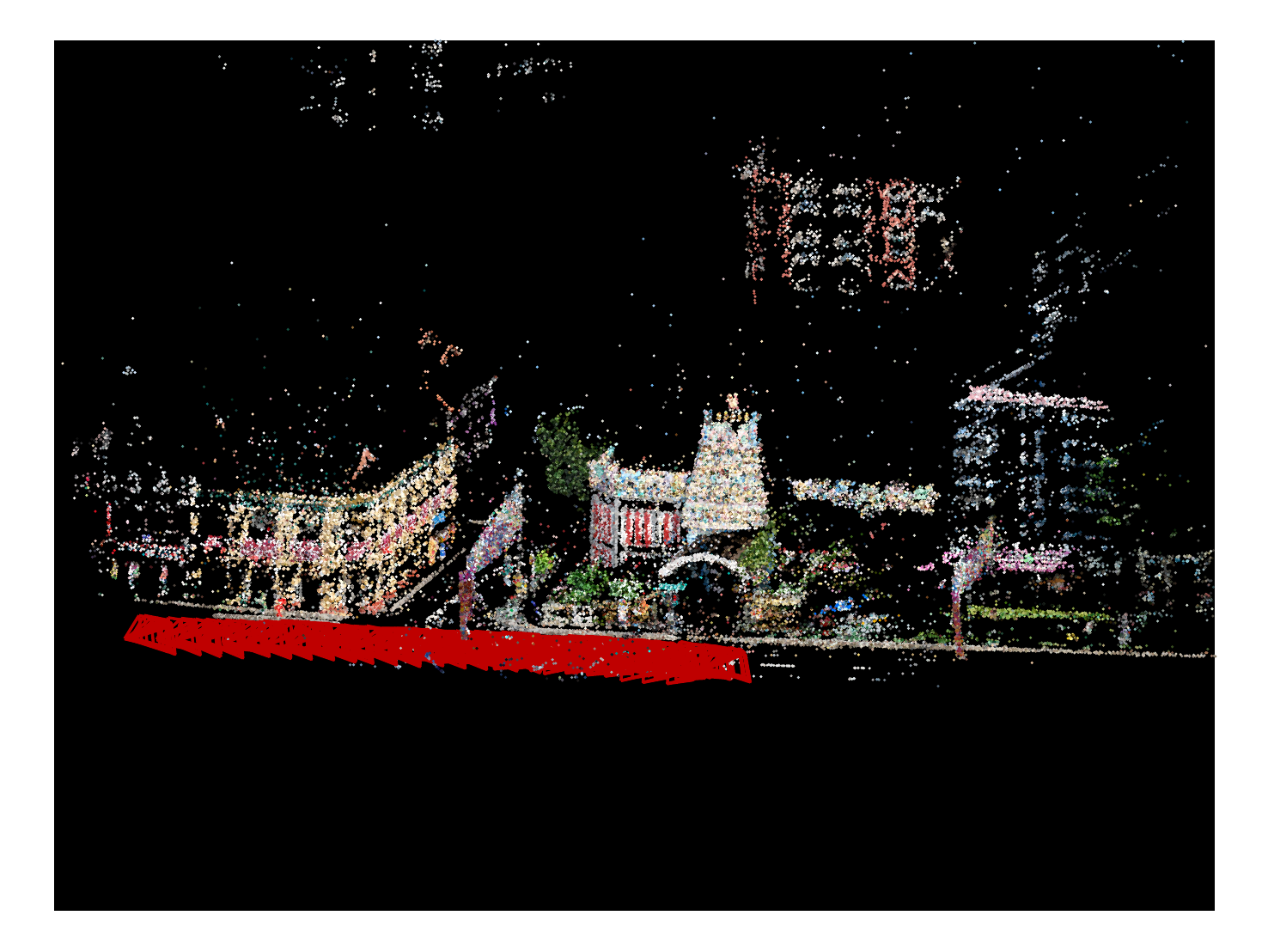} 
    \end{subfigure}
    \begin{subfigure}{0.5\textwidth}
    \centering
    \includegraphics[width=0.6\textwidth]{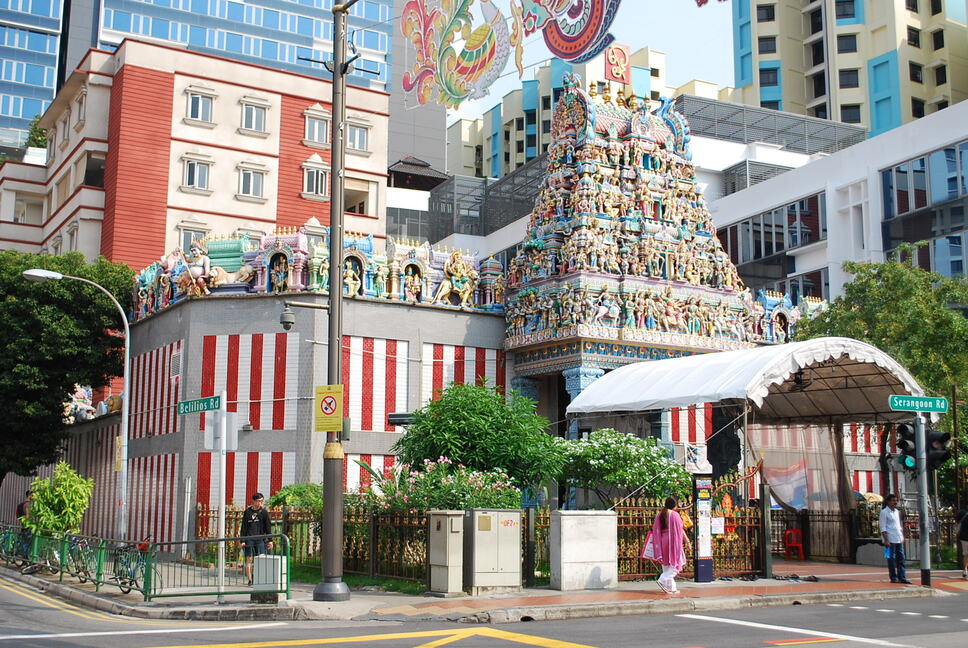}
    \end{subfigure}
    \caption{Sri Veeramakaliamman Singapore}
\end{subfigure}

\begin{subfigure}{1\textwidth}
    \begin{subfigure}{0.5\textwidth}
    \centering
    \includegraphics[width=0.75\textwidth]{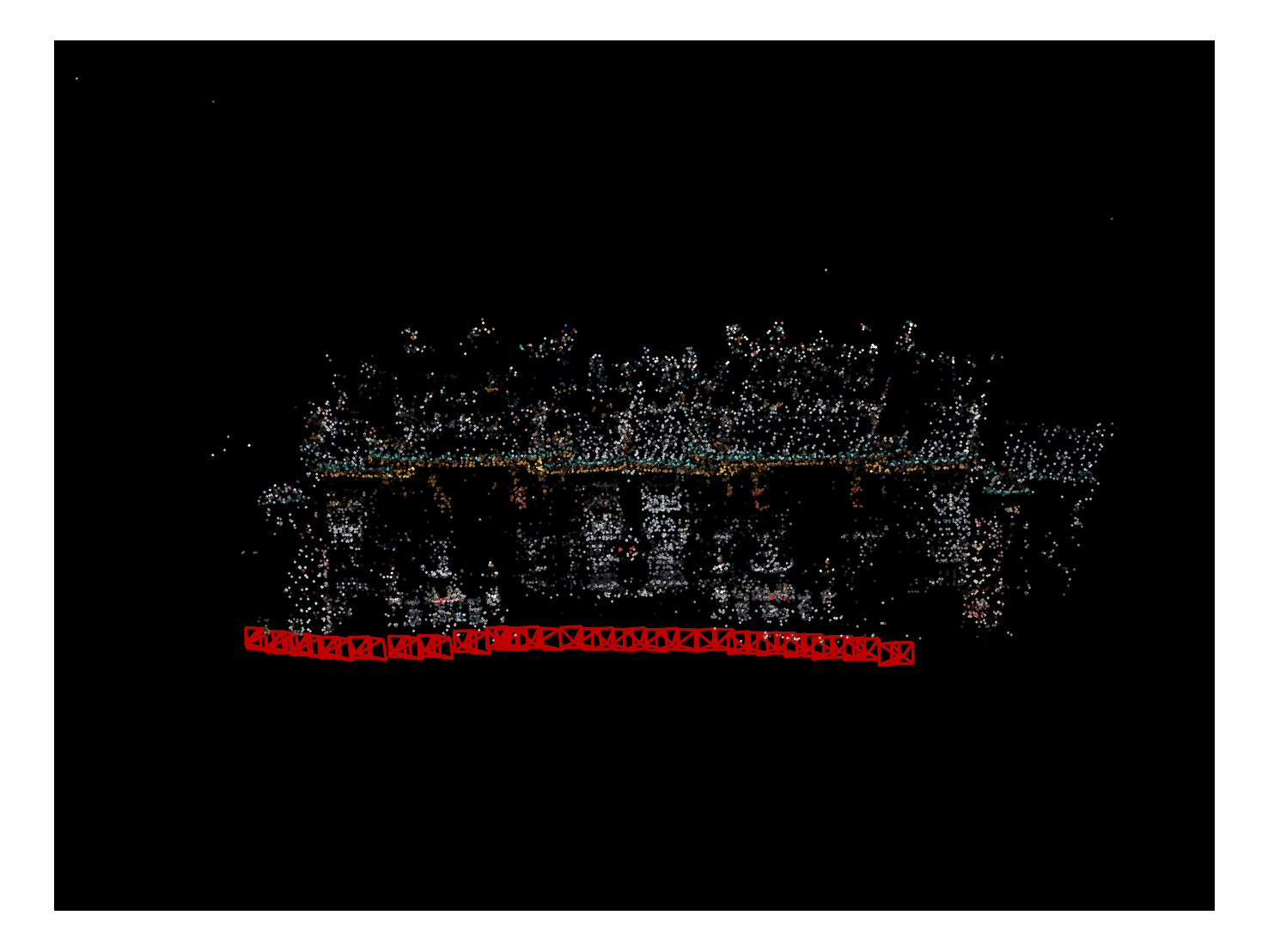} 
    \end{subfigure}
    \begin{subfigure}{0.5\textwidth}
    \centering
    \includegraphics[width=0.6\textwidth]{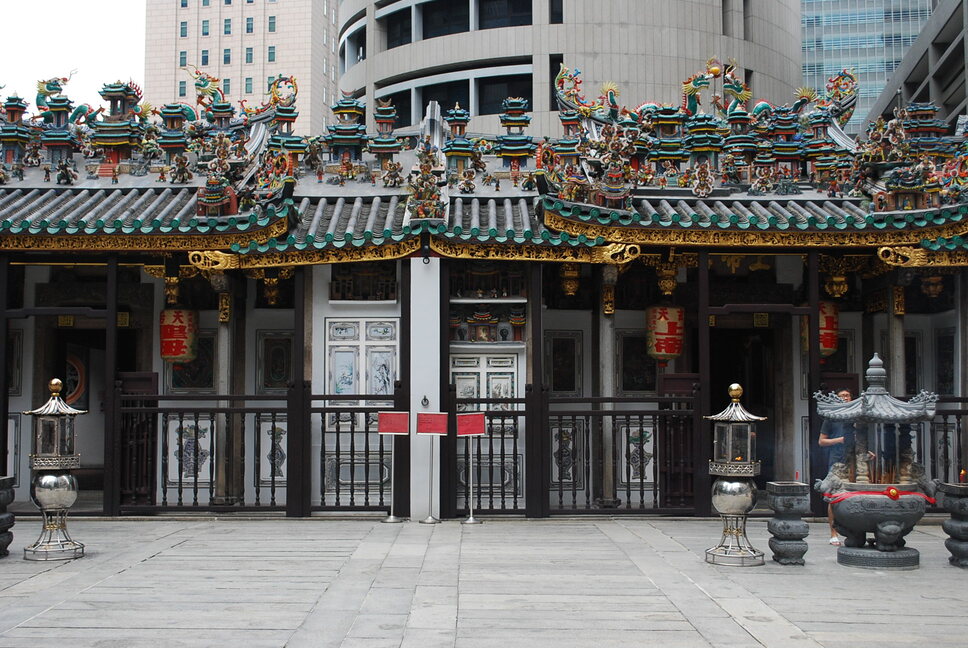}
    \end{subfigure}
    \caption{Yueh Hai Ching Temple Singapore}
\end{subfigure}

\caption{\small Single scene 3D reconstructions and recovery of camera parameters  with our method. Each pair shows on the left the triangulated point cloud and the recovered camera locations and orientations (in red) and on the right one of the input images.}
\label{fig:optimization_reconstruction3}
\end{figure}

\begin{figure}[h]
\begin{subfigure}{1\textwidth}
    \begin{subfigure}{0.5\textwidth}
    \centering
    \includegraphics[width=0.75\textwidth]{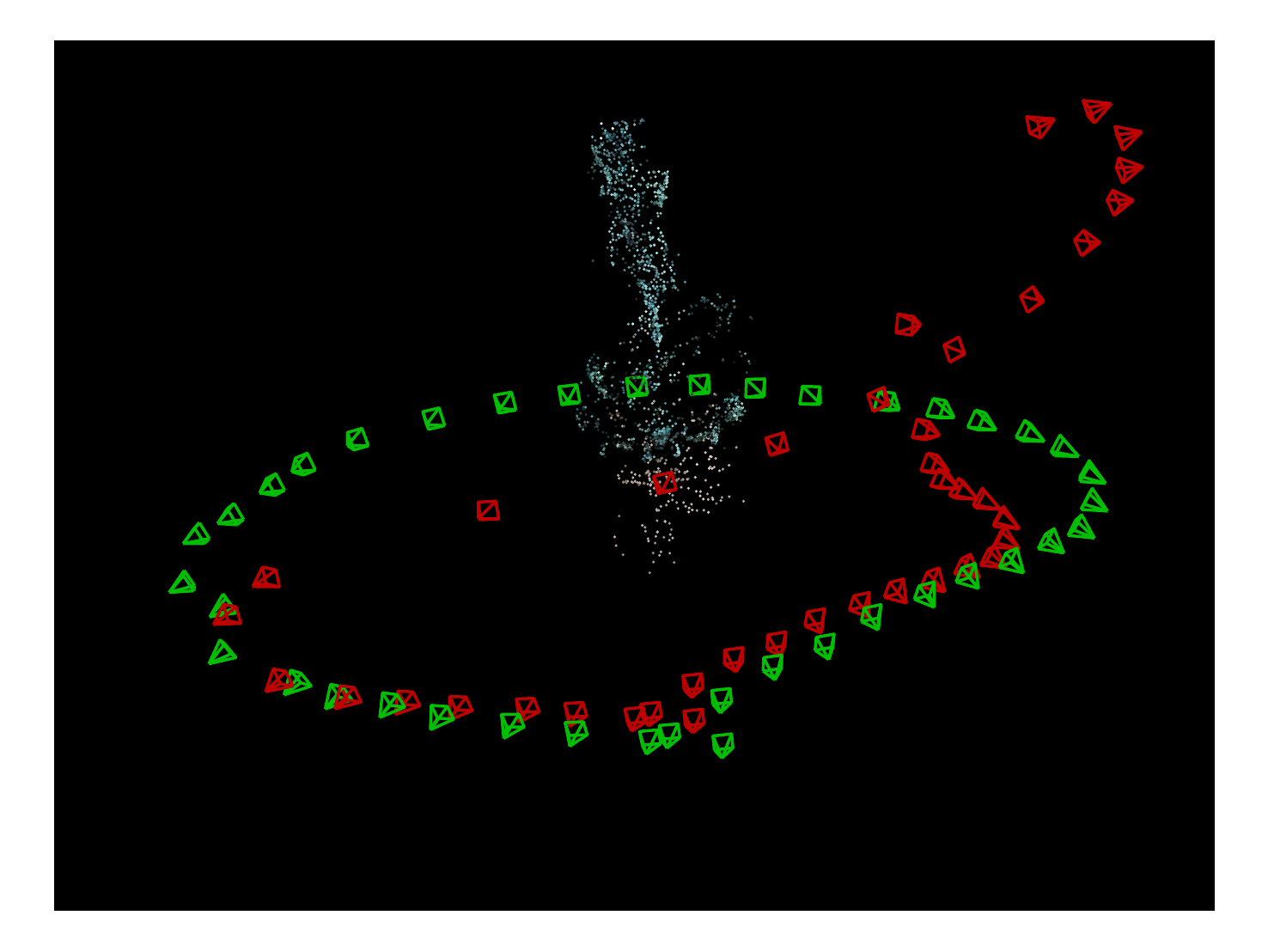} 
    \end{subfigure}
    \begin{subfigure}{0.5\textwidth}
    \centering
    \includegraphics[width=0.4\textwidth]{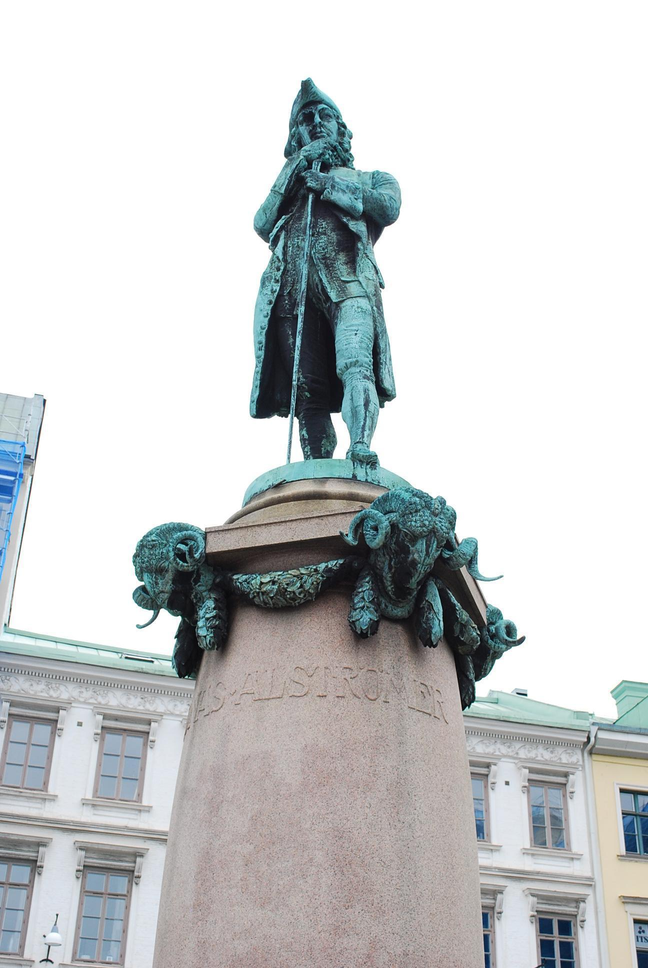}
    \end{subfigure}
\end{subfigure}
\caption{\small A failure case. Single scene 3D reconstruction and recovery of camera parameters  with our method applied to  Jonas Ahlstromer (reprojection error 8.41 pixels). The left image shows the triangulated point cloud, the recovered camera locations and orientations (in red) and the ground truth camera locations and orientations (in green). The right image is one of the input images.}
\label{fig:failure_reconstruction3}
\end{figure}

\begin{figure}[bth]
\centering

\begin{subfigure}{0.33\textwidth}
\centering
\includegraphics[width=\textwidth]{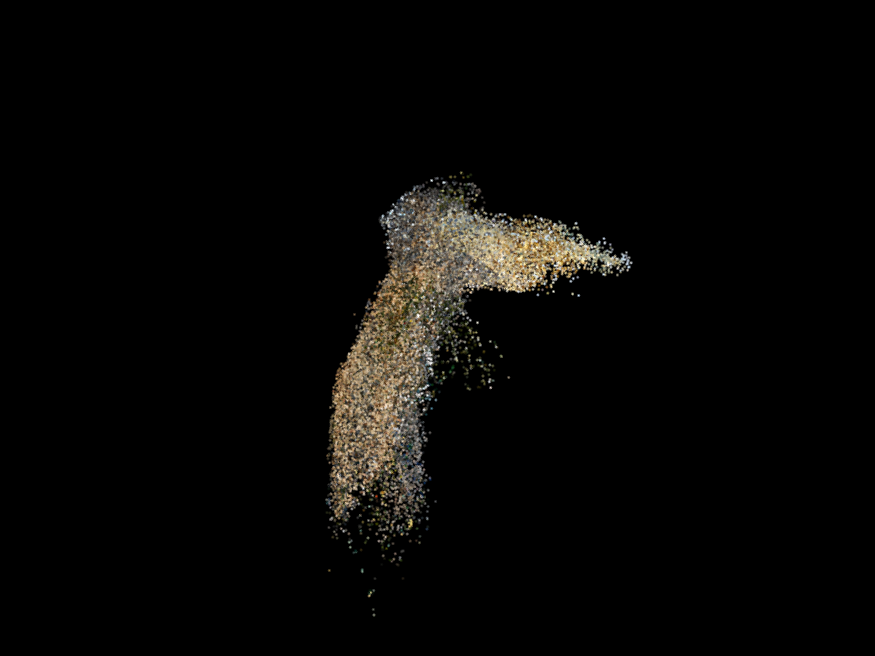} 
\caption*{Epoch 0}
\end{subfigure}
\begin{subfigure}{0.33\textwidth}
\centering
\includegraphics[width=\textwidth]{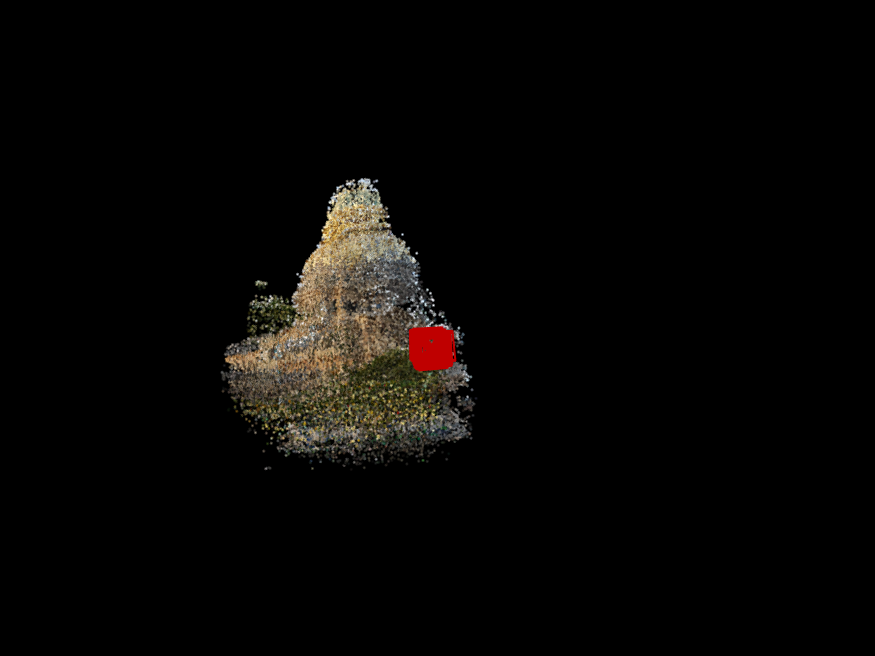} 
\caption*{Epoch 2000}
\end{subfigure}
\begin{subfigure}{0.33\textwidth}
\centering
\includegraphics[width=\textwidth]{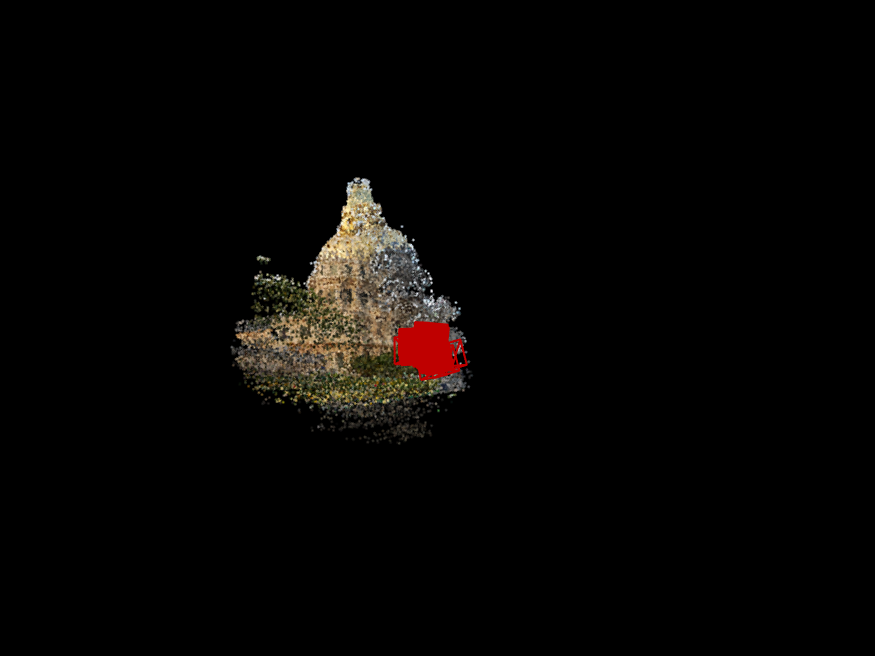} 
\caption*{Epoch 5000}
\end{subfigure}

\begin{subfigure}{0.33\textwidth}
\centering
\includegraphics[width=\textwidth]{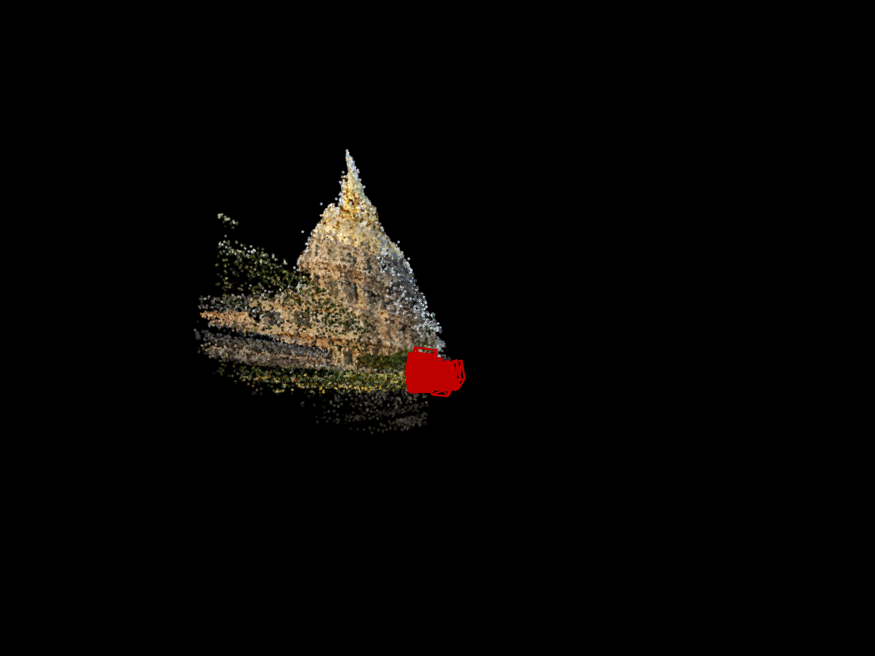} 
\caption*{Epoch 10000}
\end{subfigure}
\begin{subfigure}{0.33\textwidth}
\centering
\includegraphics[width=\textwidth]{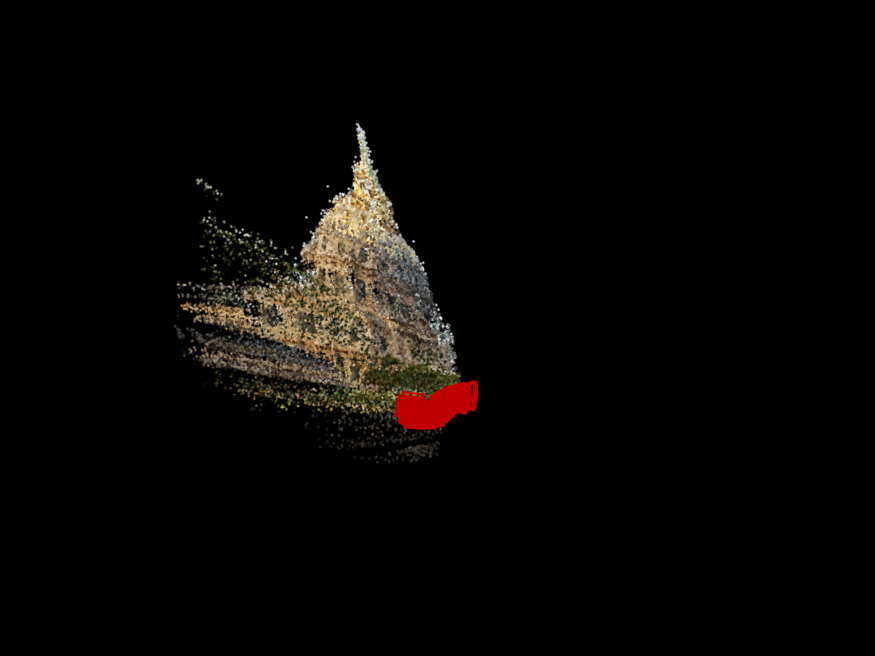} 
\caption*{Epoch 15000}
\end{subfigure}
\begin{subfigure}{0.33\textwidth}
\centering
\includegraphics[width=\textwidth]{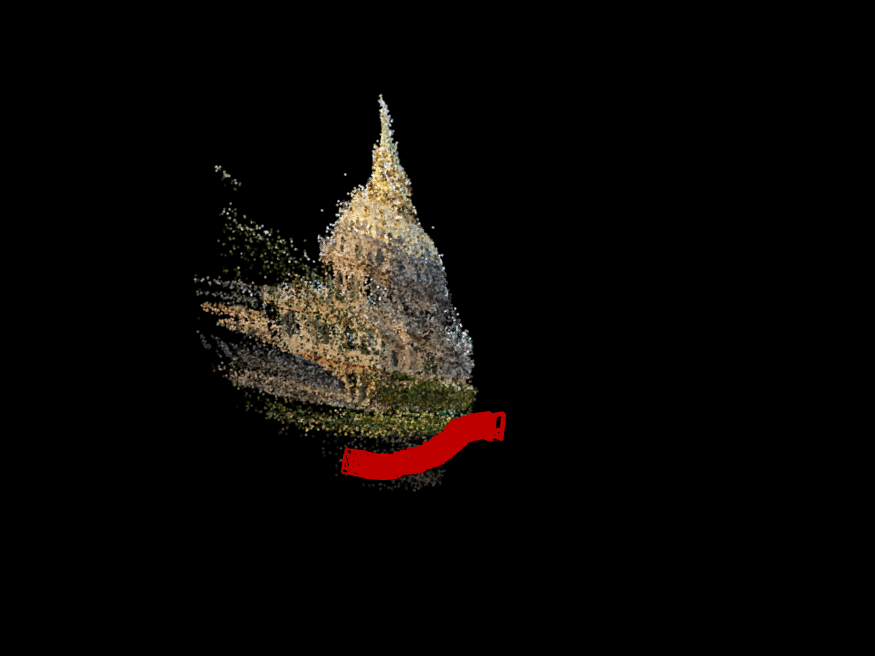} 
\caption*{Epoch 20000}
\end{subfigure}

\begin{subfigure}{0.33\textwidth}
\centering
\includegraphics[width=\textwidth]{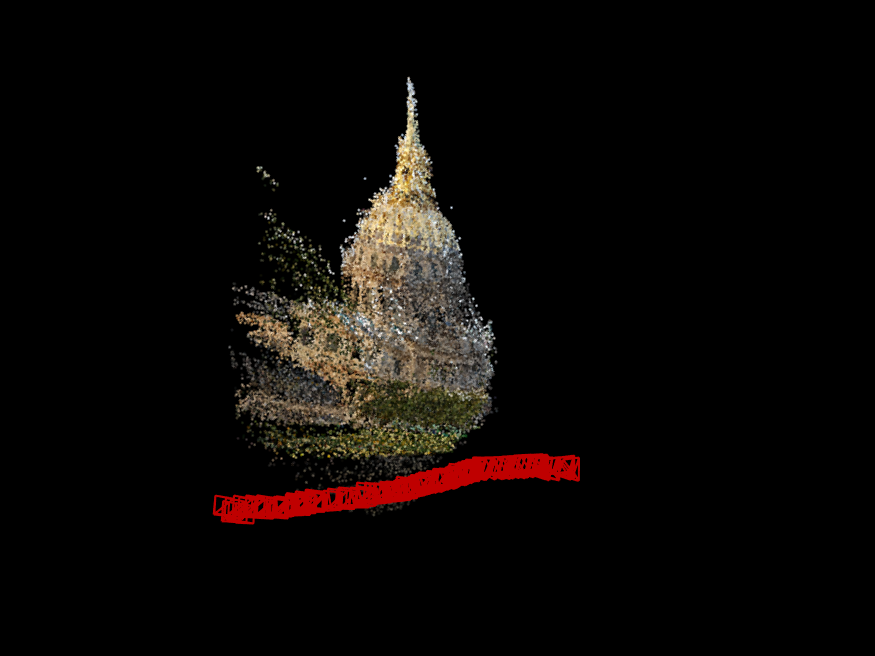} 
\caption*{Epoch 25000}
\end{subfigure}
\begin{subfigure}{0.33\textwidth}
\centering
\includegraphics[width=\textwidth]{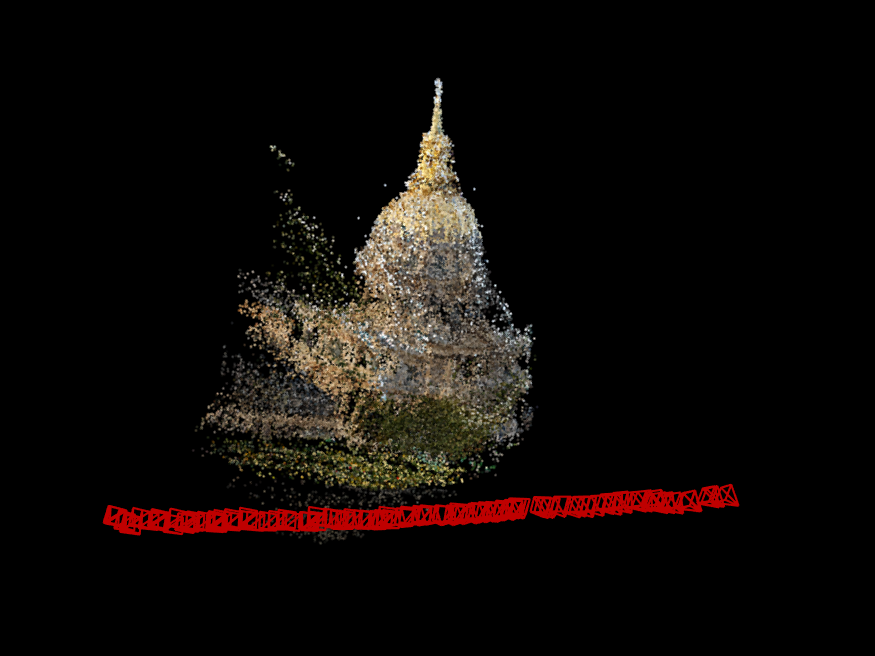} 
\caption*{Epoch 35000}
\end{subfigure}
\begin{subfigure}{0.33\textwidth}
\centering
\includegraphics[width=\textwidth]{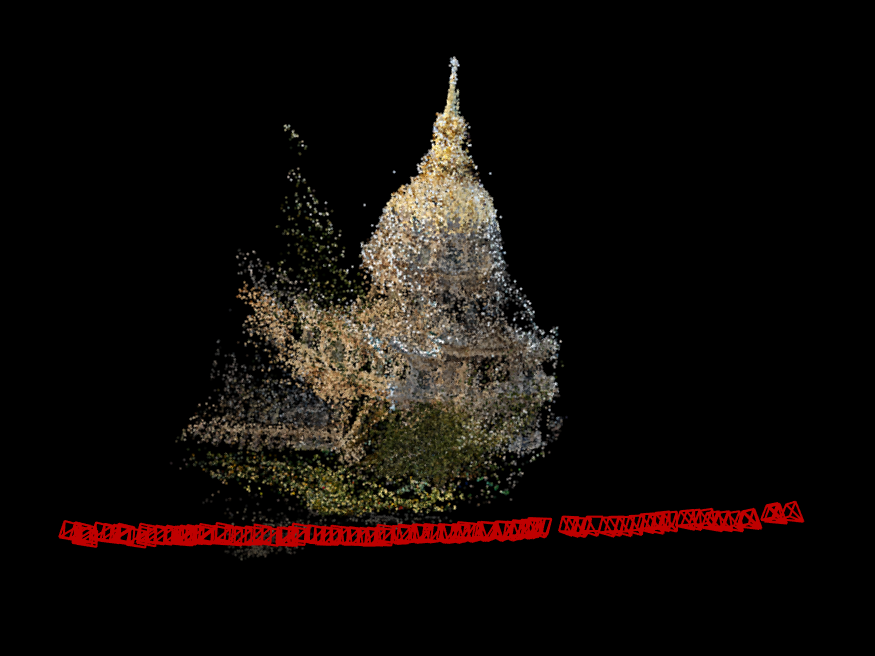}
\caption*{Epoch 45000}
\end{subfigure}

\begin{subfigure}{0.33\textwidth}
\centering
\includegraphics[width=\textwidth]{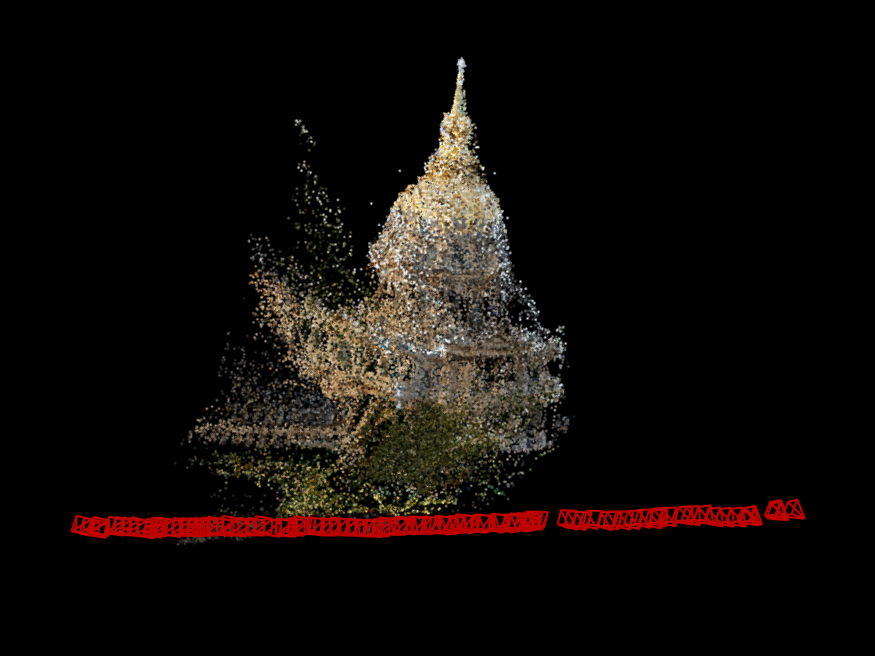} 
\caption*{Epoch 60000}
\end{subfigure}
\begin{subfigure}{0.33\textwidth}
\centering
\includegraphics[width=\textwidth]{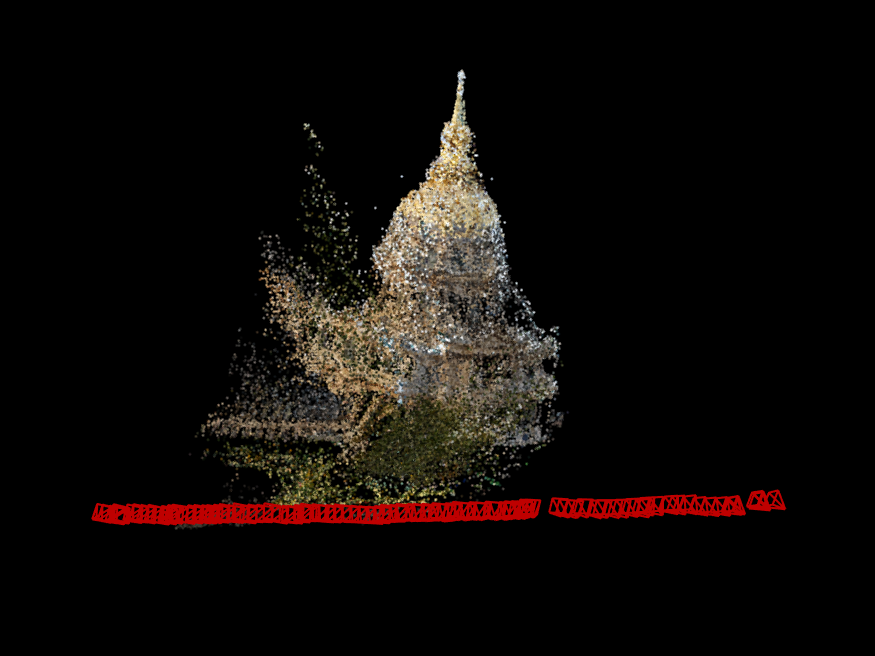}
\caption*{Epoch 70000}
\end{subfigure}
\begin{subfigure}{0.33\textwidth}
\centering
\includegraphics[width=\textwidth]{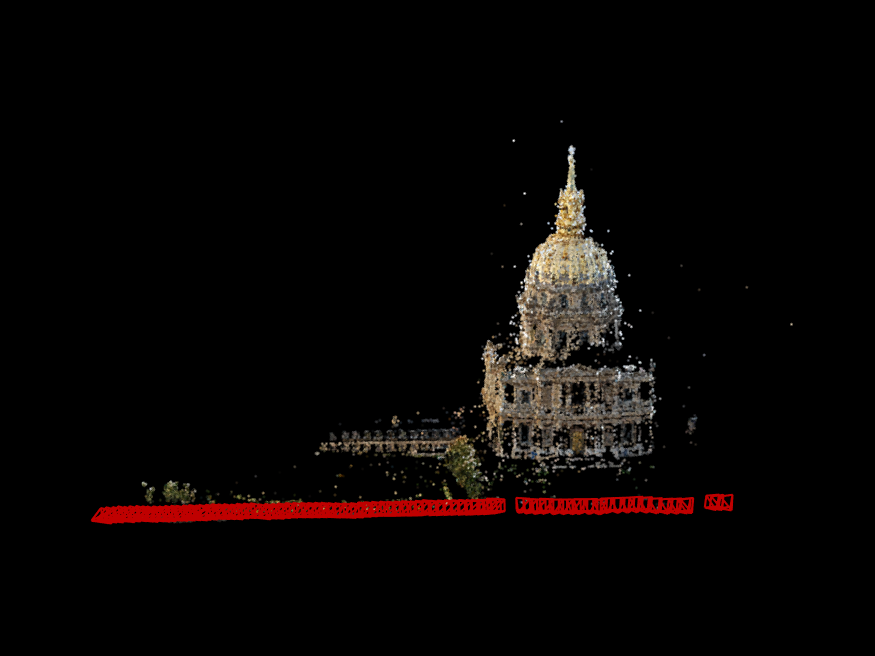}
\caption*{After BA}
\end{subfigure}

\caption{3D structure and camera parameter evolution during the optimization of the network.}
\label{fig:timeline}
\end{figure}

\paragraph{Single scene recovery.} In the single scene recovery mode given a single track tensor representing point correspondences across images of some scene we attempt to minimize the reprojection loss, where the network is used to parameterize the loss. 
Tables~\ref{tab:Projective_Results_No_BA} and~\ref{tab:Euclidean_Results_No_BA} show results of our method before and after bundle adjustment in the uncalibrated and calibrated settings. We compare our results before bundle adjustment only to global methods since sequential methods apply bundle adjustment in each iteration. Notably, already before bundle adjustment our method often achieves sub-pixel accuracies, significantly surpassing GPSFM in the uncalibrated setting and GESFM and Linear in the calibrated setting.
Figures \ref{fig:optimization_reconstruction}-\ref{fig:optimization_reconstruction3} show 3D reconstructions and camera parameter recovery in the calibrated setting. In addition, a failure case is shown in Figure \ref{fig:failure_reconstruction3}.
Figure \ref{fig:timeline} shows the evolution of structure and camera parameters during optimization. % A video footage of this evolution is provided in additional files (file names: \dm{Complete}).
%Finally, Figures \ref{fig:optimization_reconstruction}-\ref{fig:optimization_reconstruction3} show results obtained in the single scene scenario. 

\begin{table*}[tb]
    \hspace{-8pt}
    \setlength\tabcolsep{1pt} % default value: 6pt
    \tiny
    \centering
    \begin{tabular}{c}
        \begin{adjustbox}{max width=\textwidth}
        \aboverulesep=0ex
        \belowrulesep=0ex
        \renewcommand{\arraystretch}{0.9}
        \begin{tabular}[t]{|l|r|r||rrr|rrr|rrr|}
            \hline
            \multirow{3}{3em}{Scan}& \multirow{3}{3em}{\#Images} & \multirow{3}{3em}{\#Points} & \multicolumn{3}{c|}{$\tr_{\text{error}}$} & \multicolumn{3}{c|}{$R_{\text{error}}$} & \multicolumn{3}{c|}{Reprojection Error} \\
            \cline{4-12}
            & & & 
            \multicolumn{1}{c}{\textbf{Ours}} & \multirow{2}{3em}{\textbf{Ours}} & \multirow{2}{3.5em}{Colmap} & \multicolumn{1}{c}{\textbf{Ours}} & \multirow{2}{3em}{\textbf{Ours}} & \multirow{2}{3.5em}{Colmap} & \multicolumn{1}{c}{\textbf{Ours}} & \multirow{2}{3em}{\textbf{Ours}} & \multirow{2}{3.5em}{Colmap}  \\
            & & & 
            \multicolumn{1}{c}{\textbf{No BA}} & & & 
            \multicolumn{1}{c}{\textbf{No BA}} & & & 
            \multicolumn{1}{c}{\textbf{No BA}} & & \\
			\hline
Folke Filbyter & $ 40 $ & $ 21150 $ & $ 0.093 $ & $ 0.037 $ & $\mathbf{0.0}$ & $ 54.025 $ & $ 20.51 $ & $\mathbf{0.036}$ & $ 9.78 $ & $ 3.87 $ & $\mathbf{0.29}$\\
Gustav Vasa & $ 18 $ & $ 4249 $ & $ 0.193 $ & $\mathbf{0.099}$ & $ 0.1 $ & $ 2.964 $ & $\mathbf{0.839}$ & $ 0.841 $ & $ 0.62 $ & $\mathbf{0.48}$ & $\mathbf{0.48}$\\
GustavIIAdolf & $ 57 $ & $ 5813 $ & $ 0.014 $ & $\mathbf{0.004}$ & $\mathbf{0.004}$ & $ 0.068 $ & $\mathbf{0.021}$ & $\mathbf{0.021}$ & $ 0.29 $ & $\mathbf{0.26}$ & $\mathbf{0.26}$\\
Jonas Ahlstromer & $ 40 $ & $ 2021 $ & $ 0.018 $ & $\mathbf{0.011}$ & $\mathbf{0.011}$ & $ 0.051 $ & $ 0.037 $ & $\mathbf{0.036}$ & $ 0.24 $ & $\mathbf{0.22}$ & $\mathbf{0.22}$\\
Plaza De Armas Santiago & $ 240 $ & $ 26969 $ & $\mathbf{0.044}$ & $ 0.048 $ & $ 0.048 $ & $\mathbf{0.089}$ & $ 0.121 $ & $ 0.122 $ & $ 1.3 $ & $\mathbf{1.13}$ & $\mathbf{1.13}$\\
Sri Thendayuthapani Singapore & $ 98 $ & $ 88849 $ & $ 0.057 $ & $\mathbf{0.034}$ & $\mathbf{0.034}$ & $ 0.222 $ & $ 0.139 $ & $\mathbf{0.138}$ & $ 0.77 $ & $\mathbf{0.67}$ & $\mathbf{0.67}$\\
Statue Of Liberty & $ 134 $ & $ 49250 $ & $ 8.558 $ & $ 1.877 $ & $\mathbf{0.099}$ & $ 13.262 $ & $ 3.02 $ & $\mathbf{0.213}$ & $ 6.86 $ & $ 1.76 $ & $\mathbf{1.25}$\\
The Pumpkin & $ 196 $ & $ 69341 $ & $ 0.17 $ & $\mathbf{0.022}$ & $\mathbf{0.022}$ & $ 0.851 $ & $ 0.092 $ & $\mathbf{0.091}$ & $ 0.7 $ & $\mathbf{0.57}$ & $\mathbf{0.57}$\\
Tsar Nikolai I & $ 98 $ & $ 37857 $ & $ 0.024 $ & $\mathbf{0.005}$ & $\mathbf{0.005}$ & $ 0.092 $ & $\mathbf{0.018}$ & $\mathbf{0.018}$ & $ 0.38 $ & $\mathbf{0.33}$ & $\mathbf{0.33}$\\
Urban II & $ 96 $ & $ 22284 $ & $ 0.074 $ & $\mathbf{0.021}$ & $\mathbf{0.021}$ & $ 0.327 $ & $\mathbf{0.107}$ & $\mathbf{0.107}$ & $ 0.53 $ & $\mathbf{0.38}$ & $\mathbf{0.38}$\\

    \bottomrule
	\end{tabular} 
	\end{adjustbox}
	\vspace{3pt}      
    \end{tabular}
    \caption{\small Single scene results using sequential optimization in the calibrated setup. The table show results before and after bundle adjustment compared to Colmap. The table shows mean camera location error (denoted $\tr_{\mathrm{error}}$) in meters, mean orientation error (denoted $R_{\mathrm{error}}$) in degrees, and mean reprojection error in pixels. (\textit{Smaller is better.})}
    \label{tab:Euclidean_Seq_Results}
\end{table*}
\paragraph{Sequential optimization.} In some experiments, as can be seen in Table~\ref{tab:Euclidean_Results_No_BA}, our single scene recovery procedure failed to produce accurate reconstruction. In these cases (we declared failure if the reprojection error exceeded 2 pixels) we applied instead optimization at a sequential schedule. For this schedule we ordered the images greedily by the number of point tracks they share with the images that precede them in this order. Using this order, we first ran 500 optimization epochs with just the first 2 images. Then, after each 500 more epochs we add to this subset the next image in the order. As can be seen  in Table \ref{tab:Euclidean_Seq_Results}, this optimization schedule improved the reprojection error for all the failed datasets, yielding in most cases comparable accuracies to those obtained with Colmap.

% Optimization
\clearpage

\begin{figure}[bth]
\begin{subfigure}{1\textwidth}
    \begin{subfigure}{0.5\textwidth}
    \centering
    \includegraphics[width=\textwidth]{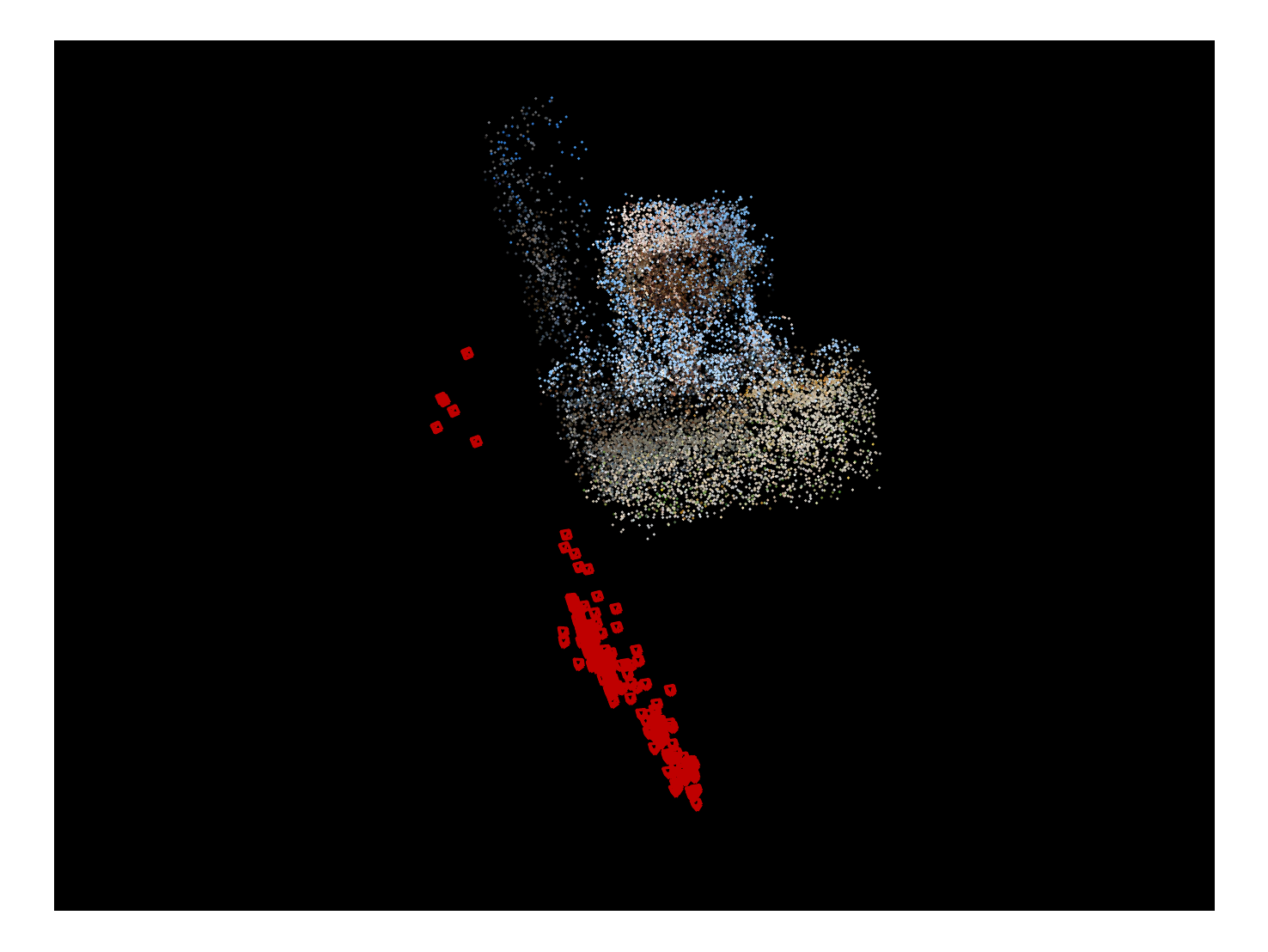} 
    \end{subfigure}
    \begin{subfigure}{0.5\textwidth}
    \centering
    \includegraphics[width=\textwidth]{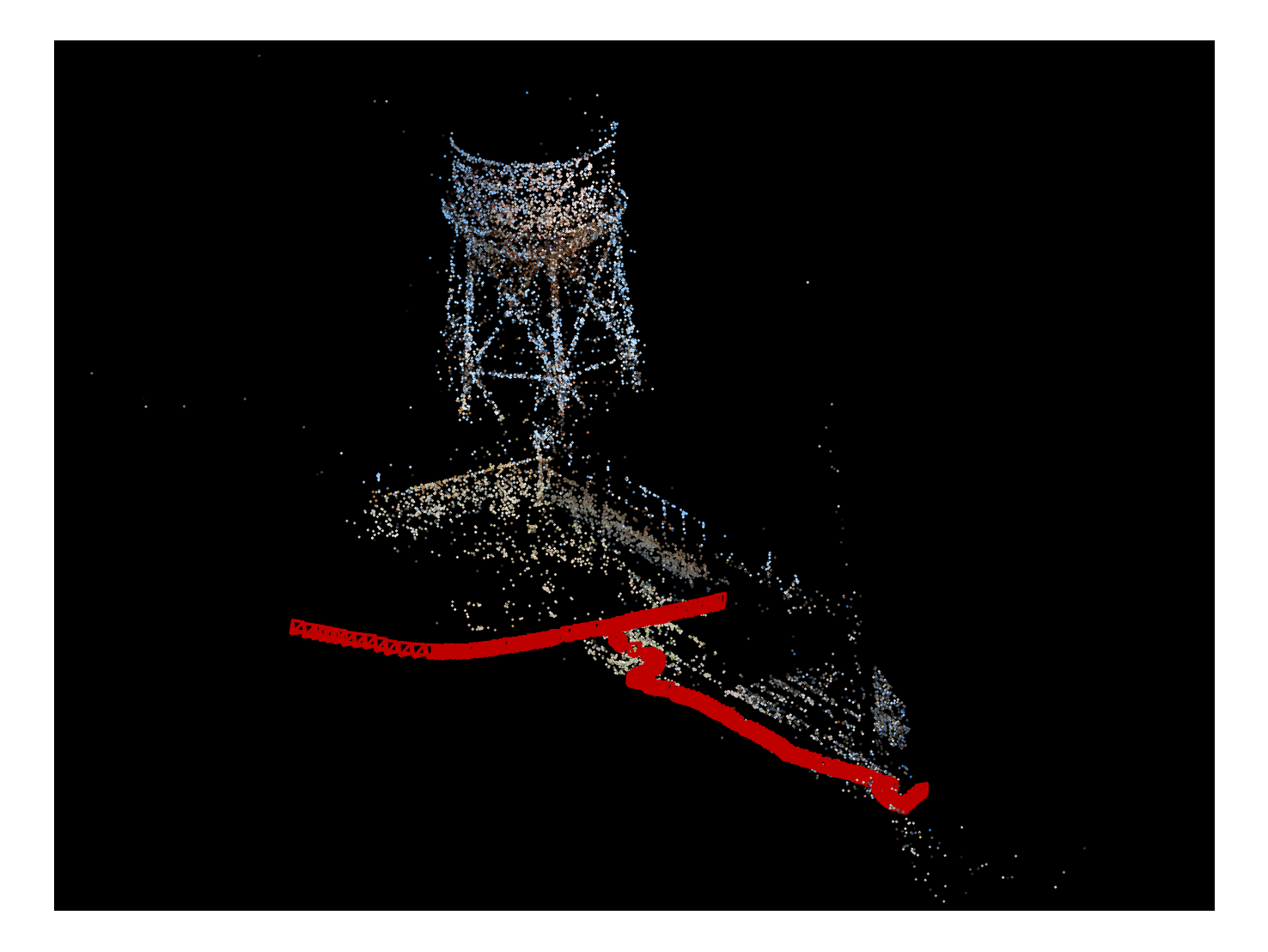} 
    \end{subfigure}
    \caption{Inference before (left) and after (right) BA}
\end{subfigure}

\begin{subfigure}{1\textwidth}
    \begin{subfigure}{0.5\textwidth}
    \centering
    \includegraphics[width=\textwidth]{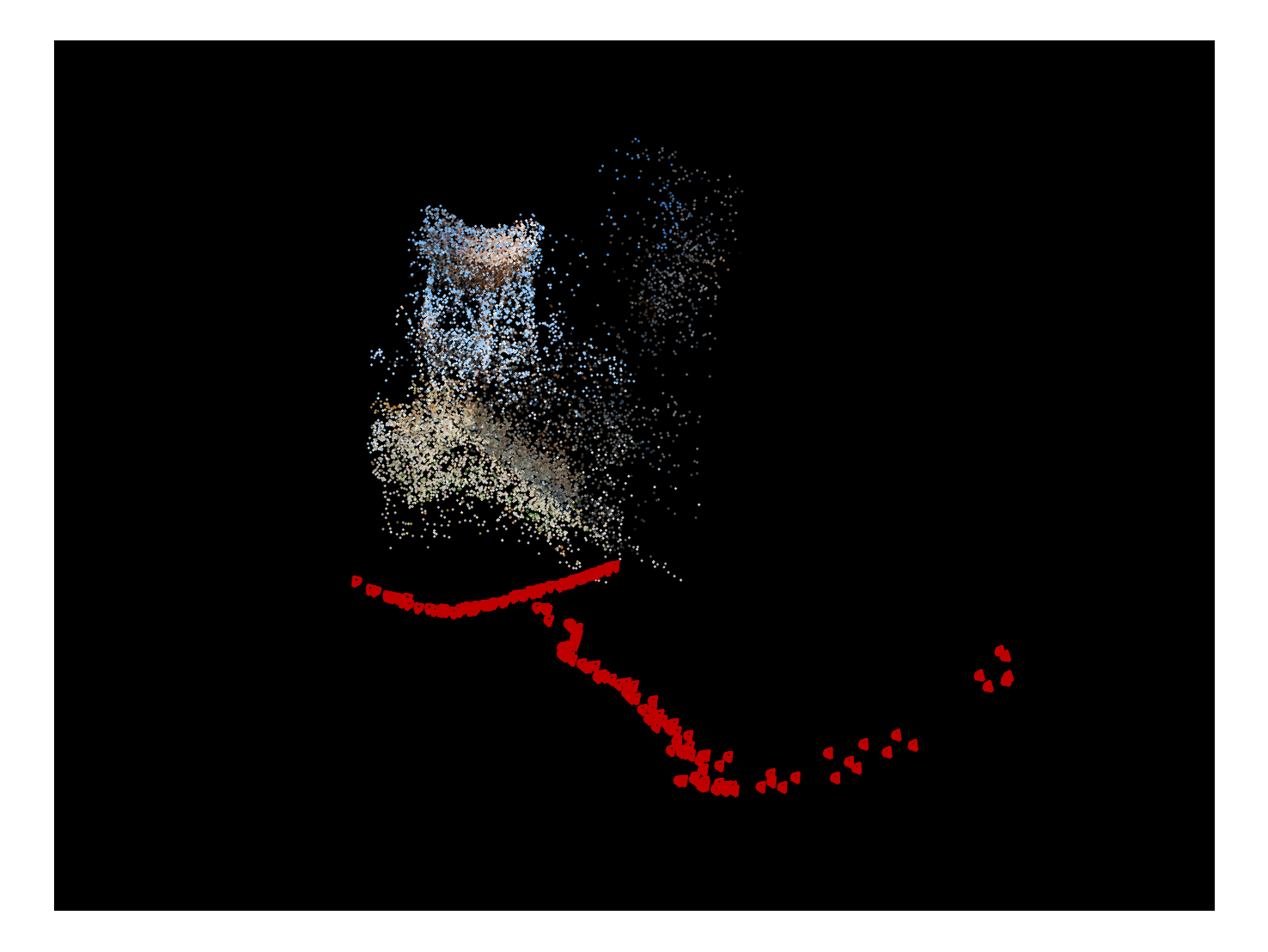} 
    \end{subfigure}
    \begin{subfigure}{0.5\textwidth}
    \centering
    \includegraphics[width=\textwidth]{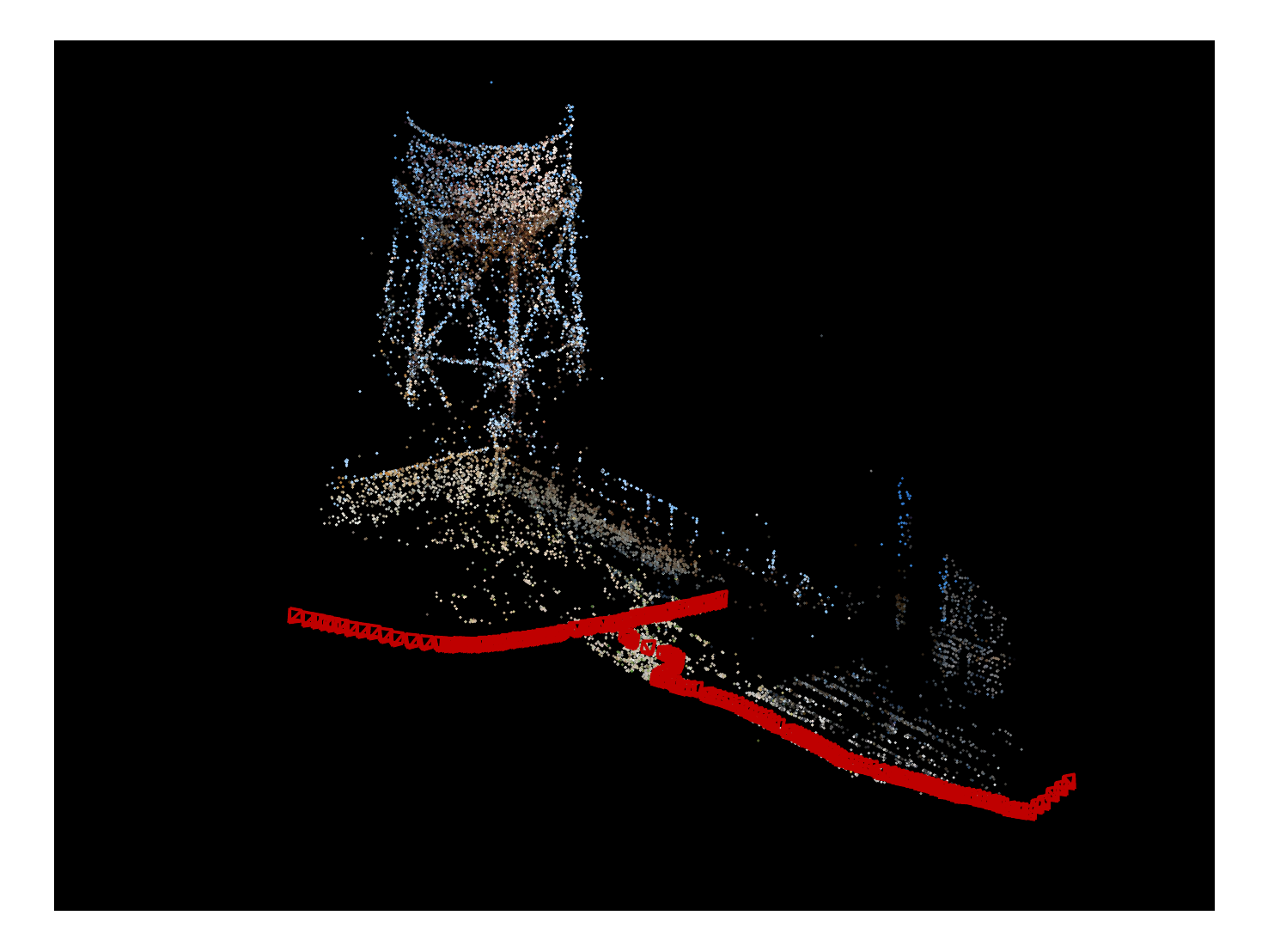} 
    \end{subfigure}
    \caption{Inference + fine tuning before (left) and after (right) BA}
\end{subfigure}

\begin{subfigure}{1\textwidth}
    \begin{subfigure}{0.5\textwidth}
    \centering
    \includegraphics[width=\textwidth]{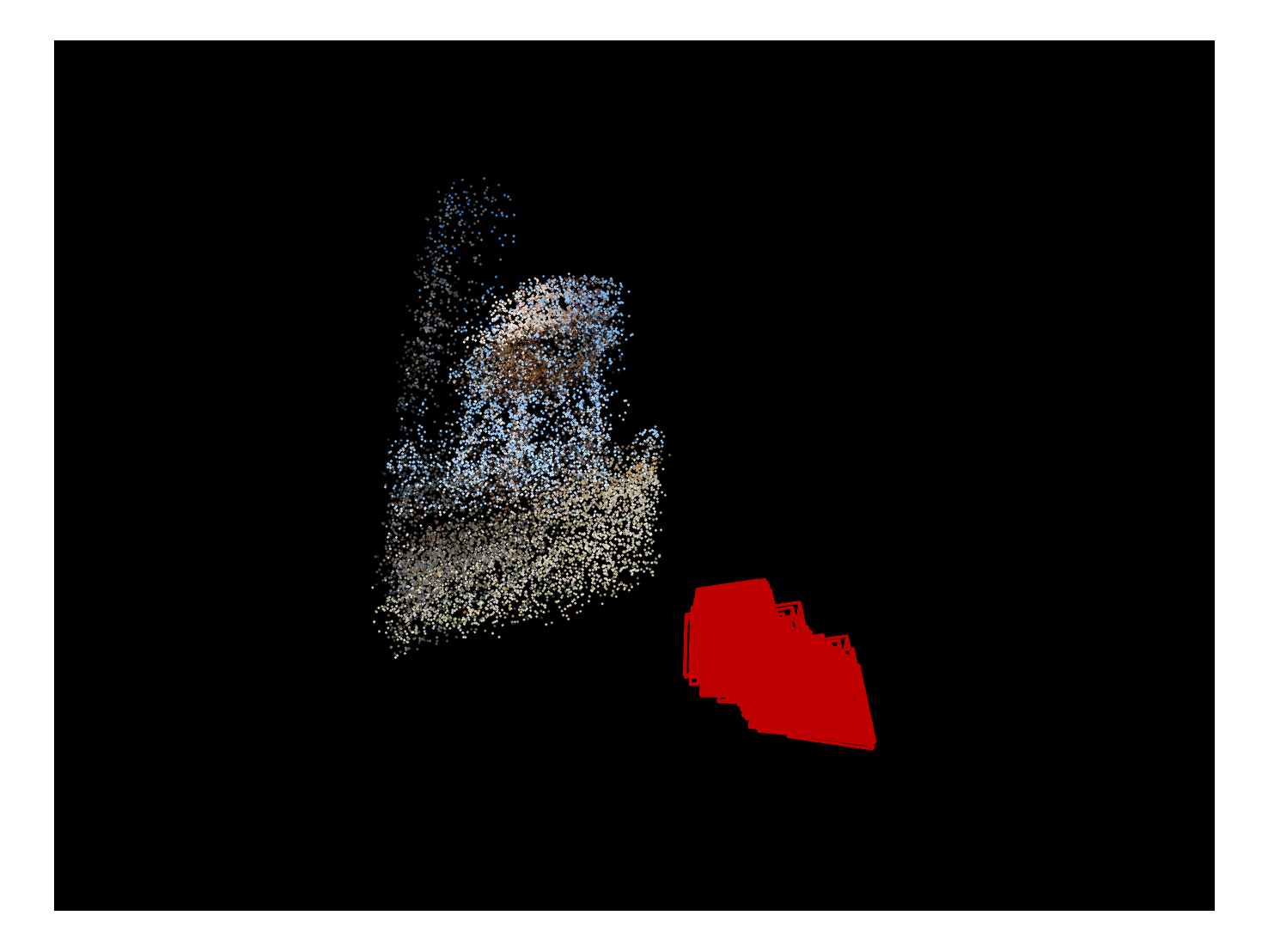} 
    \end{subfigure}
    \begin{subfigure}{0.5\textwidth}
    \centering
    \includegraphics[width=\textwidth]{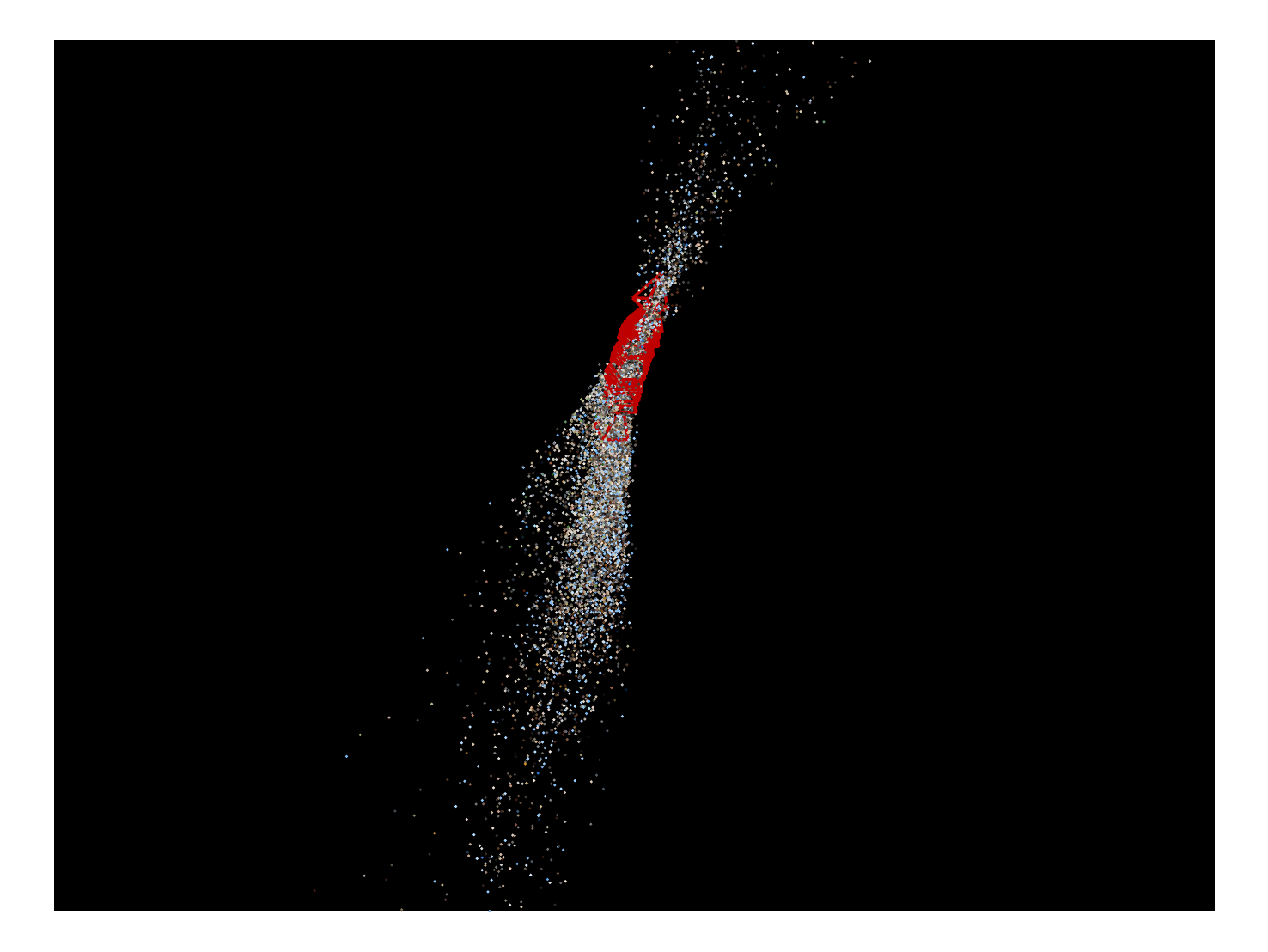} 
    \end{subfigure}
    \caption{Short optimization before (left) and after (right) BA}
\end{subfigure}
\caption{Alcatraz Water Tower. Reconstruction with our trained model. The figure shows results of inference (top row) and inference followed by fine tuning (middle row) before BA (left) and after BA (right). The bottom row shows the result of the short optimization strategy (starting with a random initialization). Each panel shows the recovered cameras positions (in red) and the recovered 3D points, corresponding to the point tracks.  It can be seen that in this case  accurate reconstruction can be obtained either by  pure  inference or inference followed by  fine tuning (+ BA). In contrast, short optimization failed to accurately recover camera positions, leading to failure of the BA.}
\label{fig:Learning_Alcatraz_Water_Tower}
\end{figure}

\begin{figure}[bth]
\begin{subfigure}{1\textwidth}
    \begin{subfigure}{0.5\textwidth}
    \centering
    \includegraphics[width=\textwidth]{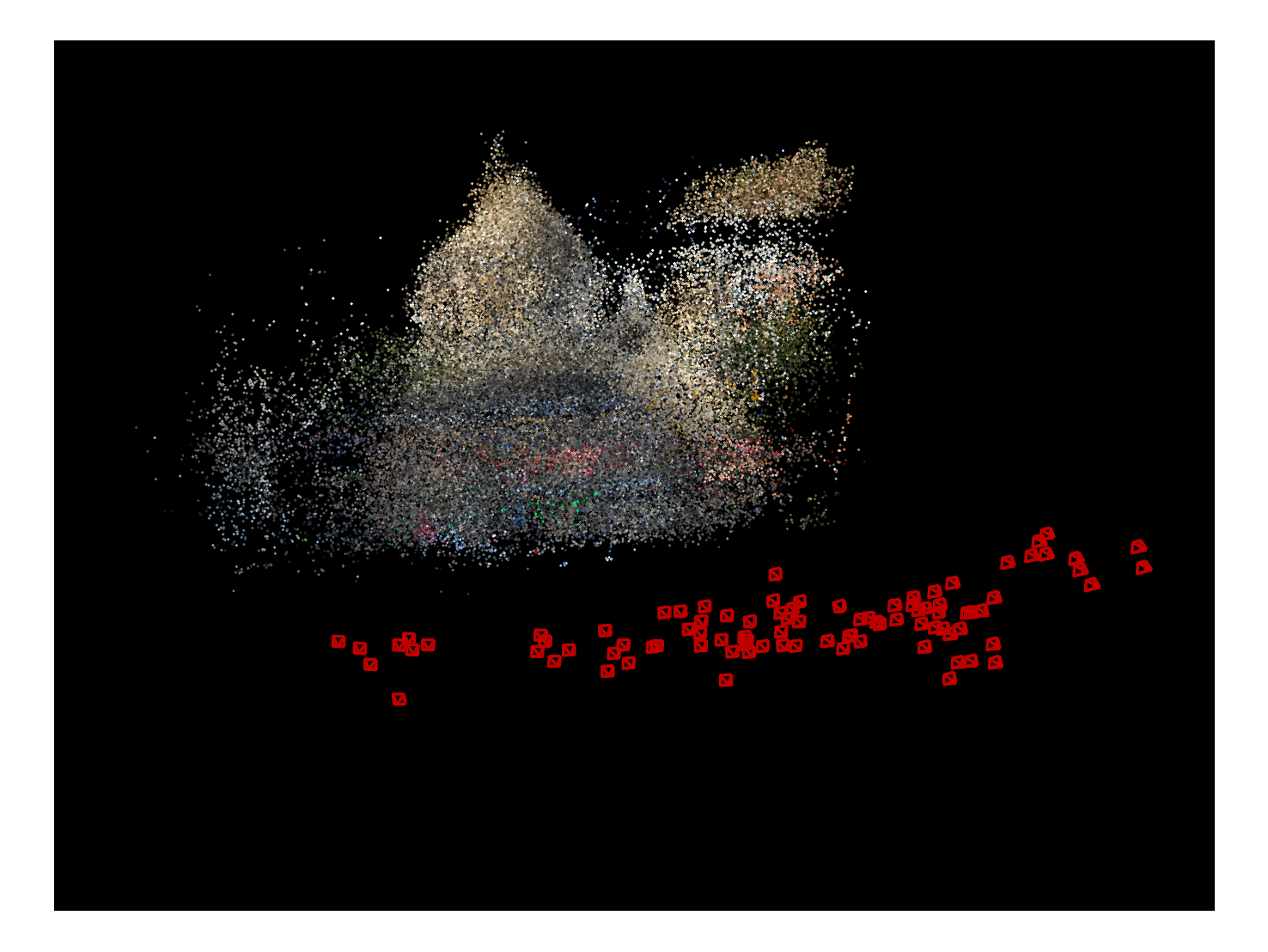} 
    \end{subfigure}
    \begin{subfigure}{0.5\textwidth}
    \centering
    \includegraphics[width=\textwidth]{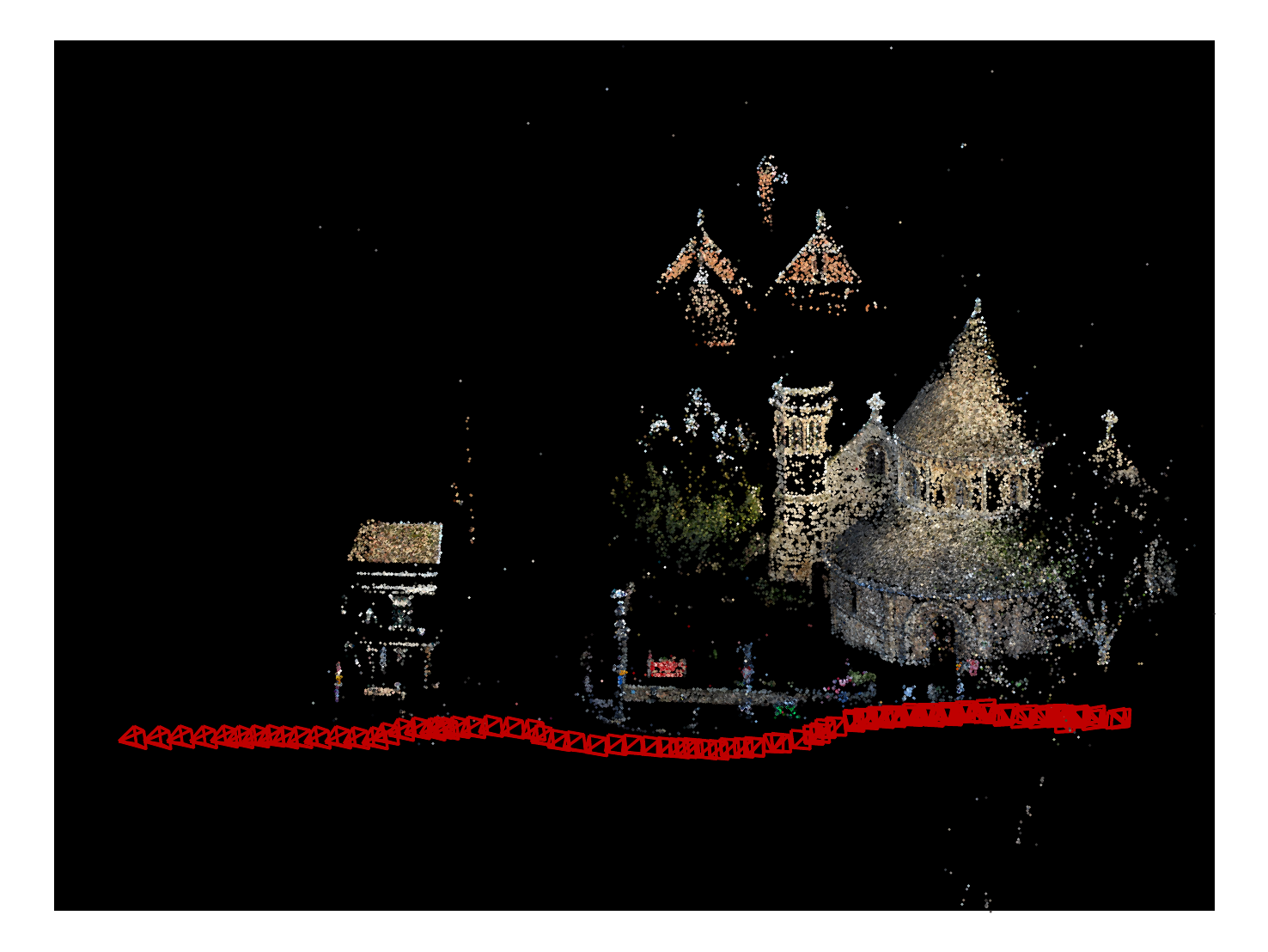} 
    \end{subfigure}
    \caption{Inference before (left) and after (right) BA}
\end{subfigure}

\begin{subfigure}{1\textwidth}
    \begin{subfigure}{0.5\textwidth}
    \centering
    \includegraphics[width=\textwidth]{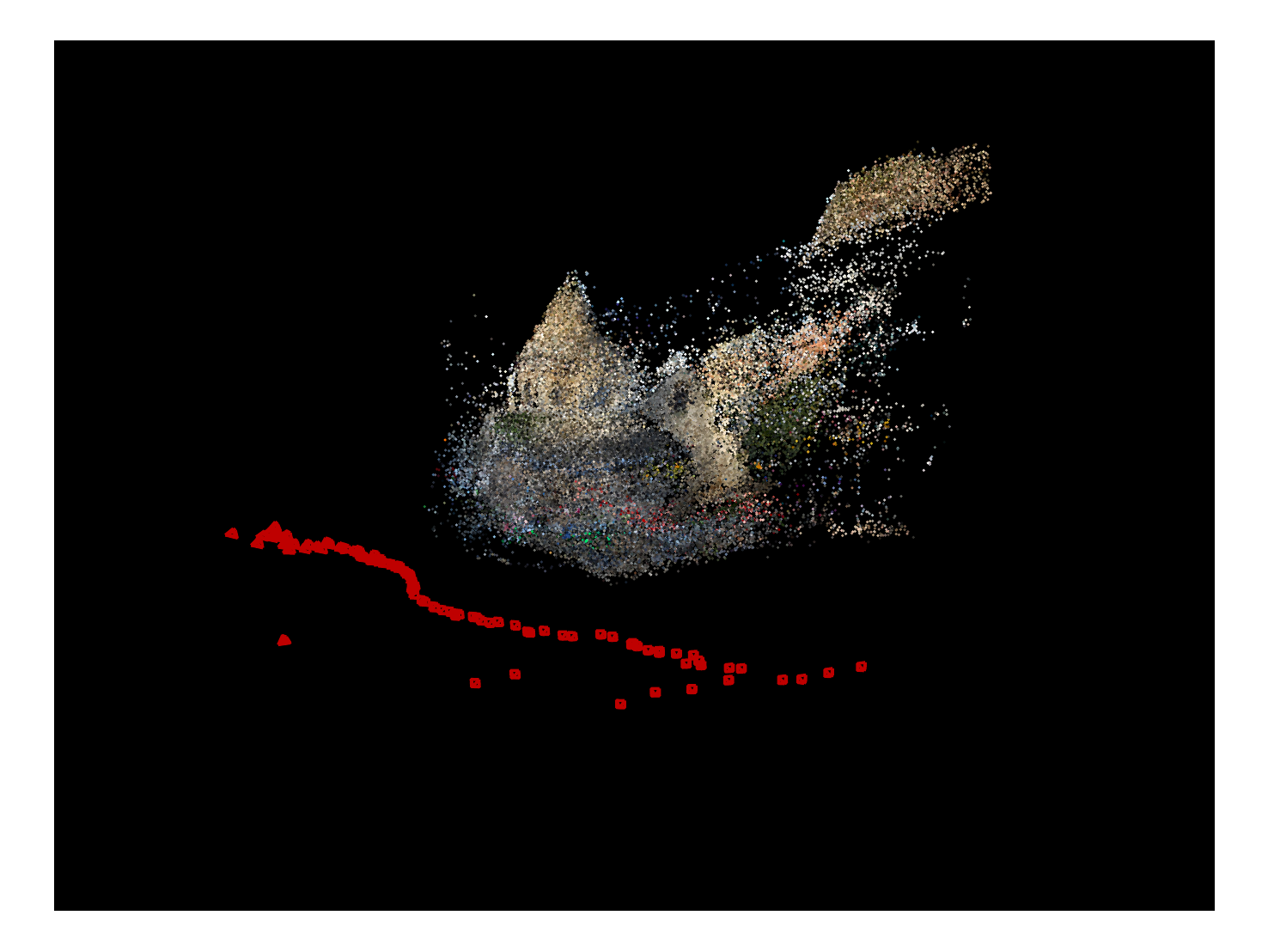} 
    \end{subfigure}
    \begin{subfigure}{0.5\textwidth}
    \centering
    \includegraphics[width=\textwidth]{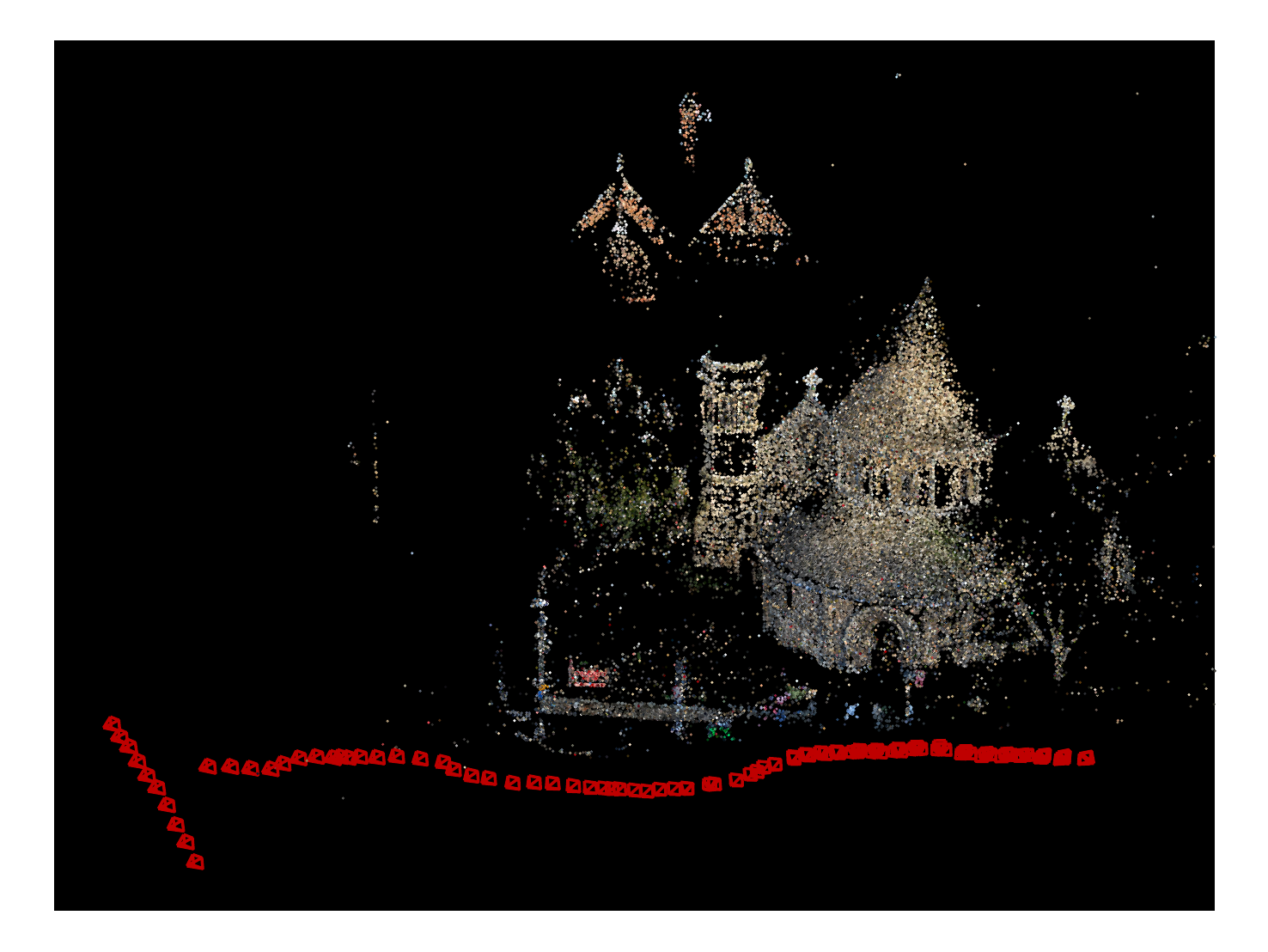} 
    \end{subfigure}
    \caption{Inference + fine tuning before (left) and after (right) BA}
\end{subfigure}

\begin{subfigure}{1\textwidth}
    \begin{subfigure}{0.5\textwidth}
    \centering
    \includegraphics[width=\textwidth]{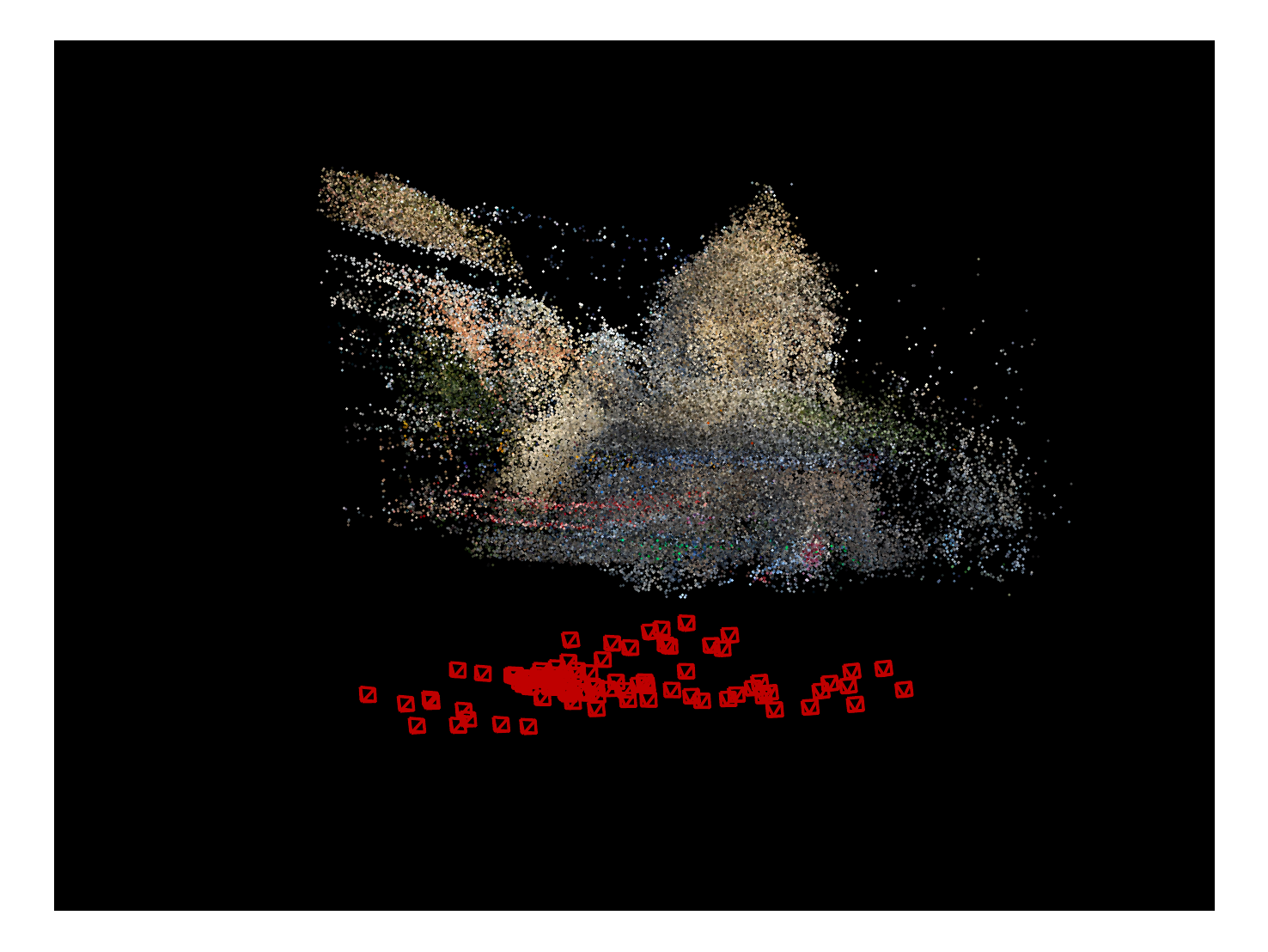} 
    \end{subfigure}
    \begin{subfigure}{0.5\textwidth}
    \centering
    \includegraphics[width=\textwidth]{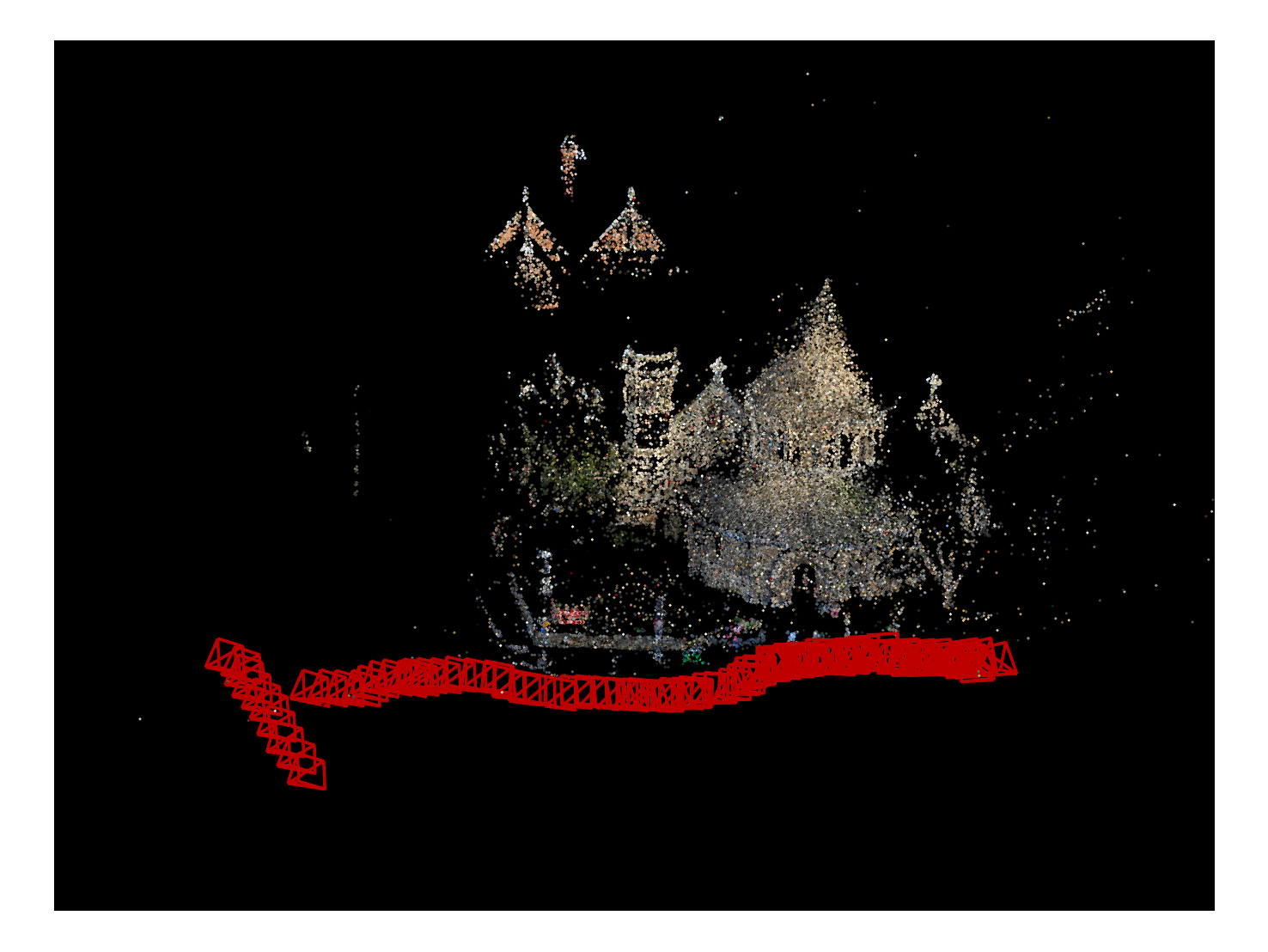} 
    \end{subfigure}
    \caption{Short optimization before (left) and after (right) BA}
\end{subfigure}
\caption{Round Church Cambridge. Reconstruction with our trained model. The figure shows results of inference (top row) and fine tuning (middle row) before BA (left) and after BA (right). The bottom row shows the result of the short optimization (starting with a random initialization). Each panel shows the recovered cameras positions (in red) and 3D points. Here fine tuning + BA yielded the most accurate reconstruction.}
\label{fig:Learning_Round_Church_Cambridge}
\end{figure}

\paragraph{Learning from multiple scenes.}
Figures \ref{fig:Learning_Alcatraz_Water_Tower} and \ref{fig:Learning_Round_Church_Cambridge} show  reconstruction results using our model before and after bundle adjustment in 3 scenarios: (i) inference using our trained model (ii) inference followed by fine tuning and (iii) short run of  optimization. Table~\ref{tab:Generalizing_times} shows execution times for our trained model. We note that using inference only yields a good initialization for bundle adjustment in a small fraction of a second. Using fine tuning yields more accurate results (See Table 2 in the paper) with execution times similar to Colmap. The short optimization generally yields less accurate results with execution times similar to fine tuning, emphasizing the importance of the trained model.

\begin{table*}[t]
    \hspace{-8pt}
    \setlength\tabcolsep{2pt} % default value: 6pt
    %\small
    \centering
    \scriptsize
    
    \begin{tabular}{c}
        \begin{adjustbox}{max width=\textwidth}
        \aboverulesep=0ex
        \belowrulesep=0ex
        \renewcommand{\arraystretch}{1}
        \begin{tabular}[t]{|l|r|r|rrrr|}
            \hline
            \multirow{2}{3em}{Scan}& \multirow{2}{3em}{\#Images} & \multirow{2}{3em}{\#Points} & \multicolumn{4}{c|}{\textbf{Time (seconds)}} \\
            \cline{4-7}
            & & & Inference & Fine tuning & BA & Colmap\\
            \hline
            Alcatraz Courtyard & 133 & 23674 & $ 0.007 $ & $ 199.125 $ & $ 43.512 $ & $ 286.0 $\\
Alcatraz Water Tower & 172 & 14828 & $ 0.007 $ & $ 110.847 $ & $ 26.44 $ & $ 130.0 $\\
Drinking Fountain Somewhere In Zurich & 14 & 5302 & $ 0.007 $ & $ 20.302 $ & $ 2.925 $ & $ 16.0 $\\
Nijo Castle Gate & 19 & 7348 & $ 0.008 $ & $ 24.493 $ & $ 4.308 $ & $ 21.0 $\\
Porta San Donato Bologna & 141 & 25490 & $ 0.007 $ & $ 194.651 $ & $ 45.416 $ & $ 170.0 $\\
Round Church Cambridge & 92 & 84643 & $ 0.014 $ & $ 360.97 $ & $ 90.092 $ & $ 229.0 $\\
Smolny Cathedral St Petersburg & 131 & 51115 & $ 0.004 $ & $ 534.528 $ & $ 101.456 $ & $ 516.0 $\\
Some Cathedral In Barcelona & 177 & 30367 & $ 0.007 $ & $ 208.542 $ & $ 55.424 $ & $ 451.0 $\\
Sri Veeramakaliamman Singapore & 157 & 130013 & $ 0.319 $ & $ 291.727 $ & $ 242.888 $ & $ 583.0 $\\
Yueh Hai Ching Temple Singapore & 43 & 13774 & $ 0.004 $ & $ 45.458 $ & $ 10.539 $ & $ 106.0 $\\
			\bottomrule
	\end{tabular}

\begin{tabular}[t]{|l|r|r|rrrr|}
            \hline
            \multirow{2}{3em}{Scan}& \multirow{2}{3em}{\#Images} & \multirow{2}{3em}{\#Points} & \multicolumn{4}{c|}{\textbf{Time (seconds)}} \\
            \cline{4-7}
            & & & Inference & Fine tuning & BA & GPSFM\\
            \hline
Alcatraz Water Tower & 172 & 14828 & $ 0.055 $ & $ 89.646 $ & $ 68.939 $ & $ 137.057 $\\
Dino 319 & 36 & 319 & $ 0.004 $ & $ 8.151 $ & $ 0.475 $ & $ 3.253 $\\
Dino 4983 & 36 & 4983 & $ 0.122 $ & $ 7.66 $ & $ 2.21 $ & $ 4.994 $\\
Dome & 85 & 84792 & $ 0.203 $ & $ 160.896 $ & $ 76.867 $ & $ 105.837 $\\
Drinking Fountain & 14 & 5302 & $ 0.01 $ & $ 18.197 $ & $ 3.016 $ & $ 3.348 $\\
Gustav Vasa & 18 & 4249 & $ 0.007 $ & $ 14.435 $ & $ 2.766 $ & $ 3.449 $\\
Nijo & 19 & 7348 & $ 0.013 $ & $ 34.397 $ & $ 3.121 $ & $ 6.37 $\\
Skansen Kronan & 131 & 28371 & $ 0.141 $ & $ 197.531 $ & $ 63.853 $ & $ 93.831 $\\
Some Cathedral In Barcelona & 177 & 30367 & $ 0.133 $ & $ 185.984 $ & $ 47.597 $ & $ 110.485 $\\
Sri Veeramakaliamman Singapore & 157 & 130013 & $ 0.314 $ & $ 473.294 $ & $ 195.374 $ & $ 301.713 $\\
			\bottomrule
	\end{tabular} 
	
	\end{adjustbox}
	\vspace{3pt}      
    \end{tabular}
    \caption{\small Execution times for our trained model. The table shows execution times in seconds in the calibrated (left) and uncalibrated (right) settings. 
    }
    \label{tab:Generalizing_times}
\end{table*}

\newpage

% Learning

\end{document}